\documentclass[10pt,twocolumn,letterpaper]{article}

\usepackage[pagenumbers]{cvpr} 

\definecolor{cvprblue}{rgb}{0.21,0.49,0.74}
\usepackage[pagebackref,breaklinks,colorlinks,allcolors=cvprblue]{hyperref}
\usepackage{multirow}
\usepackage{multicol}
\usepackage{xcolor}
\usepackage{fontawesome}
\usepackage{colortbl}
\usepackage{subcaption}
\usepackage[most]{tcolorbox}
\usepackage{utfsym}
\tcbuselibrary{breakable}

\def\paperID{18476} 
\def\confName{CVPR}
\def\confYear{2026}

\title{LEMON: How Well Do MLLMs Perform Temporal Multimodal Understanding on Instructional Videos?}

\author{Zhuang Yu, Lei Shen\\
Shanghai Jiao Tong University
\and 
Jing Zhao\\
East China Normal University
\and 
Shiliang Sun*\\
Shanghai Jiao Tong University\\
{\tt\small shiliangsun@gmail.com}
}

\begin{document}
\maketitle

\begin{abstract}
Recent multimodal large language models (MLLMs) have shown remarkable progress across vision, audio, and language tasks, yet their performance on long-form, knowledge-intensive, and temporally structured educational content remains largely unexplored. To bridge this gap, we introduce \textbf{LEMON}, a \textbf{L}ecture-based \textbf{E}valuation benchmark for \textbf{M}ultim\textbf{O}dal u\textbf{N}derstanding, focusing on STEM lecture videos that require long-horizon reasoning and cross-modal integration. LEMON comprises 2,277 video segments spanning 5 disciplines and 29 courses, with an average duration of 196.1 seconds, yielding 4,181 high-quality QA pairs, including 3,413 multiple-choice and 768 open-ended questions. Distinct from existing video benchmarks, LEMON features: (1) semantic richness and disciplinary density, (2) tightly coupled video-audio-text modalities, (3) explicit temporal and pedagogical structure, and (4) contextually linked multi-turn questioning. It further encompasses six major tasks and twelve subtasks, covering the full cognitive spectrum from perception to reasoning and then to generation. Comprehensive experiments reveal substantial performance gaps across tasks, highlighting that even state-of-the-art MLLMs like GPT-4o struggle with temporal reasoning and instructional prediction. We expect LEMON to serve as an extensible and challenging benchmark for advancing multimodal perception, reasoning, and generation in long-form instructional contents.
\end{abstract}

\begin{figure}
    \centering
    \includegraphics[width=1\linewidth]{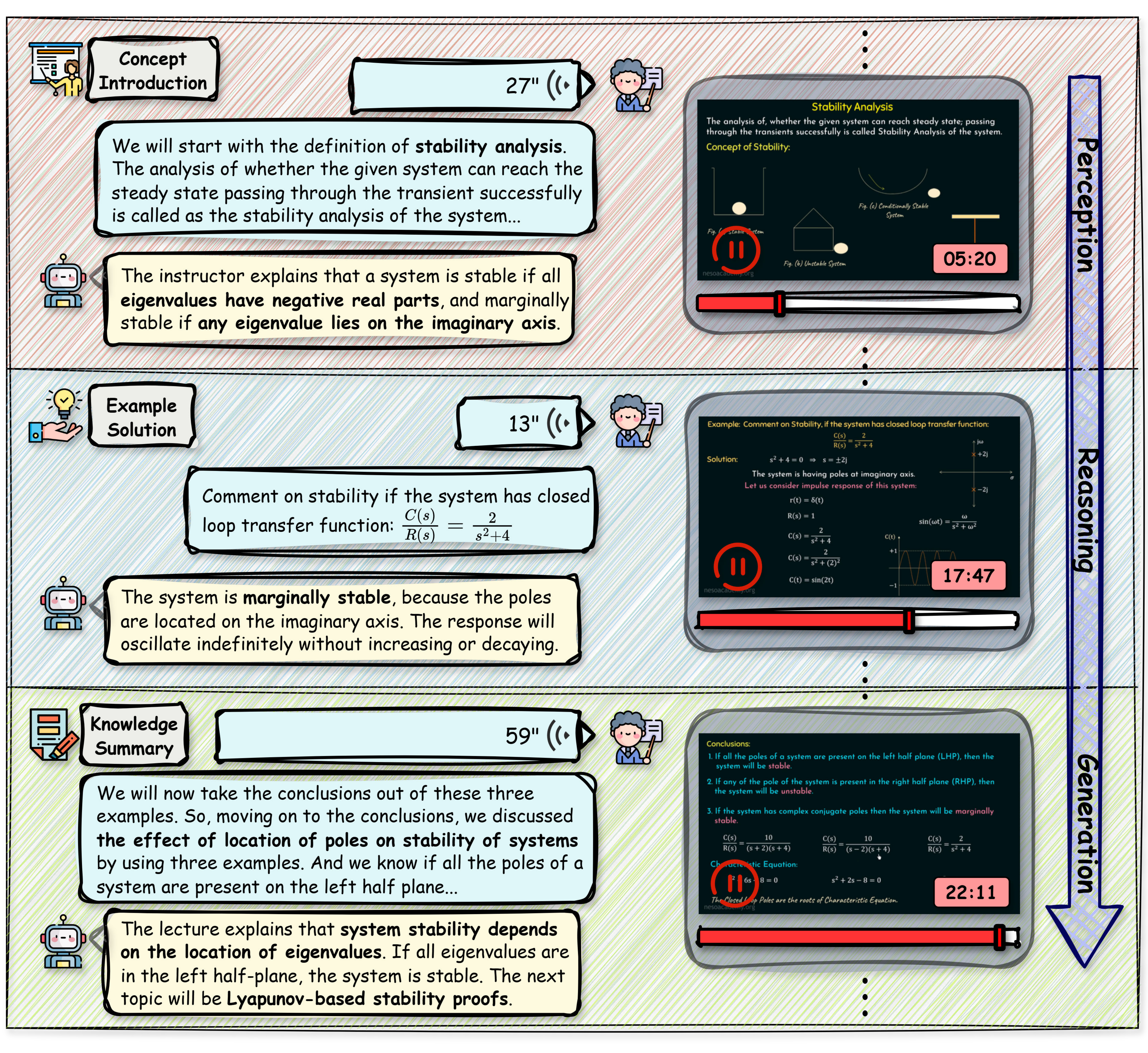}
    \caption{Characteristics of online lecture videos that motivate the design of LEMON. The teaching process naturally integrates multimodal cues (visual slides, speech, and text), covering from \textbf{\textit{Perception}} (concept introduction) to \textbf{\textit{Reasoning}} (example solution) and finally to \textbf{\textit{Generation}} (knowledge summary).}
    \label{fig: OL Charactaristic}
\end{figure}

\begin{table*}[t]
    \centering
    \setlength{\tabcolsep}{4pt}
    \resizebox{1.0\linewidth}{!}{
    \begin{tabular}{lcccccccccccccc}
        \toprule
        \midrule
        \multirow{2}{*}{\textbf{Benchmark}} & 
        \multirow{2}{*}{\textbf{\#Videos}} & 
        \multirow{2}{*}{\textbf{\#QA Pairs}} & 
        \multirow{2}{*}{\textbf{Len. (s)}} & 
        \multicolumn{3}{c}{\textbf{Modality}} & 
        \multicolumn{2}{c}{\textbf{Annotation}} &
        \multicolumn{2}{c}{\textbf{QA Type}} &
        \multicolumn{4}{c}{\textbf{Benchmark Characteristics}} \\
        \cmidrule(lr){5-7} \cmidrule(lr){8-9} \cmidrule(lr){10-11} \cmidrule(lr){12-15}
        &  &  &  & \textbf{Video} & \textbf{Audio} & \textbf{Subtitle} & \textbf{Auto.} & \textbf{Man.} & \textbf{Close-End} & \textbf{Open-End} & \textbf{Multi-Turn} & \textbf{Multi-Level} & \textbf{Causality} & \textbf{Hierarchy}  \\
        
        \midrule
        MSRVTT-QA \cite{xu2017video} & 2,990 & 72,821 & 15.2 & \textcolor{teal}{\faCheck} & \textcolor{Red}{\faTimes} & \textcolor{Red}{\faTimes} & \textcolor{teal}{\faCheck} & \textcolor{Red}{\faTimes} & \textcolor{teal}{\faCheck} & \textcolor{Red}{\faTimes} & \textcolor{Red}{\faTimes} & \textcolor{Red}{\faTimes} & \textcolor{Red}{\faTimes} & \textcolor{Red}{\faTimes} \\
        TGIF-QA \cite{jang2017tgif} & 9,575 & 8,506 & 3.0 & \textcolor{teal}{\faCheck} & \textcolor{Red}{\faTimes} & \textcolor{Red}{\faTimes} & \textcolor{teal}{\faCheck} & \textcolor{teal}{\faCheck} & \textcolor{teal}{\faCheck} & \textcolor{Red}{\faTimes} & \textcolor{Red}{\faTimes} & \textcolor{Red}{\faTimes} & \textcolor{Red}{\faTimes} & \textcolor{Red}{\faTimes} \\
        How2QA \cite{li2020hero} & 1,166 & 2,852 & 15.3 & \textcolor{teal}{\faCheck} & \textcolor{Red}{\faTimes} & \textcolor{teal}{\faCheck} & \textcolor{Red}{\faTimes} & \textcolor{teal}{\faCheck} & \textcolor{teal}{\faCheck} & \textcolor{Red}{\faTimes} & \textcolor{Red}{\faTimes} & \textcolor{Red}{\faTimes} & \textcolor{Red}{\faTimes} & \textcolor{Red}{\faTimes} \\
        NExT-QA \cite{xiao2021next} & 1,000 & 8,564 & 39.5 & \textcolor{teal}{\faCheck} & \textcolor{Red}{\faTimes} & \textcolor{Red}{\faTimes} & \textcolor{teal}{\faCheck} & \textcolor{Red}{\faTimes} & \textcolor{teal}{\faCheck} & \textcolor{teal}{\faCheck} & \textcolor{Red}{\faTimes} & \textcolor{Red}{\faTimes} & \textcolor{teal}{\faCheck} & \textcolor{teal}{\faCheck} \\
        Video-Bench \cite{ning2023video} & 5,917 & 17,036 & 56.0 & \textcolor{teal}{\faCheck} & \textcolor{teal}{\faCheck} & \textcolor{Red}{\faTimes} & \textcolor{teal}{\faCheck} & \textcolor{teal}{\faCheck} & \textcolor{teal}{\faCheck} & \textcolor{Red}{\faTimes} & \textcolor{Red}{\faTimes} & \textcolor{Red}{\faTimes} & \textcolor{Red}{\faTimes} & \textcolor{teal}{\faCheck} \\
        MVBench \cite{li2024mvbench} & 3,641 & 4,000 & 16.0 & \textcolor{teal}{\faCheck} & \textcolor{Red}{\faTimes} & \textcolor{Red}{\faTimes} & \textcolor{teal}{\faCheck} & \textcolor{Red}{\faTimes} & \textcolor{teal}{\faCheck} & \textcolor{Red}{\faTimes} & \textcolor{Red}{\faTimes} & \textcolor{Red}{\faTimes} & \textcolor{teal}{\faCheck} & \textcolor{teal}{\faCheck} \\
        
        \midrule
        EgoSchema \cite{mangalam2023egoschema} & 5,063 & 5,063 & 180.0 & \textcolor{teal}{\faCheck} & \textcolor{Red}{\faTimes} & \textcolor{Red}{\faTimes} & \textcolor{teal}{\faCheck} & \textcolor{teal}{\faCheck} & \textcolor{teal}{\faCheck} & \textcolor{Red}{\faTimes} & \textcolor{Red}{\faTimes} & \textcolor{Red}{\faTimes} & \textcolor{teal}{\faCheck} & \textcolor{Red}{\faTimes} \\
        MovieChat-1K \cite{song2024moviechat} & 130 & 1,950 & 500 & \textcolor{teal}{\faCheck} & \textcolor{Red}{\faTimes} & \textcolor{Red}{\faTimes} & \textcolor{Red}{\faTimes} & \textcolor{teal}{\faCheck} & \textcolor{teal}{\faCheck} & \textcolor{teal}{\faCheck} & \textcolor{Red}{\faTimes} & \textcolor{Red}{\faTimes} & \textcolor{teal}{\faCheck} & \textcolor{Red}{\faTimes}\\
        VideoMME \cite{fu2025video} & 900 & 2,700 & 1017.9 & \textcolor{teal}{\faCheck} & \textcolor{teal}{\faCheck} & \textcolor{teal}{\faCheck} & \textcolor{Red}{\faTimes} & \textcolor{teal}{\faCheck} & \textcolor{teal}{\faCheck} & \textcolor{Red}{\faTimes} & \textcolor{Red}{\faTimes} & \textcolor{teal}{\faCheck} & \textcolor{teal}{\faCheck} & \textcolor{teal}{\faCheck} \\ 
        LongVideoBench \cite{wu2024longvideobench} & 3,763 & 6,678 & 473.0 & \textcolor{teal}{\faCheck} & \textcolor{Red}{\faTimes} & \textcolor{teal}{\faCheck} & \textcolor{Red}{\faTimes} & \textcolor{teal}{\faCheck} & \textcolor{teal}{\faCheck} & \textcolor{Red}{\faTimes} & \textcolor{Red}{\faTimes} & \textcolor{teal}{\faCheck} & \textcolor{Red}{\faTimes} & \textcolor{teal}{\faCheck} \\
        MLVU \cite{zhou2024mlvu} & 1,730 & 3,102 & 930 & \textcolor{teal}{\faCheck} & \textcolor{Red}{\faTimes} & \textcolor{Red}{\faTimes} & \textcolor{teal}{\faCheck} & \textcolor{teal}{\faCheck} & \textcolor{teal}{\faCheck} & \textcolor{teal}{\faCheck} & \textcolor{Red}{\faTimes} & \textcolor{teal}{\faCheck} & \textcolor{teal}{\faCheck} & \textcolor{teal}{\faCheck} \\
        StreamingBench \cite{lin2024streamingbench} & 900 & 4,500 & 243.1 & \textcolor{teal}{\faCheck} & \textcolor{teal}{\faCheck} & \textcolor{Red}{\faTimes} & \textcolor{teal}{\faCheck} & \textcolor{teal}{\faCheck} & \textcolor{teal}{\faCheck} & \textcolor{Red}{\faTimes} & \textcolor{teal}{\faCheck} & \textcolor{teal}{\faCheck} & \textcolor{teal}{\faCheck} & \textcolor{teal}{\faCheck} \\
        OvO-Bench \cite{niu2025ovo} & 644 & 2,814 & 428.9 & \textcolor{teal}{\faCheck} & \textcolor{Red}{\faTimes} & \textcolor{Red}{\faTimes} & \textcolor{teal}{\faCheck} & \textcolor{teal}{\faCheck} & \textcolor{teal}{\faCheck} & \textcolor{Red}{\faTimes} & \textcolor{teal}{\faCheck} & \textcolor{teal}{\faCheck} & \textcolor{teal}{\faCheck} & \textcolor{teal}{\faCheck} \\
        OmniMMI \cite{wang2025omnimmi} & 1,121 & 2,290 & 324.3 & \textcolor{teal}{\faCheck} & \textcolor{teal}{\faCheck} & \textcolor{Red}{\faTimes} & \textcolor{Red}{\faTimes} & \textcolor{teal}{\faCheck} & \textcolor{teal}{\faCheck} & \textcolor{Red}{\faTimes} & \textcolor{teal}{\faCheck} & \textcolor{teal}{\faCheck} & \textcolor{teal}{\faCheck} & \textcolor{teal}{\faCheck} \\
        WorldSense \cite{hong2025worldsense} & 1,662 & 3,172 & 141.1 & \textcolor{teal}{\faCheck} & \textcolor{teal}{\faCheck} & \textcolor{teal}{\faCheck} & \textcolor{Red}{\faTimes} & \textcolor{teal}{\faCheck} & \textcolor{teal}{\faCheck} & \textcolor{Red}{\faTimes} & \textcolor{Red}{\faTimes} & \textcolor{teal}{\faCheck} & \textcolor{teal}{\faCheck} & \textcolor{teal}{\faCheck} \\
        \midrule
        \textbf{LEMON (Ous)} & 2,277 & 4,181 & 196.1 & \textcolor{teal}{\faCheck} & \textcolor{teal}{\faCheck} & \textcolor{teal}{\faCheck} & \textcolor{teal}{\faCheck} & \textcolor{teal}{\faCheck} & \textcolor{teal}{\faCheck} & \textcolor{teal}{\faCheck} & \textcolor{teal}{\faCheck} & \textcolor{teal}{\faCheck} & \textcolor{teal}{\faCheck} & \textcolor{teal}{\faCheck}\\
        \midrule
        \bottomrule
    \end{tabular}
    }
    \caption{Comparison between LEMON and existing benchmarks, across several key aspects: the numbel of videos (\textbf{\#Videos}), number of QA pairs (\textbf{\#QA Pairs}), average video length (\textbf{Len.}), the input modality (\textbf{Video}, \textbf{Audio} and \textbf{Subtitle}), the method of annotation (\textbf{Auto.}/\textbf{Man.} for automatic/manual), the type of question (\textbf{Close-End} or \textbf{Open-End}). \textbf{Multi-Turn} denotes multi-round dialogue format. \textbf{Multi-Level} covers multiple duration levels. \textbf{Causality} reflects temporal causal understanding. \textbf{Hierarchy} indicates structured task organization.}
    \label{tab:benchmark comparison}
\end{table*}



\section{Introduction}
\label{sec:intro}
The rapid development of multimodal large language models (MLLMs) \cite{yin2024survey} have demonstrated remarkable capabilities across vision, language and audio understanding \cite{zhang2021vinvl, lin2023video, chu2024qwen2, yang2025qwen3}, achieving impressive scores on exisiting multimodal benchmarks \cite{chaoyou2023mme, lu2023mathvista, yu2023mm, fang2024mmbench, li2024mvbench, fu2025video}. Models such as GPT-4o \cite{hurst2024gpt} and Gemini 2.5 Pro \cite{comanici2025gemini} have extended the frontier of multimodal reasoning by enabling unified perception and generation. With the evolution from static multimodal alignment to dynamic spatiotemporal interaction, the ability to multimodal understanding, temporal reasoning, and cross-linguistic generation over sequential video data has become a critical milestone for the next generation of MLLMs \cite{lin2024streamingbench, huang2024online, niu2025ovo}. 

Despite their impressive generalization, video understanding remains one of the most demanding frontiers for MLLMs, as it requires temporal perception, cross-modal grounding, and causal reasoning over sequential inputs \cite{lin2024streamingbench, wang2025proactivevideoqa}. However, current benchmarks still fall short of reflecting the complexity of real-world interactive scenarios \cite{niu2025ovo}. Early video understanding benchmarks \cite{caba2015activitynet, xu2016msr, jang2017tgif, xu2017video, yu2019activitynet} typically focused on short clips and simple recognition-based QA, while later efforts expanded video duration and domain diversity to probe temporal reasoning and multimodal comprehension \cite{mangalam2023egoschema, li2024mvbench, li2024vitatecs, liu2024tempcompass, zhou2024mlvu, fu2025video}. Nevertheless, most benchmarks  remain constrained by visual streams without corresponding audio, overlooking the inherently multimodal nature of real-world video understanding \cite{wang2025omnimmi, hong2025worldsense}. Moreover, most existing benchmarks primarily focus on open-domain content such as entertainment, sports, and movies \cite{li2024omnibench, yao2025timechat, tang2025video}, leaving the educational domain largely unexplored. Therefore, an essential question arises: \textit{How well do MLLMs preform temporal multimodal understanding on instructional videos?}

To address these limitations, we introduce \textbf{LEMON}, a new benchmark named \textbf{\textit{L}}ecture-based \textbf{\textit{E}}valuation for \textbf{\textit{M}}ultim\textbf{\textit{O}}dal u\textbf{\textit{N}}derstanding, designed to evaluate MLLMs' temporal multimodal understanding on instructional videos. In contrast to existing benchmark that rely on fragmented, visually dominated clips \cite{lei2018tvqa, li2020hero}, LEMON focuses on online lecture videos, a domain that naturally integrates visual, auditory, and textual modalities in a temporally coherent and semantically dense manner \cite{zhang20252}. The benchmark is built upon videos drawn from STEM disciplines, emcompassing \textbf{Mathematics}, \textbf{Artificial Intelligence}, \textbf{Computer Science}, \textbf{Electronic Engineering}, and \textbf{Robotics}, spanning 29 distinct courses. Unlike other open-domain content, online lectures embody structured knowledge delivery, pedagogical intent, and cross-modal synchronization, where instructors continuously switch between spoken explanations, written derivations, and visual demonstrations, as shown in Figure \ref{fig: OL Charactaristic}. These tightly coupled modalities form rich temporal dependencies that demand both fine-grained perception and long-term causal reasoning, making online education videos an ideal testbed for evaluating MLLMs’ ability to perform streaming understanding and multimodal reasoning in complex, knowledge intensive environments.

Table \ref{tab:benchmark comparison} shows that LEMON provides a more comprehensive evaluation for video understanding compared to existing benchmarks, comprising 2,277 lecture videos and 4,181 QA pairs. Each lecture segment is accompanied by synchronized video, audio and subtitle streams, which provides rich multimodal cues for evaluating temporal understanding. Beyond scale and diversity, LEMON distinguishes itself through three unique characteristics:
\begin{itemize}
    \item \textbf{Temporally Causal Educational Scenarios.} LEMON focuses on educational content, where multimodal information unfolds in a strongly causal and temporally continuous manner. This enables a rigorous evaluation of models’ streaming perception and long-term reasoning capabilities, reflecting real-world temporal dependencies.
    \item \textbf{Cognitively Hierarchical Multimodal Design.} LEMON organizes audio, video, and subtitles into a coherent multimodal hierarchy, encompassing six major categories and twelve subtasks that span the cognitive spectrum from \textbf{\textit{Perception}} to \textbf{\textit{Reasoning}} and then to \textbf{\textit{Generation}}, thereby linking low-level multimodal alignment with high-level semantic understanding.
    \item \textbf{Interactive Multi-turn Evaluation Format.} LEMON adopts a multi-turn dialogue setting, where interrelated questions require MLLMs to track contexts, integrate modalities, and perform temporal reasoning across turns. This design encourages a more realistic assessment of how MLLMs comprehend, predict, and generate within dynamic, multimodal streams.
\end{itemize}

Finally, we conduct extensive evaluations across a board spectrum of MLLMs, including open-source omni models, open-source video models and proprietary models. Experimental results reveal that while current MLLMs demonstrate promising multimodal comprehension abilities like visual perception, they still struggle with temporal coherence and causal reasoning over continuous instructional sequences, underscoring a clear gap between existing capabilities and real-world multimodal comprehension. 

\begin{figure*}[t]
    \centering
    \includegraphics[width=1.0\linewidth]{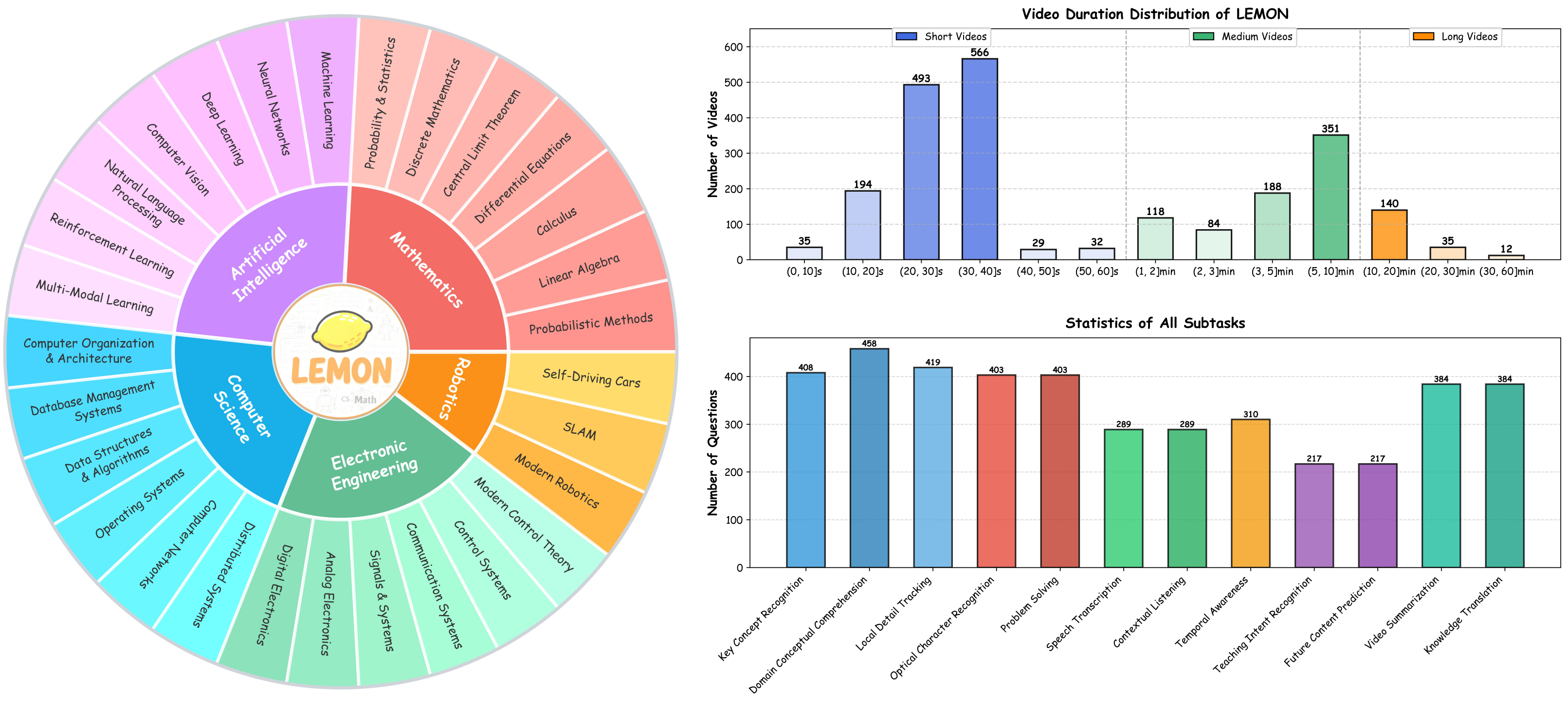}
    \caption{\textbf{Statistics overview of LEMON.} Left: Video categories included in LEMON. Top Right: Video duration distribution of LEMON. Bottom Right: Statistics on the number of all subtasks.}
    \label{fig: statistic overview}
\end{figure*}

\section{Related Work}
\textbf{Multimodal Large Language Models.} Recent advances in MLLMs have substantially expanded traditional LLMs toward visual, auditory and video understanding \cite{chen2023vlp, jiang2024effectiveness, wang2023large, yin2024survey, chu2024qwen2, yao2024minicpm, bai2025qwen2}. Early video MLLMs such as VideoChat \cite{li2023videochat}, Video-LLaMA \cite{zhang2023video}, and Video-LLaVA \cite{lin2023video} employed attention-based fusion mechanisms to process short video content. Subsequent models, including MovieChat \cite{song2024moviechat} and ShareGPT4Video \cite{chen2024sharegpt4video}, introduced memory modules to enable recursive video reasoning across extended sequences, while models like LongVA \cite{zhang2024long}, LongLLaVA \cite{wang2024longllava} and LongVU \cite{shen2024longvu} further extended contextual length, enabling more effective processing of long video sequences. Beyond offline settings, online MLLMs such as Flash-VStream \cite{zhang2024flash}, VideoLLM-Online \cite{chen2024videollm}, and VideoChat-Online \cite{huang2025online} enable incremental reasoning and real-time response as frames stream in, simulating human-like perception. Meanwhile, Omni MLLMs such as Mini-Omni2 \cite{xie2024mini}, VITA-1.5 \cite{fu2025vita} and Qwen3-Omni \cite{xu2025qwen3} pursue modality unification by integrating text, image, audio and video within shared tokenization and unified attention. Recent works, like IXC2.5-OL \cite{zhang2024internlm} and MiniCPM-o 2.6 \cite{openbmb2025minicpmo}, support streaming and real-time interaction, bridging perception and cognition in a continuous omnimodal input flow. Despite these advances, it remains an open challenge for MLLMs to reason over long temporal spans, sustain dialogue coherence, and adapt to specialized educational domains.

\noindent
\textbf{Video Understanding Benchmarks.} With the rapid development of MLLMs, video understanding has become a crucial platform for assessing their perception and reasoning capabilities \cite{zhang2024long, wang2024videoagent, xu2023retrieval}. Early benchmarks, such as MSRVTT-QA \cite{xu2017video} and TGIF-QA \cite{jang2017tgif}, primarily focus on short video clips with simple question-answer pairs. As MLLMs demonstrated stronger cross-modal reasoning and memory abilities, research gradually shifted toward long-form and semantically complex video comprehension, exemplified by datasets like VideoMME \cite{fu2025video} and MLVU \cite{zhou2024mlvu}, which emphasize temporal grounding, causal reasoning, and multi-event inference. Meanwhile, evaluation tasks have expanded from simple classification \cite{kay2017kinetics} and action recognition \cite{wang2023paxion, wu2024star, li2024mvbench} to temporal reasoning\cite{fu2025video, wu2024longvideobench, wang2025omnimmi, hong2025worldsense}, event prediction \cite{niu2025ovo, lin2024streamingbench} and video captioning \cite{zhou2024mlvu, chen2024sharegpt4video}. More recently, the emergence of online or streaming benchmarks, such as VStream-QA \cite{zhang2024flash} and StreamingBench \cite{lin2024streamingbench} has further simulated real-time perception by requiring temporally causal processing of sequential frames rather than post-hoc analysis \cite{niu2025ovo, wang2025omnimmi}. This paradigm not only challenges models to maintain temporal coherence and long-term memory but also better mirrors how humans perceive and reason over continuous visual streams, promising for driving MLLMs toward more realistic and temporally grounded video understanding.

 
\section{LEMON}

\subsection{Data Construction} \label{sec: data construction}
\textbf{Video Collection.} We choose online lecture videos as the basis of our benchmark as they inherently exhibit streaming characteristics: sequential information flow, strong temporal causality, rich multimodal interactions, and sustained discourse continuity. Unlike videos from other domains, lectures require long-term reasoning and cross-modal alignment, making them ideal for evaluating real-world streaming understanding. As shown in the left part of Figure \ref{fig: statistic overview}, we collect videos across five STEM disciplines: \textbf{Mathematics}, \textbf{Artificial Intelligence}, \textbf{Computer Science}, \textbf{Electronic Engineering}, and \textbf{Robotics}, encompassing 29 distinct courses from YouTube \footnote{\url{https://www.youtube.com}}. Each video is carefully selected to ensure clear audio-visual quality, high resolution, presence of explicit instructional or explanatory content, and coherent multimodal cues that facilitate question generation. For each video, we extracted the audio track using FFmpeg \footnote{\url{https://www.ffmpeg.org}} and generated high-quality subtitles via Whisper-v3 \footnote{\url{https://huggingface.co/openai/whisper-large-v3}}, ensuring precise synchronization among modalities for subsequent QA construction.

\noindent
\textbf{QA Generation and Annotation.} Building upon the collected videos, we employ a hybrid human-AI pipeline to construct high-quality QA pairs, with detailed procedures and examples provided in the supplementary material. Specifically, each lecture video is segmented into 10-minute clips, and their subtitles are input to GPT-4o \cite{hurst2024gpt} to automatically generate initial questions with corresponding temporal spans. For each referenced spans, we extract key video frames and provide them, together with the generated questions, to Gemini 2.0 Flash \cite{google2024gemini2}, which produced preliminary answer options or labels. Importantly, to promote contextual reasoning, each subtask contains sequentially dependent questions, with subsequent questions built upon previous ones through cues like ``\textit{Based on your answer to the previous question}''. Subsequently, several human annotators, guided by task-specific prompts, carefully review each entire video, verifying and refining the generated questions and answers to ensure that each question is answerable through combining video with audio or video with subtitles. Finally, we obtain a total of 4,181 QA pairs spanning six major categories and twelve subtask, as illustrated in the bottom right of Figure \ref{fig: statistic overview}.

\noindent
\textbf{Quality Review.} To ensure the reliability of the constructed QA pairs, we conduct a multi-stage quality review. After the initial human annotation, independent reviewers rechecked each question, verifying the correctness and coherence of the question, answer options, and temporal references. Subsequently, we employ the Qwen2.5-Omni \cite{xu2025qwen2} model to re-evaluate all QA pairs, with and without audio or subtitles, assessing their true multimodal dependency. QA pairs whose correctness remained unaffected by the removal of a modality were further revised or discarded to ensure that each question genuinely required multimodal comprehension. Based on these results, further manual adjustments are made to refine the final dataset, where more detailed descriptions, examples, and implementation specifics are provided in the supplementary material.

\begin{figure*}[t]
    \centering
    \includegraphics[width=1\linewidth]{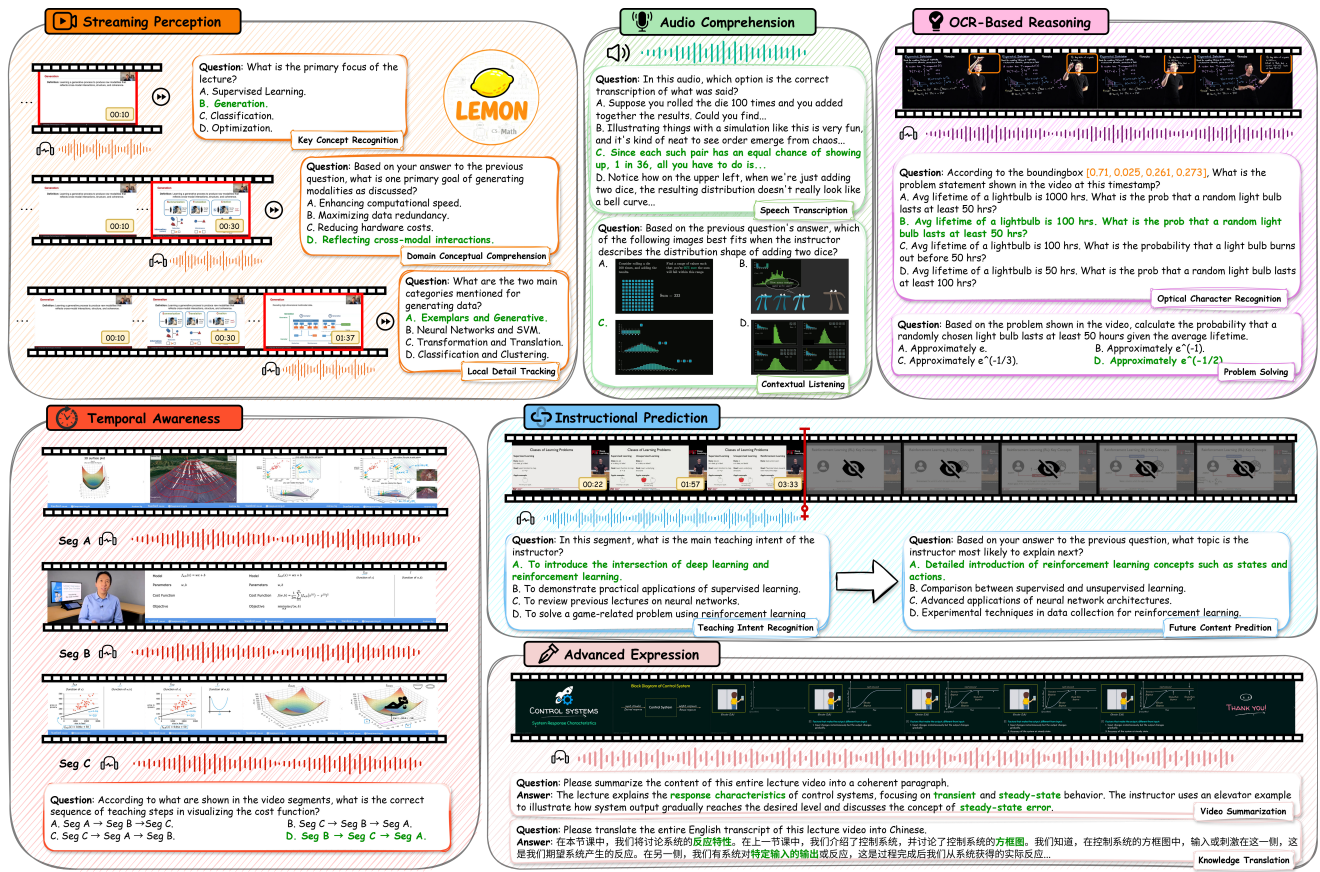}
    \caption{\textbf{Examples of LEMON.} Each multiple-choice task forms a chain of interdependent subtasks, where each answer informs the next. The open-ended task assesses a model’s ability to produce coherent summaries and cross-lingual translations.}
    \label{fig: task examples}
\end{figure*}

\begin{table*}[t]
    \centering
    \setlength{\tabcolsep}{4pt}
    \resizebox{1.0\linewidth}{!}{
    \begin{tabular}{l c c cccc ccc ccc c ccc ccc c}
        \toprule
        \midrule
        
        \multirow{2}{*}{\textbf{Models}} & 
        \multirow{2}{*}{\textbf{\#Params}} & 
        \multirow{2}{*}{\textbf{\#Frames}} & 
        \multicolumn{4}{c}{\textbf{SP}} &
        \multicolumn{3}{c}{\textbf{OR}} &
        \multicolumn{3}{c}{\textbf{AC}} &
        \multirow{2}{*}{\textbf{TA}} &
        \multicolumn{3}{c}{\textbf{IP}} &
        \multicolumn{3}{c}{\textbf{AE}} &
        \multirow{2}{*}{\textbf{Overall}} \\
        
        \cmidrule(lr){4-7} \cmidrule(lr){8-10} \cmidrule(lr){11-13} \cmidrule(lr){15-17} \cmidrule(lr){18-20}

        & & & KCR & DCC & LDT & Avg. & OCR & PS & Avg. & ST & CL & Avg. & & TIR & FCP & Avg. & VS & KT & Avg. & \\ \midrule

        \rowcolor{gray!15}
        \multicolumn{21}{c}{\textbf{\textit{Proprietary MLLMs}}} \\ \midrule
        GPT-4o$^{*}$ \cite{hurst2024gpt} & - & 64 & 88.25 & 86.49 & 78.89 & 84.54 & 93.55 & 60.79 & 77.17 & - & \textcolor{red!60!black}{\textbf{85.61}}$^{\dagger}$ & \textcolor{red!60!black}{\textbf{85.61}} & 29.03 & 80.78 & \textcolor{red!60!black}{\textbf{75.58}} & 78.18 & 61.08 & 39.31 & 50.19 & 67.45 \\
        GPT-5$^{*}$ \cite{openai2025gpt5} & - & 64 & 86.51 & 87.67 & 83.60 & 85.93 & \textcolor{red!60!black}{\textbf{97.27}} & 84.35 & 90.81 & - & 83.71$^{\dagger}$ & 83.71 & 23.71 & 84.01 & 64.98 & 74.49 & 57.05 & \textcolor{red!60!black}{\textbf{60.46}} & 58.76 & 69.57\\
        Gemini 2.5 Pro$^{*}$ \cite{comanici2025gemini} & - & 64 / 1fps & 88.25 & \textcolor{red!60!black}{\textbf{88.25}} & \textcolor{red!60!black}{\textbf{84.15}} & \textcolor{red!60!black}{\textbf{86.88}} & 96.03 & \textcolor{red!60!black}{\textbf{87.75}} & \textcolor{red!60!black}{\textbf{91.89}} & 31.82 & 79.92 & 55.87 & 22.58 & 81.71 & 64.98 & 73.35 & 56.45 & 59.07 & 57.76 & 64.72 \\
        Gemini 2.5 Flash$^{*}$ \cite{comanici2025gemini} & - & 64 / 1fps & \textcolor{red!60!black}{\textbf{89.42}} & 87.67 & 83.02 & 86.70 & 94.54 & 80.94 & 87.73 & 31.44 & 79.92 & 55.68 & 23.87 & 82.44 & 67.25 & 74.85 & 20.48 & 27.31 & 23.90 & 58.79\\
        Claude 4.5 Sonnet$^{*}$ \cite{Anthropic2025ClaudeSonnet45} & - & 64 & 84.47 & 82.98 & 78.30 & 81.91 & 91.56 & 61.61 & 76.58 & - & 78.79 & 78.79 & 25.48 & 83.09 & 72.81 & 77.95 & 61.18 & 59.80 & 60.49 & 66.87 \\ 
        Grok 4$^{*}$ \cite{grok4_xai_2025} & - & 64 & 84.77 & 83.60 & 77.79 & 82.05 & 85.61 & 52.57 & 69.09 & 34.09 & 74.62 & 54.36 & 21.94 & 82.63 & 66.82 & 74.72 & \textcolor{red!60!black}{\textbf{62.76}} & 58.38 & \textcolor{red!60!black}{\textbf{60.57}} & 60.46 \\
        \midrule

        \rowcolor{gray!15}
        \multicolumn{21}{c}{\textbf{\textit{Open-Source Omni MLLMs}}} \\ \midrule
        Qwen3-Omni$^{*}$ \cite{xu2025qwen3} & 30B-A3B & 64 / 1fps & 85.93 & 84.19 & 78.95 & 83.02 & 94.79 & 58.82 & 76.80 & - & 75.38$^{\dagger}$ & 75.38 & 65.81 & 84.91 & 75.00 & \textcolor{red!60!black}{\textbf{79.95}} & 60.85 & 54.07 & 57.46 & \textcolor{red!60!black}{\textbf{73.07}} \\
        IXC2.5-OL \cite{zhang2024internlm} & 7B & Adaptive$^{\ddagger}$ & 82.23 & 77.05 & 80.67 & 79.98 & 75.68 & 36.48 & 56.08 & - & 21.80& 21.80 & 10.00 & 50.46 & 56.25 & 53.36 & 39.59 & 0.48 & 20.04 & 40.21 \\
        Baichuan-Omni-1.5 \cite{li2025baichuan} & 7B & 1fps & 84.17 & 70.83 & 71.67 & 75.56 & 80.89 & 35.24 & 58.07 & 76.47 & 27.34 & 51.90 & \textcolor{red!60!black}{\textbf{66.77}} & 78.02 & 55.76 & 66.89 & 5.64 & 9.10 & 7.37 & 54.43 \\
        MiniCPM-o 2.6 \cite{openbmb2025minicpmo} & 8B & 1fps & 78.68 & 76.79 & 71.13 & 75.53 & 87.10 & 32.75 & 59.93 & 32.18 & 52.94 & 42.56 & 55.81 & 75.54 & 61.45 & 68.49 & 55.95 & 16.97 & 36.46 & 56.46 \\
        Ola \cite{liu2025ola} & 7B & 1fps & 59.25 & 66.50 & 64.43 & 63.39 & 75.39 & 48.41 & 61.82 & 31.49 & 32.18 & 31.83 & 38.06 & 76.18 & 72.35 & 74.26 & 54.31 & 31.36 & 42.83 & 52.05 \\
        M4 \cite{wang2025omnimmi} & 7B & 1fps & 66.68 & 51.48 & 48.37 & 55.51 & 27.05 & 29.78 & 28.42 & 69.44 & 27.34 & 48.35 & 27.42 & 69.72 & 42.40 & 56.06 & 44.85 & 2.22 & 23.54 & 39.88 \\
        video-SALMONN 2+ \cite{tang2025video} & 7B & 64 & 85.65 & 78.04 & 71.52 & 78.41 & 86.60 & 44.17 & 65.39 & \textcolor{red!60!black}{\textbf{80.32}} & 74.39 & 77.34 & 23.55 & 82.63 & 72.35 & 77.49 & 49.03 & 24.45 & 36.74 & 59.82 \\ 
        \midrule
        
        \rowcolor{gray!15}
        \multicolumn{21}{c}{\textbf{\textit{Open-Source Video MLLMs}}} \\ \midrule
        ShareGPT4Video \cite{chen2024sharegpt4video} & 8B & 64 & 72.41 & 62.13 & 65.00 & 66.51 & 47.64 & 32.75 & 40.20 & - & 39.10 & 39.10 & 9.35 & 80.32 & 58.53 & 69.42 & 46.72 & 9.11 & 27.92 & 42.08 \\
        LongVA \cite{zhang2024long} & 7B & 64 & 51.96 & 51.96 & 55.10 & 53.01 & 44.42 & 23.82 & 34.12 & - & 41.18 & 41.18 & 39.35 & 60.05 & 49.31 & 54.68 & 44.64 & 22.51 & 33.58 & 42.65 \\
        LLaVA-OneVision \cite{li2024llava} & 7B & 64 & 83.78 & 71.87 & 69.79 & 75.15 & 88.59 & 35.89 & 62.24 & - & 50.52 & 50.52 & 24.19 & 82.68 & 54.84 & 68.76 & 47.28 & 22.24 & 34.76 & 52.60 \\
        LongVU \cite{shen2024longvu} & 7B & 1fps & 79.12 & 67.72 & 75.49 & 74.11 & 58.81 & 39.45 & 49.13 & - & 39.79 & 39.79 & 25.16 & 78.94 & 53.92 & 66.43 & 24.89 & 0.98 & 12.94 & 44.59 \\
        Dispider \cite{qian2025dispider} & 7B & 100 clips & 63.06 & 55.28 & 53.73 & 57.36 & 69.48 & 33.25 & 51.37 & - & 32.53 & 32.53 & - & 76.18 & 52.07 & 64.13 & - & - & - & 51.34 \\
        VideoLLaMA3 \cite{zhang2025videollama} & 7B & 180 & 78.54 & 82.19 & 82.19 & 80.97 & 62.50 & 25.00 & 43.75 & - & 32.18 & 32.18 & 24.52 & 40.09 & 32.26 & 36.18 & 55.85 & 29.80 & 42.83 & 43.40 \\
        InternVL3 \cite{zhu2025internvl3} & 8B & 64 & 84.82 & 81.71 & 82.23 & 82.92 & 94.29 & 41.94 & 68.12 & - & 64.01 & 64.01 & 23.23 & \textcolor{red!60!black}{\textbf{85.85}} & 62.67 & 74.26 & 54.69 & 29.00 & 41.84 & 59.06 \\
        Video-XL-2 \cite{qin2025video} & 8B & 64 & 77.56 & 72.90 & 75.49 & 75.32 & 86.60 & 42.68 & 64.64 & - & 43.07 & 43.07 & 50.65 & 83.09 & 65.44 & 74.26 & 35.02 & 21.71 & 28.37 & 56.05 \\

        \midrule
        \bottomrule
    \end{tabular}
    }
    \caption{\textbf{The overall performances of MLLMs on LEMON.} ``Avg.'' denotes the average accuracy. ``Overall'' represents the mean accuracy across all six tasks. $*$ indicates models evaluated via their official APIs. In AC task, models without audio support are given the ground-truth ST answers for the CL task. $\dagger$ indicates that the model itself does not support audio but uses dedicated audio models (GPT-4o-audio for GPT-4o and GPT-5, Qwen audio for Qwen 3-Omni) to process ST tasks, and then provides the transcription results to the main model. $\ddagger$ indicates models handle video inputs in an adaptive manner, without a predefined frame sampling rate or count. }
    \label{tab: overall performance}
\end{table*}

\subsection{Dataset Statistics} \label{sec: dataset statistics}
As summarized in Figure \ref{fig: statistic overview}, our proposed LEMON benchmark consists of 2,277 video segment with synchronized audio across 5 disciplines and 29 courses, averaging 196.1 seconds in lengths, and includes 3,413 multiple-choice questions and 768 open-ended questions. Motivated by the versatility for temporal multimodal understanding, we design six core task categories: \textbf{Streaming Perception (SP)}, \textbf{OCR-based Reasoning (OR)}, \textbf{Audio Comprehension (AC)}, \textbf{Temporal Awareness (TA)}, \textbf{Instructional Prediction (IP)}, and \textbf{Advanced Expression (AE)}. Furthermore, LEMON embodies three distinctive aspects: (i) realistic streaming scenarios with multi-turn dialogue; (ii) complete video and audio streams with aligned transcripts or subtitles; and (iii) multi-level cognitive abilities across perception, reasoning, and generation. Specifically, \textit{\textbf{Perception}} includes \textit{Key Concept Recognition}, \textit{Domain Conceptual Comprehension}, \textit{Local Detail Tracking}, \textit{Optical Character Recognition}, \textit{Speech Transcription}, and \textit{Teaching Intent Recognition}; \textit{\textbf{Reasoning}} covers \textit{Problem Solving}, \textit{Temporal Ordering}, \textit{Contextual Listening}, and \textit{Future Content Prediction}; \textit{\textbf{Generation}} consists of \textit{Video Summarization} and \textit{Knowledge Translation}.

\subsection{Benchmark Tasks} \label{sec: benchmark tasks}
The LEMON benchmark evaluates six tasks, each targeting a distinct aspect of temporal multimodal understanding.

\subsubsection{Streaming Perception}
This task assesses the ability of models to perceive and interpret multimodal information in a continuous video stream, emphasizing context accumulation and temporal grounding rather than static observation.
\begin{itemize}
    \item \textbf{Key Concept Recognition (KCR)}: Identify main concepts or ideas within a segment; 
    \item \textbf{Domain Conceptual Comprehension (DCC)}: Connect visual cues to discipline knowledge such as formulas, definitions, or theoretical principles;
    \item \textbf{Local Detail Tracking (LDT)}: Capture fine-grained factual detail, numerical value, or descriptive statement.
\end{itemize}

\subsubsection{OCR-Based Reasoning}
This task measures how well a model can identify text in videos and leverage it for reasoning. While OCR alone tests whether a model can directly recognize text on slides or boards, combining audio information often helps resolve cases where the text is unclear or partially occluded, reflecting real classroom scenarios.
\begin{itemize}
    \item \textbf{Optical Character Recognition (OCR)}: Recognize and extract questions appearing in videos, optionally leveraging audio or subtitles for unclear.
    \item \textbf{Problem Solving (PS)}: Leverage the extracted texts to solve discipline problems or answer questions that require reasoning over the textual content.
\end{itemize}

\subsubsection{Audio Comprehension}
This task focuses on evaluating models to understand and utilize auditory information embedded in lecture videos, which is crucial when visual cues alone are insufficient. It examines both transcription accuracy and contextual reasoning based on spoken content.
\begin{itemize}
    \item \textbf{Speech Transcription (ST)}: Convert spoken language in the video into accurate textual representation, ensuring alignment with visual or contextual cues.
    \item \textbf{Contextual Listening (CL)}: Comprehend semantic meaning and identify corresponding visual content that semantically aligns with the speech segments.
\end{itemize}

\subsubsection{Temporal Awareness}
This task evaluates a model’s ability to capture and reason about temporal dependencies across video segments, reflecting its understanding of causal and sequential relationships within a continuous stream. Models are required to infer the correct chronological or logical order of events, actions, or instructional steps, demonstrating awareness of how visual and auditory information evolves over time.

\subsubsection{Instructional Prediction}
This task focuses on forecasting upcoming instructional events or content based on the current segment of a streaming lecture, which requires the model to infer pedagogical intent and understand why an instructor is explaining a concept and what is likely to cover next.
\begin{itemize}
    \item \textbf{Teaching Intent Recognition (TIR)}: Identify the instructor's underlying teaching objective within the current segment, such as introducing a new concept, emphasizing a key idea, or preparing for a topic transition.
    \item \textbf{Future Content Prediction (FCP)}: Anticipate the next topic, example, or instructional step likely to appear, based on the contextual flow of the ongoing lecture.
\end{itemize}

\subsubsection{Advanced Expression}
This task evaluates a model’s capacity for high-level multimodal generation, requiring it not only to understand and recognize visual content but also to produce coherent, semantically enriched responses while emphasizing highlight discipline-specific concepts and terminology.
\begin{itemize}
    \item \textbf{Video Summarization (VS)}: Generate a coherent and informative summary of the lecture segment, focusing on instructional objectives and core concepts conveyed.
    \item \textbf{Knowledge Translation (KT)}: Translate the instructional content into other target languages, ensuring both semantic fidelity and academic clarity, especially for concepts or terminology.
\end{itemize}

\begin{figure*}[h]
    \centering
    \begin{subfigure}[t]{0.33\linewidth}
        \centering
        \includegraphics[width=0.9\linewidth]{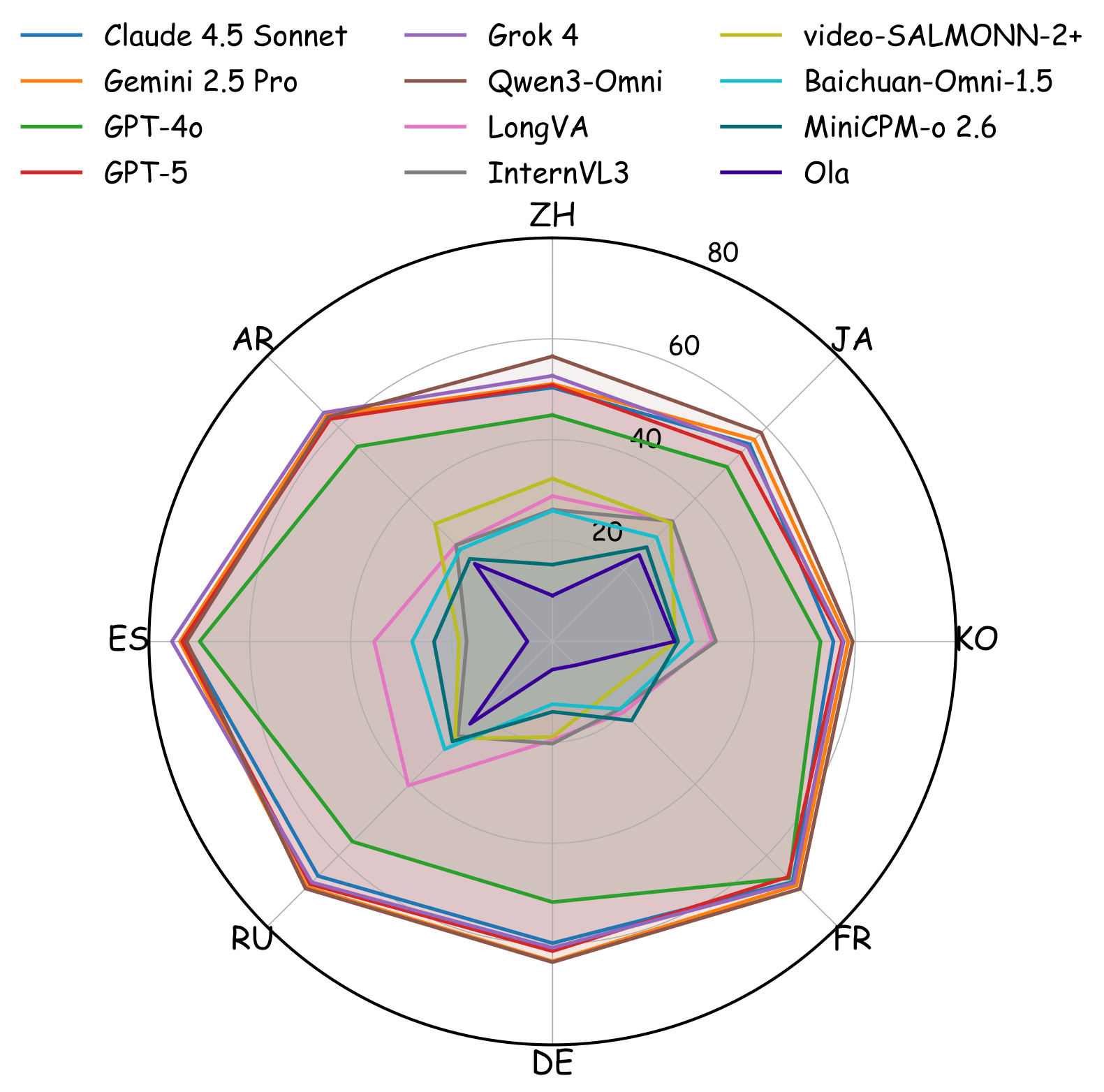}
        \caption{Multilingual performance for MLLMs.}
        \label{subfig: multilingual}
    \end{subfigure}
    \hfill
    \begin{subfigure}[t]{0.66\linewidth}
        \centering
        \includegraphics[width=1\linewidth]{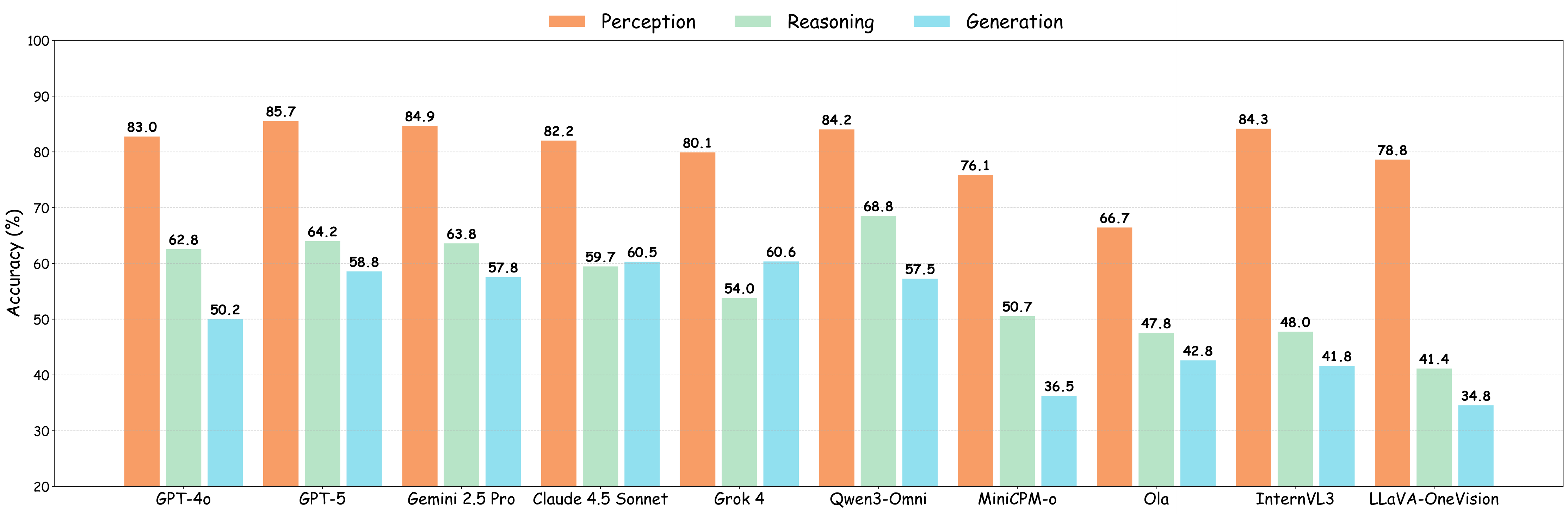}
        \caption{Task performance for MLLMs.}
        \label{subfig: task}
    \end{subfigure}
    \caption{\textbf{Analysis of model performance across languages and tasks.} (a) Average performance on different languages measured by \textit{BLEU} \cite{papineni-etal-2002-bleu}, \textit{ROUGE-L} \cite{lin-2004-rouge}, and \textit{BERTScore} \cite{zhang2019bertscore}. \textbf{ZH}: Chinese, \textbf{JA}: Japanese, \textbf{KO}: Korean, \textbf{FR}: French, \textbf{DE}: German, \textbf{RU}: Russian, \textbf{ES}: Spanish, \textbf{AR}: Arabic. (b) Task performance comparison shows strengths in \textit{\textbf{Perception}} and weaknesses in \textit{\textbf{Reasoning}} and \textit{\textbf{Generation}}.}
    \label{fig: multilingual and task}
\end{figure*}

\section{Experiments}

\subsection{Settings}
We evaluate three categories of MLLMs on the LEMON benchmark, which contains lecture videos with synchronized visual, textual, and auditory modalities. The experimental models include: (i) \textbf{Proprietary MLLMs}, such as GPT-4o \cite{hurst2024gpt} and Gemini 2.5 Pro \cite{comanici2025gemini}, which represent state-of-the-art commercial systems with strong multimodal reasoning capabilities; (ii) \textbf{Open-Source Omni MLLMs}, such as Qwen3-Omni \cite{xu2025qwen3} and MiniCPM-o 2.6 \cite{openbmb2025minicpmo}, designed for unified multimodal interaction across text, image, audio, and video; and (iii) \textbf{Open-Source Video MLLMs}, such as VideoLLaMA3 \cite{zhang2025videollama} and LongVA \cite{geng2025longvale}, specialized for long-range temporal understanding and video-grounded reasoning. All models are assessed under a unified prompting and video-chunking protocol to ensure fair comparison. All evaluations are conducted in a zero-shot setting based on their official implementations or APIs.

\subsection{Main Results}
Table \ref{tab: overall performance} presents the overall performance of 21 representative MLLMs evaluated on the LEMON benchmark. Our evaluation brings several important findings as follows.

\textbf{Proprietary MLLMs demonstrate stronger performance on most LEMON tasks, significantly outperforming open-source counterparts.} Proprietary models like GPT-5 and Gemini 2.5 Pro consistently lead across most tasks, demonstrating strong multimodal understanding and reasoning. These findings highlight a persistent gap in multimodal alignment and reasoning robustness between proprietary and open-source models.

\begin{table}[t]
    \centering
    \resizebox{1.0\linewidth}{!}{
    \begin{tabular}{l ccc}
        \toprule
        \midrule
        
        \multirow{2}{*}{\textbf{Models}} & 
        \multicolumn{3}{c}{\textbf{Overall}} \\

        \cmidrule(lr){2-4}

        & Video & +Subtitle & +Audio \\ \midrule
        Gemini 2.5 Pro & 82.74 & \textbf{84.53}$_{\textcolor{teal}{\uparrow 1.79}}$ & \textbf{84.04}$_{\textcolor{teal}{\uparrow 1.30}}$ \\
        Gemini 2.5 Flash & \textbf{82.81} & 84.44$_{\textcolor{teal}{\uparrow 1.63}}$ & 83.09$_{\textcolor{teal}{\uparrow 0.28}}$ \\
        Grok 4 & 75.02 & 78.53$_{\textcolor{teal}{\uparrow 3.51}}$ & 75.29$_{\textcolor{teal}{\uparrow 0.27}}$ \\
        Qwen3-Omni$^{\triangledown}$ & 77.46 & 79.92$_{\textcolor{teal}{\uparrow 2.46}}$ & - \\
        IXC2.5-OL$^{\triangledown}$ & 44.00 & 63.14$_{\textcolor{teal}{\uparrow 19.14}}$ & - \\
        Baichuan-Omni-1.5 & 57.78 & 71.64$_{\textcolor{teal}{\uparrow 13.96}}$ & 66.84$_{\textcolor{teal}{\uparrow 9.06}}$ \\
        MiniCPM-o 2.6$^{\triangledown}$ & 63.58 & 67.98$_{\textcolor{teal}{\uparrow 4.40}}$ & - \\
        Ola & 68.62 & 80.00$_{\textcolor{teal}{\uparrow 11.38}}$ & 66.49$_{\textcolor{Red}{\downarrow 2.13}}$ \\
        M4 & 47.68 & 50.66$_{\textcolor{teal}{\uparrow 2.98}}$ & 46.66$_{\textcolor{Red}{\downarrow 1.02}}$ \\
        video-SALMONN 2+ & 69.83 & 73.76$_{\textcolor{teal}{\uparrow 3.93}}$ & 73.76$_{\textcolor{teal}{\uparrow 3.93}}$ \\

        \midrule
        \bottomrule
    \end{tabular}
    }
    \caption{\textbf{Impact of audio information for Omni MLLMs on SP, OR and IP tasks.} We evaluate three input settings: video-only, video with subtitles, and video with audios. $\triangledown$ indicates models that do not support simultaneous input of video and audio. }
    \label{tab: audio impact on omni mllms}
\end{table}

\begin{table}[t]
    \centering 
    \begin{tabular}{l cc}
        \toprule
        \midrule
        \multirow{2}{*}{\textbf{Models}} & 
        \multicolumn{2}{c}{\textbf{Overall}} \\

        \cmidrule(lr){2-3}

        & Video & +Subtitle \\ \midrule
        GPT-4o & \textbf{77.21} & \textbf{79.96}$_{\textcolor{teal}{\uparrow 2.75}}$ \\
        Claude 4.5 Sonnet & 73.93 & 78.81$_{\textcolor{teal}{\uparrow 4.88}}$ \\
        ShareGPT4Video & 34.46 & 58.71$_{\textcolor{teal}{\uparrow 24.25}}$ \\
        LongVA & 36.93 & 47.27$_{\textcolor{teal}{\uparrow 10.34}}$ \\
        LLaVA-OneVision & 63.66 & 68.72$_{\textcolor{teal}{\uparrow 5.06}}$ \\
        LongVU & 56.38 & 63.22$_{\textcolor{teal}{\uparrow 16.84}}$ \\
        Dispider & 57.07 & 57.62$_{\textcolor{teal}{\uparrow 0.55}}$ \\
        VideoLLaMA3 & 68.31 & 53.63$_{\textcolor{Red}{\downarrow 14.68}}$ \\
        InternVL3 & 76.16 & 75.10$_{\textcolor{Red}{\downarrow 1.06}}$ \\
        Video-XL-2 & 70.67 & 71.41$_{\textcolor{teal}{\uparrow 0.74}}$ \\

        \midrule
        \bottomrule
    \end{tabular}
    \caption{\textbf{Impact of text information for Video MLLMs on SP, OR and IP tasks.} For video MLLMs, we compare performance between video-only and video with subtitles settings.}
    \label{tab: audio impact on video mllms}
\end{table}

\textbf{Open-source MLLMs show balanced perception but still lag behind in complex reasoning and generation.} As shown in Figure \ref{subfig: task}, despite their relatively small parameter scale (7B/8B), these models achieve solid performance in perceptual tasks, particularly in \textit{Streaming Perception}, yet exhibit clear limitations in reasoning tasks such as \textit{Problem Solving}. In addition to reasoning gaps, their generative outputs often remain coarse or lack precision, indicating that current models excel at low-level multimodal alignment but have difficulty producing coherent and detail rich outputs.

\begin{figure*}[t]
    \centering
    \includegraphics[width=\linewidth]{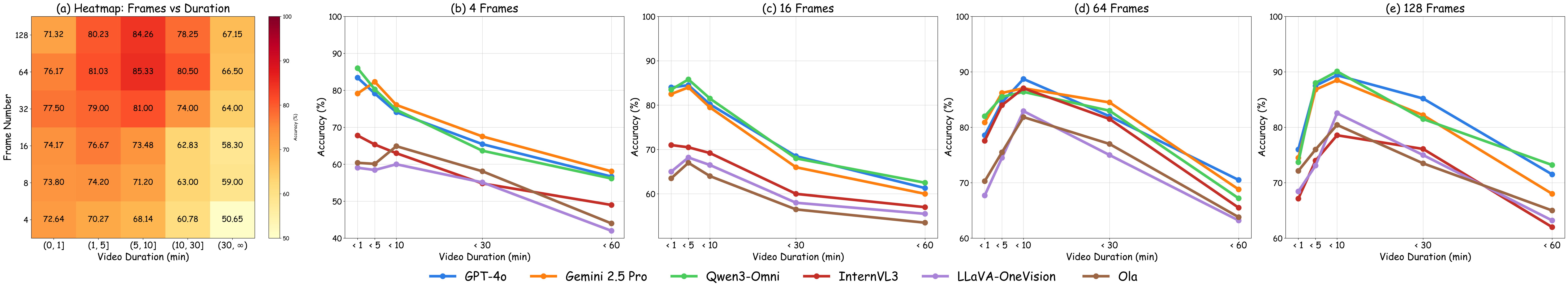}
    \caption{\textbf{Impact of frame sampling and video duration on Streaming Perception performance.} Accuracy decreases with longer videos under sparse sampling, while dense sampling improves performance for longer clips but introduces redundancy in short ones.}
    \label{fig: frame duration analysis}
\end{figure*}

\begin{figure}
    \centering
    \includegraphics[width=0.7\linewidth]{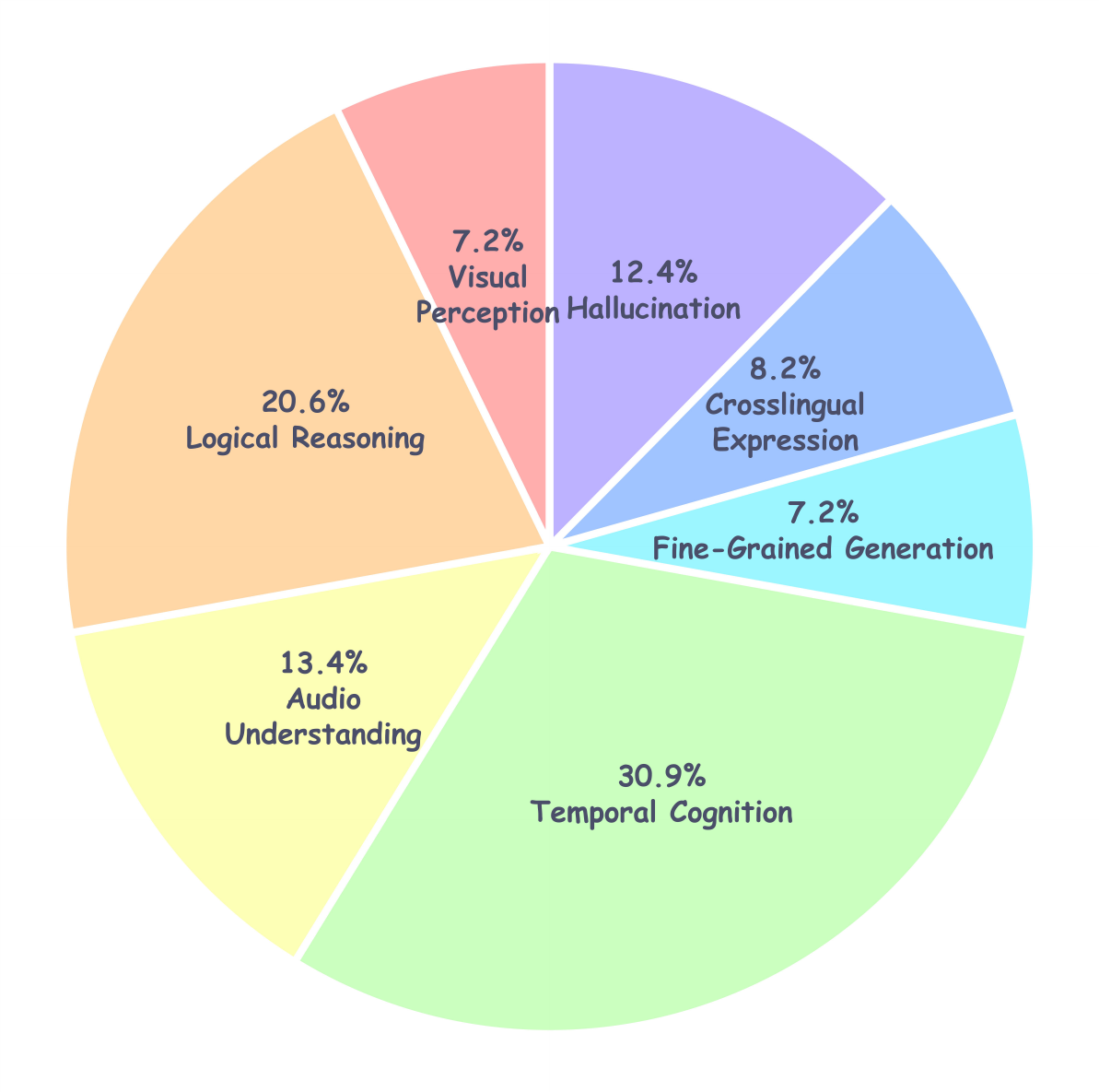}
    \caption{\textbf{Error distribution of MLLMs on LEMON.} Sampled from error cases in the generated results of all evaluated models.}
    \label{fig: error distribution}
\end{figure}

\textbf{All MLLMs exhibit pronounced limitations in temporal causal reasoning.} As reflected in their consistently low performance on \textit{Temporal Awareness} and \textit{Future Content Prediction}, almost all models struggle to capture long-range dependencies, track evolving multimodal signals, and infer the causal structure underlying sequential events, indicating that despite strong perceptual abilities, current systems remain far from achieving robust temporal modeling or reliable causal inference in instructional video settings.


\subsection{Further Analysis}
To better understand the behavior and limitations of current MLLMs on instructional videos, we conduct further analysis from multiple perspectives, including modality impact, linguistic coverage, temporal sensitivity, and error type.

\textbf{Audio and subtitle inputs generally enhance multimodal performance of MLLMs.} As shown in Tables \ref{tab: audio impact on omni mllms} and \ref{tab: audio impact on video mllms}, incorporating subtitles or audio mostly improves performance compared to video-only setting. Subtitles consistently provide the most stable and substantial gain, as they offer explicit textual grounding for visual scenes and spoken explanations typical in instructional content. In contrast, audio input offers far less predictable gains, and omnimodels like Ola and M4 exhibit negative performance shifts, revealing that current MLLMs still face challenges in effectively extracting and aligning semantic information from raw audio signals, particularly when processing complex narrations or domain-specific terminology in teaching videos.

\textbf{Current MLLMs exhibit weak cross-lingual generation, especially for Asian languages.} As shown in Figure \ref{subfig: multilingual}, most models achieve relatively higher performance in Western languages such as FR and DE, while their performance drops in ZH, JA, and KO. The discrepancy indicates a persistent imbalance in multilingual support, with most models focused on English-centric or Latin-script corpora, leaving Asian languages underrepresented in multimodal alignment. Even advanced proprietary models like GPT, Gemini, and Claude show noticeable degradation in Asian language generation when describing or reasoning over instructional content, indicating that cross-lingual semantic grounding and language-specific stylistic modeling remain key challenges.

\textbf{Video performance in instructional scenarios is sensitive to both frame count and video duration.} Figure \ref{fig: frame duration analysis} shows that sparse frame sampling can lead to steadily decreasing accuracy as instructional video length grows. In contrast, denser sampling (64 or 128 frames) can introduce visual redundancy in short clips, often due to repetitive or static teaching scenes, slightly lowering accuracy at first; as video length grows, accuracy improves, but eventually declines for very long videos. This trend indicates that current models still struggle with comprehension and memory retention over extended video sequences, especially when following procedural or step-by-step demonstrations.

\textbf{The error distribution reveals key weaknesses across modalities and tasks in LEMON.} We analyze the failure cases of each task and present the error distribution of different models on LEMON in Figure \ref{fig: error distribution}. Overall, the main shortcomings can be summarized as: 
(i) \textit{\textbf{Limited audio understanding despite strong visual perception.}} Most models can accurately recognize objects and scenes, but often fail to capture the semantics of spoken instructions. 
(ii) \textit{\textbf{Insufficient complex reasoning, especially for temporal or causal dependencies.}} Models struggle with tasks that require understanding procedural steps or causal reasoning common in instructional domains.  
(iii) \textit{\textbf{Focus on coarse-grained descriptions while ignoring fine-grained instructional details.}} Many MLLMs produce only general overviews of demonstrations, overlooking specific difinitions, formulas, concepts or scientific terms.  
(iv) \textit{\textbf{Prevalent hallucinations.}} Models sometimes generate responses inconsistent with input modalities, misinterpret prompt information, or violate required output formats, such as misunderstanding spoken cues or introducing irrelevant instructional steps.  
Detailed examples and further analyses of each error type are provided in the supplementary material.
\section{Conclusion}
In this work, we conduct a comprehensive evaluation of state-of-the-art MLLMs on the LEMON benchmark, which focuses on instructional videos integrating visual, audio, and textual modalities. Our findings indicate that, although current models perform well in visual perception and basic multimodal alignment, significant challenges persist in audio comprehension, temporal and causal reasoning, as well as fine-grained or cross-lingual content generation. Proprietary models generally outperform open-source models, particularly in reasoning and multilingual tasks, yet hallucinations and output inconsistencies remain prevalent. By analyzing these limitations, we hope to encourage future research toward building MLLMs with enhanced omnimodal understanding in realistic instructional and real-world video contexts, ultimately leading to more reliable and contextually grounded multimodal systems.

{
     \small
     \bibliographystyle{ieeenat_fullname}
     \bibliography{main}

@String(AAAI = {AAAI})

@article{yin2024survey,
  title={A survey on multimodal large language models},
  author={Yin, Shukang and Fu, Chaoyou and Zhao, Sirui and Li, Ke and Sun, Xing and Xu, Tong and Chen, Enhong},
  journal={National Science Review},
  volume={11},
  number={12},
  pages={nwae403},
  year={2024},
  publisher={Oxford University Press}
}

@article{hurst2024gpt,
  title={Gpt-4o system card},
  author={Hurst, Aaron and Lerer, Adam and Goucher, Adam P and Perelman, Adam and Ramesh, Aditya and Clark, Aidan and Ostrow, AJ and Welihinda, Akila and Hayes, Alan and Radford, Alec and others},
  journal={arXiv preprint arXiv:2410.21276},
  year={2024}
}

@article{comanici2025gemini,
  title={Gemini 2.5: Pushing the frontier with advanced reasoning, multimodality, long context, and next generation agentic capabilities},
  author={Comanici, Gheorghe and Bieber, Eric and Schaekermann, Mike and Pasupat, Ice and Sachdeva, Noveen and Dhillon, Inderjit and Blistein, Marcel and Ram, Ori and Zhang, Dan and Rosen, Evan and others},
  journal={arXiv preprint arXiv:2507.06261},
  year={2025}
}

@article{lin2023video,
  title={Video-llava: Learning united visual representation by alignment before projection},
  author={Lin, Bin and Ye, Yang and Zhu, Bin and Cui, Jiaxi and Ning, Munan and Jin, Peng and Yuan, Li},
  journal={arXiv preprint arXiv:2311.10122},
  year={2023}
}

@inproceedings{zhang2021vinvl,
  title={Vinvl: Revisiting visual representations in vision-language models},
  author={Zhang, Pengchuan and Li, Xiujun and Hu, Xiaowei and Yang, Jianwei and Zhang, Lei and Wang, Lijuan and Choi, Yejin and Gao, Jianfeng},
  booktitle={Proceedings of the IEEE/CVF Conference on Computer Vision and Pattern Recognition},
  pages={5579--5588},
  year={2021}
}

@article{yang2025qwen3,
  title={Qwen3 technical report},
  author={Yang, An and Li, Anfeng and Yang, Baosong and Zhang, Beichen and Hui, Binyuan and Zheng, Bo and Yu, Bowen and Gao, Chang and Huang, Chengen and Lv, Chenxu and others},
  journal={arXiv preprint arXiv:2505.09388},
  year={2025}
}

@article{chu2024qwen2,
  title={Qwen2-audio technical report},
  author={Chu, Yunfei and Xu, Jin and Yang, Qian and Wei, Haojie and Wei, Xipin and Guo, Zhifang and Leng, Yichong and Lv, Yuanjun and He, Jinzheng and Lin, Junyang and others},
  journal={arXiv preprint arXiv:2407.10759},
  year={2024}
}

@article{fang2024mmbench,
  title={Mmbench-video: A long-form multi-shot benchmark for holistic video understanding},
  author={Fang, Xinyu and Mao, Kangrui and Duan, Haodong and Zhao, Xiangyu and Li, Yining and Lin, Dahua and Chen, Kai},
  journal={Advances in Neural Information Processing Systems},
  volume={37},
  pages={89098--89124},
  year={2024}
}

@inproceedings{fu2025video,
  title={Video-mme: The first-ever comprehensive evaluation benchmark of multi-modal llms in video analysis},
  author={Fu, Chaoyou and Dai, Yuhan and Luo, Yongdong and Li, Lei and Ren, Shuhuai and Zhang, Renrui and Wang, Zihan and Zhou, Chenyu and Shen, Yunhang and Zhang, Mengdan and others},
  booktitle={Proceedings of the Computer Vision and Pattern Recognition Conference},
  pages={24108--24118},
  year={2025}
}

@inproceedings{li2024mvbench,
  title={Mvbench: A comprehensive multi-modal video understanding benchmark},
  author={Li, Kunchang and Wang, Yali and He, Yinan and Li, Yizhuo and Wang, Yi and Liu, Yi and Wang, Zun and Xu, Jilan and Chen, Guo and Luo, Ping and others},
  booktitle={Proceedings of the IEEE/CVF Conference on Computer Vision and Pattern Recognition},
  pages={22195--22206},
  year={2024}
}

@article{chaoyou2023mme,
  title={Mme: A comprehensive evaluation benchmark for multimodal large language models},
  author={Chaoyou, Fu and Peixian, Chen and Yunhang, Shen and Yulei, Qin and Mengdan, Zhang and Xu, Lin and Jinrui, Yang and Xiawu, Zheng and Ke, Li and Xing, Sun and others},
  journal={arXiv preprint arXiv:2306.13394},
  volume={3},
  year={2023}
}

@article{lu2023mathvista,
  title={Mathvista: Evaluating math reasoning in visual contexts with gpt-4v, bard, and other large multimodal models},
  author={Lu, Pan and Bansal, Hritik and Xia, Tony and Liu, Jiacheng and Li, Chunyuan and Hajishirzi, Hannaneh and Cheng, Hao and Chang, Kai-Wei and Galley, Michel and Gao, Jianfeng},
  journal={CoRR},
  year={2023}
}

@article{yu2023mm,
  title={Mm-vet: Evaluating large multimodal models for integrated capabilities},
  author={Yu, Weihao and Yang, Zhengyuan and Li, Linjie and Wang, Jianfeng and Lin, Kevin and Liu, Zicheng and Wang, Xinchao and Wang, Lijuan},
  journal={arXiv preprint arXiv:2308.02490},
  year={2023}
}

@article{lin2024streamingbench,
  title={Streamingbench: Assessing the gap for mllms to achieve streaming video understanding},
  author={Lin, Junming and Fang, Zheng and Chen, Chi and Wan, Zihao and Luo, Fuwen and Li, Peng and Liu, Yang and Sun, Maosong},
  journal={arXiv preprint arXiv:2411.03628},
  year={2024}
}

@inproceedings{niu2025ovo,
  title={OVO-Bench: How Far is Your Video-LLMs from Real-World Online Video Understanding?},
  author={Niu, Junbo and Li, Yifei and Miao, Ziyang and Ge, Chunjiang and Zhou, Yuanhang and He, Qihao and Dong, Xiaoyi and Duan, Haodong and Ding, Shuangrui and Qian, Rui and others},
  booktitle={Proceedings of the Computer Vision and Pattern Recognition Conference},
  pages={18902--18913},
  year={2025}
}

@article{huang2024online,
  title={Online video understanding: A comprehensive benchmark and memory-augmented method},
  author={Huang, Zhenpeng and Li, Xinhao and Li, Jiaqi and Wang, Jing and Zeng, Xiangyu and Liang, Cheng and Wu, Tao and Chen, Xi and Li, Liang and Wang, Limin},
  journal={arXiv preprint arXiv:2501.00584},
  year={2024}
}

@article{wang2025proactivevideoqa,
  title={Proactivevideoqa: A comprehensive benchmark evaluating proactive interactions in video large language models},
  author={Wang, Yueqian and Meng, Xiaojun and Wang, Yifan and Zhang, Huishuai and Zhao, Dongyan},
  journal={arXiv preprint arXiv:2507.09313},
  year={2025}
}

@inproceedings{jang2017tgif,
  title={Tgif-qa: Toward spatio-temporal reasoning in visual question answering},
  author={Jang, Yunseok and Song, Yale and Yu, Youngjae and Kim, Youngjin and Kim, Gunhee},
  booktitle={Proceedings of the IEEE/CVF Conference on Computer Vision and Pattern Recognition},
  pages={2758--2766},
  year={2017}
}

@inproceedings{xu2017video,
  title={Video question answering via gradually refined attention over appearance and motion},
  author={Xu, Dejing and Zhao, Zhou and Xiao, Jun and Wu, Fei and Zhang, Hanwang and He, Xiangnan and Zhuang, Yueting},
  booktitle={Proceedings of the 25th ACM international conference on Multimedia},
  pages={1645--1653},
  year={2017}
}

@inproceedings{yu2019activitynet,
  title={Activitynet-qa: A dataset for understanding complex web videos via question answering},
  author={Yu, Zhou and Xu, Dejing and Yu, Jun and Yu, Ting and Zhao, Zhou and Zhuang, Yueting and Tao, Dacheng},
  booktitle={Proceedings of the AAAI Conference on Artificial Intelligence},
  volume={33},
  number={01},
  pages={9127--9134},
  year={2019}
}

@inproceedings{caba2015activitynet,
  title={Activitynet: A large-scale video benchmark for human activity understanding},
  author={Caba Heilbron, Fabian and Escorcia, Victor and Ghanem, Bernard and Carlos Niebles, Juan},
  booktitle={Proceedings of the IEEE/CVF Conference on Computer Vision and Pattern Recognition},
  pages={961--970},
  year={2015}
}

@inproceedings{xu2016msr,
  title={Msr-vtt: A large video description dataset for bridging video and language},
  author={Xu, Jun and Mei, Tao and Yao, Ting and Rui, Yong},
  booktitle={Proceedings of the IEEE/CVF Conference on Computer Vision and Pattern Recognition},
  pages={5288--5296},
  year={2016}
}

@inproceedings{li2024vitatecs,
  title={Vitatecs: A diagnostic dataset for temporal concept understanding of video-language models},
  author={Li, Shicheng and Li, Lei and Liu, Yi and Ren, Shuhuai and Liu, Yuanxin and Gao, Rundong and Sun, Xu and Hou, Lu},
  booktitle={European Conference on Computer Vision},
  pages={331--348},
  year={2024},
  organization={Springer}
}

@article{liu2024tempcompass,
  title={Tempcompass: Do video llms really understand videos?},
  author={Liu, Yuanxin and Li, Shicheng and Liu, Yi and Wang, Yuxiang and Ren, Shuhuai and Li, Lei and Chen, Sishuo and Sun, Xu and Hou, Lu},
  journal={arXiv preprint arXiv:2403.00476},
  year={2024}
}

@article{mangalam2023egoschema,
  title={Egoschema: A diagnostic benchmark for very long-form video language understanding},
  author={Mangalam, Karttikeya and Akshulakov, Raiymbek and Malik, Jitendra},
  journal={Advances in Neural Information Processing Systems},
  volume={36},
  pages={46212--46244},
  year={2023}
}

@article{zhou2024mlvu,
  title={Mlvu: A comprehensive benchmark for multi-task long video understanding},
  author={Zhou, Junjie and Shu, Yan and Zhao, Bo and Wu, Boya and Xiao, Shitao and Yang, Xi and Xiong, Yongping and Zhang, Bo and Huang, Tiejun and Liu, Zheng},
  journal={arXiv preprint arXiv:2406.04264},
  year={2024}
}

@inproceedings{wang2025omnimmi,
  title={OmniMMI: A Comprehensive Multi-modal Interaction Benchmark in Streaming Video Contexts},
  author={Wang, Yuxuan and Wang, Yueqian and Chen, Bo and Wu, Tong and Zhao, Dongyan and Zheng, Zilong},
  booktitle={Proceedings of the Computer Vision and Pattern Recognition Conference},
  pages={18925--18935},
  year={2025}
}

@article{hong2025worldsense,
  title={Worldsense: Evaluating real-world omnimodal understanding for multimodal llms},
  author={Hong, Jack and Yan, Shilin and Cai, Jiayin and Jiang, Xiaolong and Hu, Yao and Xie, Weidi},
  journal={arXiv preprint arXiv:2502.04326},
  year={2025}
}

@article{li2024omnibench,
  title={Omnibench: Towards the future of universal omni-language models},
  author={Li, Yizhi and Zhang, Ge and Ma, Yinghao and Yuan, Ruibin and Zhu, Kang and Guo, Hangyu and Liang, Yiming and Liu, Jiaheng and Wang, Zekun and Yang, Jian and others},
  journal={arXiv preprint arXiv:2409.15272},
  year={2024}
}

@article{yao2025timechat,
  title={TimeChat-Online: 80\% Visual Tokens are Naturally Redundant in Streaming Videos},
  author={Yao, Linli and Li, Yicheng and Wei, Yuancheng and Li, Lei and Ren, Shuhuai and Liu, Yuanxin and Ouyang, Kun and Wang, Lean and Li, Shicheng and Li, Sida and others},
  journal={arXiv preprint arXiv:2504.17343},
  year={2025}
}

@article{tang2025video,
  title={video-SALMONN 2: Captioning-Enhanced Audio-Visual Large Language Models},
  author={Tang, Changli and Li, Yixuan and Yang, Yudong and Zhuang, Jimin and Sun, Guangzhi and Li, Wei and Ma, Zejun and Zhang, Chao},
  journal={arXiv preprint arXiv:2506.15220},
  year={2025}
}

@article{lei2018tvqa,
  title={Tvqa: Localized, compositional video question answering},
  author={Lei, Jie and Yu, Licheng and Bansal, Mohit and Berg, Tamara L},
  journal={arXiv preprint arXiv:1809.01696},
  year={2018}
}

@article{li2020hero,
  title={Hero: Hierarchical encoder for video+ language omni-representation pre-training},
  author={Li, Linjie and Chen, Yen-Chun and Cheng, Yu and Gan, Zhe and Yu, Licheng and Liu, Jingjing},
  journal={arXiv preprint arXiv:2005.00200},
  year={2020}
}

@article{zhang20252,
  title={2.5 years in class: A multimodal textbook for vision-language pretraining},
  author={Zhang, Wenqi and Zhang, Hang and Li, Xin and Sun, Jiashuo and Shen, Yongliang and Lu, Weiming and Zhao, Deli and Zhuang, Yueting and Bing, Lidong},
  journal={arXiv preprint arXiv:2501.00958},
  year={2025}
}

@inproceedings{xiao2021next,
  title={Next-qa: Next phase of question-answering to explaining temporal actions},
  author={Xiao, Junbin and Shang, Xindi and Yao, Angela and Chua, Tat-Seng},
  booktitle={Proceedings of the IEEE/CVF Conference on Computer Vision and Pattern Recognition},
  pages={9777--9786},
  year={2021}
}

@article{ning2023video,
  title={Video-bench: A comprehensive benchmark and toolkit for evaluating video-based large language models},
  author={Ning, Munan and Zhu, Bin and Xie, Yujia and Lin, Bin and Cui, Jiaxi and Yuan, Lu and Chen, Dongdong and Yuan, Li},
  journal={arXiv preprint arXiv:2311.16103},
  year={2023}
}

@inproceedings{song2024moviechat,
  title={Moviechat: From dense token to sparse memory for long video understanding},
  author={Song, Enxin and Chai, Wenhao and Wang, Guanhong and Zhang, Yucheng and Zhou, Haoyang and Wu, Feiyang and Chi, Haozhe and Guo, Xun and Ye, Tian and Zhang, Yanting and others},
  booktitle={Proceedings of the IEEE/CVF Conference on Computer Vision and Pattern Recognition},
  pages={18221--18232},
  year={2024}
}

@article{wu2024longvideobench,
  title={Longvideobench: A benchmark for long-context interleaved video-language understanding},
  author={Wu, Haoning and Li, Dongxu and Chen, Bei and Li, Junnan},
  journal={Advances in Neural Information Processing Systems},
  volume={37},
  pages={28828--28857},
  year={2024}
}

@inproceedings{geng2025longvale,
  title={Longvale: Vision-audio-language-event benchmark towards time-aware omni-modal perception of long videos},
  author={Geng, Tiantian and Zhang, Jinrui and Wang, Qingni and Wang, Teng and Duan, Jinming and Zheng, Feng},
  booktitle={Proceedings of the Computer Vision and Pattern Recognition Conference},
  pages={18959--18969},
  year={2025}
}

@article{chen2023vlp,
  title={Vlp: A survey on vision-language pre-training},
  author={Chen, Fei-Long and Zhang, Du-Zhen and Han, Ming-Lun and Chen, Xiu-Yi and Shi, Jing and Xu, Shuang and Xu, Bo},
  journal={Machine Intelligence Research},
  volume={20},
  number={1},
  pages={38--56},
  year={2023},
  publisher={Springer}
}

@article{jiang2024effectiveness,
  title={Effectiveness assessment of recent large vision-language models},
  author={Jiang, Yao and Yan, Xinyu and Ji, Ge-Peng and Fu, Keren and Sun, Meijun and Xiong, Huan and Fan, Deng-Ping and Khan, Fahad Shahbaz},
  journal={Visual Intelligence},
  volume={2},
  number={1},
  pages={17},
  year={2024},
  publisher={Springer}
}

@article{wang2023large,
  title={Large-scale multi-modal pre-trained models: A comprehensive survey},
  author={Wang, Xiao and Chen, Guangyao and Qian, Guangwu and Gao, Pengcheng and Wei, Xiao-Yong and Wang, Yaowei and Tian, Yonghong and Gao, Wen},
  journal={Machine Intelligence Research},
  volume={20},
  number={4},
  pages={447--482},
  year={2023},
  publisher={Springer}
}

@article{li2023videochat,
  title={Videochat: Chat-centric video understanding},
  author={Li, KunChang and He, Yinan and Wang, Yi and Li, Yizhuo and Wang, Wenhai and Luo, Ping and Wang, Yali and Wang, Limin and Qiao, Yu},
  journal={arXiv preprint arXiv:2305.06355},
  year={2023}
}

@article{zhang2023video,
  title={Video-llama: An instruction-tuned audio-visual language model for video understanding},
  author={Zhang, Hang and Li, Xin and Bing, Lidong},
  journal={arXiv preprint arXiv:2306.02858},
  year={2023}
}

@article{yao2024minicpm,
  title={Minicpm-v: A gpt-4v level mllm on your phone},
  author={Yao, Yuan and Yu, Tianyu and Zhang, Ao and Wang, Chongyi and Cui, Junbo and Zhu, Hongji and Cai, Tianchi and Li, Haoyu and Zhao, Weilin and He, Zhihui and others},
  journal={arXiv preprint arXiv:2408.01800},
  year={2024}
}

@article{bai2025qwen2,
  title={Qwen2. 5-vl technical report},
  author={Bai, Shuai and Chen, Keqin and Liu, Xuejing and Wang, Jialin and Ge, Wenbin and Song, Sibo and Dang, Kai and Wang, Peng and Wang, Shijie and Tang, Jun and others},
  journal={arXiv preprint arXiv:2502.13923},
  year={2025}
}

@article{chen2024sharegpt4video,
  title={Sharegpt4video: Improving video understanding and generation with better captions},
  author={Chen, Lin and Wei, Xilin and Li, Jinsong and Dong, Xiaoyi and Zhang, Pan and Zang, Yuhang and Chen, Zehui and Duan, Haodong and Tang, Zhenyu and Yuan, Li and others},
  journal={Advances in Neural Information Processing Systems},
  volume={37},
  pages={19472--19495},
  year={2024}
}

@article{zhang2024long,
  title={Long context transfer from language to vision},
  author={Zhang, Peiyuan and Zhang, Kaichen and Li, Bo and Zeng, Guangtao and Yang, Jingkang and Zhang, Yuanhan and Wang, Ziyue and Tan, Haoran and Li, Chunyuan and Liu, Ziwei},
  journal={arXiv preprint arXiv:2406.16852},
  year={2024}
}

@article{wang2024longllava,
  title={Longllava: Scaling multi-modal llms to 1000 images efficiently via a hybrid architecture},
  author={Wang, Xidong and Song, Dingjie and Chen, Shunian and Chen, Junyin and Cai, Zhenyang and Zhang, Chen and Sun, Lichao and Wang, Benyou},
  journal={arXiv preprint arXiv:2409.02889},
  year={2024}
}

@article{shen2024longvu,
  title={Longvu: Spatiotemporal adaptive compression for long video-language understanding},
  author={Shen, Xiaoqian and Xiong, Yunyang and Zhao, Changsheng and Wu, Lemeng and Chen, Jun and Zhu, Chenchen and Liu, Zechun and Xiao, Fanyi and Varadarajan, Balakrishnan and Bordes, Florian and others},
  journal={arXiv preprint arXiv:2410.17434},
  year={2024}
}

@article{zhang2024flash,
  title={Flash-vstream: Memory-based real-time understanding for long video streams},
  author={Zhang, Haoji and Wang, Yiqin and Tang, Yansong and Liu, Yong and Feng, Jiashi and Dai, Jifeng and Jin, Xiaojie},
  journal={arXiv preprint arXiv:2406.08085},
  year={2024}
}

@inproceedings{huang2025online,
  title={Online Video Understanding: OVBench and VideoChat-Online},
  author={Huang, Zhenpeng and Li, Xinhao and Li, Jiaqi and Wang, Jing and Zeng, Xiangyu and Liang, Cheng and Wu, Tao and Chen, Xi and Li, Liang and Wang, Limin},
  booktitle={Proceedings of the Computer Vision and Pattern Recognition Conference},
  pages={3328--3338},
  year={2025}
}

@inproceedings{chen2024videollm,
  title={Videollm-online: Online video large language model for streaming video},
  author={Chen, Joya and Lv, Zhaoyang and Wu, Shiwei and Lin, Kevin Qinghong and Song, Chenan and Gao, Difei and Liu, Jia-Wei and Gao, Ziteng and Mao, Dongxing and Shou, Mike Zheng},
  booktitle={Proceedings of the IEEE/CVF Conference on Computer Vision and Pattern Recognition},
  pages={18407--18418},
  year={2024}
}

@article{fu2025vita,
  title={Vita-1.5: Towards gpt-4o level real-time vision and speech interaction},
  author={Fu, Chaoyou and Lin, Haojia and Wang, Xiong and Zhang, Yi-Fan and Shen, Yunhang and Liu, Xiaoyu and Cao, Haoyu and Long, Zuwei and Gao, Heting and Li, Ke and others},
  journal={arXiv preprint arXiv:2501.01957},
  year={2025}
}

@article{xie2024mini,
  title={Mini-omni2: Towards open-source gpt-4o with vision, speech and duplex capabilities},
  author={Xie, Zhifei and Wu, Changqiao},
  journal={arXiv preprint arXiv:2410.11190},
  year={2024}
}

@article{xu2025qwen3,
  title={Qwen3-omni technical report},
  author={Xu, Jin and Guo, Zhifang and Hu, Hangrui and Chu, Yunfei and Wang, Xiong and He, Jinzheng and Wang, Yuxuan and Shi, Xian and He, Ting and Zhu, Xinfa and others},
  journal={arXiv preprint arXiv:2509.17765},
  year={2025}
}

@article{zhang2024internlm,
  title={Internlm-xcomposer2. 5-omnilive: A comprehensive multimodal system for long-term streaming video and audio interactions},
  author={Zhang, Pan and Dong, Xiaoyi and Cao, Yuhang and Zang, Yuhang and Qian, Rui and Wei, Xilin and Chen, Lin and Li, Yifei and Niu, Junbo and Ding, Shuangrui and others},
  journal={arXiv preprint arXiv:2412.09596},
  year={2024}
}

@techreport{openbmb2025minicpmo,
  author       = {OpenBMB},
  title        = {MiniCPM-o 2.6: A GPT-4o Level MLLM for Vision, Speech, and Multimodal Live Streaming on Your Phone},
  institution  = {OpenBMB},
  year         = {2025},
  note         = {Technical report}
}

@inproceedings{wang2024videoagent,
  title={Videoagent: Long-form video understanding with large language model as agent},
  author={Wang, Xiaohan and Zhang, Yuhui and Zohar, Orr and Yeung-Levy, Serena},
  booktitle={European Conference on Computer Vision},
  pages={58--76},
  year={2024},
  organization={Springer}
}

@article{xu2023retrieval,
  title={Retrieval-based video language model for efficient long video question answering},
  author={Xu, Jiaqi and Lan, Cuiling and Xie, Wenxuan and Chen, Xuejin and Lu, Yan},
  journal={arXiv preprint arXiv:2312.04931},
  year={2023}
}

@article{kay2017kinetics,
  title={The kinetics human action video dataset},
  author={Kay, Will and Carreira, Joao and Simonyan, Karen and Zhang, Brian and Hillier, Chloe and Vijayanarasimhan, Sudheendra and Viola, Fabio and Green, Tim and Back, Trevor and Natsev, Paul and others},
  journal={arXiv preprint arXiv:1705.06950},
  year={2017}
}

@article{wang2023paxion,
  title={Paxion: Patching action knowledge in video-language foundation models},
  author={Wang, Zhenhailong and Blume, Ansel and Li, Sha and Liu, Genglin and Cho, Jaemin and Tang, Zineng and Bansal, Mohit and Ji, Heng},
  journal={Advances in Neural Information Processing Systems},
  volume={36},
  pages={20729--20749},
  year={2023}
}

@article{wu2024star,
  title={Star: A benchmark for situated reasoning in real-world videos},
  author={Wu, Bo and Yu, Shoubin and Chen, Zhenfang and Tenenbaum, Joshua B and Gan, Chuang},
  journal={arXiv preprint arXiv:2405.09711},
  year={2024}
}

@techreport{openai2025gpt5,
  author       = {OpenAI},
  title        = {GPT‑5: Our Smartest, Fastest, Most Useful Model Yet},
  institution  = {OpenAI},
  year         = {2025},
  month        = aug,
  note         = {System card / technical overview}
}

@article{xu2025qwen2,
  title={Qwen2. 5-omni technical report},
  author={Xu, Jin and Guo, Zhifang and He, Jinzheng and Hu, Hangrui and He, Ting and Bai, Shuai and Chen, Keqin and Wang, Jialin and Fan, Yang and Dang, Kai and others},
  journal={arXiv preprint arXiv:2503.20215},
  year={2025}
}

@techreport{Anthropic2025ClaudeSonnet45,
  author       = {Anthropic},
  title        = {Introducing Claude Sonnet 4.5},
  institution  = {Anthropic},
  year         = {2025},
  month        = {sep},
  day          = {29},
  type         = {Technical Report},
  url          = {https://www.anthropic.com/news/claude-sonnet-4-5},
  note         = {Technical Report}
}

@article{li2025baichuan,
  title={Baichuan-omni-1.5 technical report},
  author={Li, Yadong and Liu, Jun and Zhang, Tao and Chen, Song and Li, Tianpeng and Li, Zehuan and Liu, Lijun and Ming, Lingfeng and Dong, Guosheng and Pan, Da and others},
  journal={arXiv preprint arXiv:2501.15368},
  year={2025}
}

@article{liu2025ola,
  title={Ola: Pushing the frontiers of omni-modal language model},
  author={Liu, Zuyan and Dong, Yuhao and Wang, Jiahui and Liu, Ziwei and Hu, Winston and Lu, Jiwen and Rao, Yongming},
  journal={arXiv preprint arXiv:2502.04328},
  year={2025}
}

@inproceedings{qian2025dispider,
  title={Dispider: Enabling video llms with active real-time interaction via disentangled perception, decision, and reaction},
  author={Qian, Rui and Ding, Shuangrui and Dong, Xiaoyi and Zhang, Pan and Zang, Yuhang and Cao, Yuhang and Lin, Dahua and Wang, Jiaqi},
  booktitle={Proceedings of the Computer Vision and Pattern Recognition Conference},
  pages={24045--24055},
  year={2025}
}

@article{zhang2025videollama,
  title={Videollama 3: Frontier multimodal foundation models for image and video understanding},
  author={Zhang, Boqiang and Li, Kehan and Cheng, Zesen and Hu, Zhiqiang and Yuan, Yuqian and Chen, Guanzheng and Leng, Sicong and Jiang, Yuming and Zhang, Hang and Li, Xin and others},
  journal={arXiv preprint arXiv:2501.13106},
  year={2025}
}

@article{zhu2025internvl3,
  title={Internvl3: Exploring advanced training and test-time recipes for open-source multimodal models},
  author={Zhu, Jinguo and Wang, Weiyun and Chen, Zhe and Liu, Zhaoyang and Ye, Shenglong and Gu, Lixin and Tian, Hao and Duan, Yuchen and Su, Weijie and Shao, Jie and others},
  journal={arXiv preprint arXiv:2504.10479},
  year={2025}
}

@article{qin2025video,
  title={Video-XL-2: Towards Very Long-Video Understanding Through Task-Aware KV Sparsification},
  author={Qin, Minghao and Liu, Xiangrui and Liang, Zhengyang and Shu, Yan and Yuan, Huaying and Zhou, Juenjie and Xiao, Shitao and Zhao, Bo and Liu, Zheng},
  journal={arXiv preprint arXiv:2506.19225},
  year={2025}
}

@article{li2024llava,
  title={Llava-onevision: Easy visual task transfer},
  author={Li, Bo and Zhang, Yuanhan and Guo, Dong and Zhang, Renrui and Li, Feng and Zhang, Hao and Zhang, Kaichen and Zhang, Peiyuan and Li, Yanwei and Liu, Ziwei and others},
  journal={arXiv preprint arXiv:2408.03326},
  year={2024}
}

@techreport{grok4_xai_2025,
  author       = {xAI},
  title        = {Grok 4: The Latest Flagship Model from xAI},
  institution  = {xAI, Inc.},
  year         = {2025},
  note         = {Version v1.0}
}

@inproceedings{papineni-etal-2002-bleu,
    title = "{B}leu: a Method for Automatic Evaluation of Machine Translation",
    author = "Papineni, Kishore  and
      Roukos, Salim  and
      Ward, Todd  and
      Zhu, Wei-Jing",
    editor = "Isabelle, Pierre  and
      Charniak, Eugene  and
      Lin, Dekang",
    booktitle = "Proceedings of the 40th Annual Meeting of the Association for Computational Linguistics",
    month = jul,
    year = "2002",
    address = "Philadelphia, Pennsylvania, USA",
    publisher = "Association for Computational Linguistics",
    url = "https://aclanthology.org/P02-1040/",
    doi = "10.3115/1073083.1073135",
    pages = "311--318"
}

@inproceedings{lin-2004-rouge,
    title = "{ROUGE}: A Package for Automatic Evaluation of Summaries",
    author = "Lin, Chin-Yew",
    booktitle = "Text Summarization Branches Out",
    month = jul,
    year = "2004",
    address = "Barcelona, Spain",
    publisher = "Association for Computational Linguistics",
    url = "https://aclanthology.org/W04-1013/",
    pages = "74--81"
}

@article{zhang2019bertscore,
  title={Bertscore: Evaluating text generation with bert},
  author={Zhang, Tianyi and Kishore, Varsha and Wu, Felix and Weinberger, Kilian Q and Artzi, Yoav},
  journal={arXiv preprint arXiv:1904.09675},
  year={2019}
}

@misc{google2024gemini2,
  title        = {Introducing Gemini 2.0: Our New AI Model for the Agentic Era},
  author       = {{Google DeepMind}},
  year         = {2024},
  month        = dec,
  note         = {Google DeepMind Blog Post}
}
}

\clearpage
\setcounter{page}{1}
\maketitlesupplementary

\setcounter{figure}{6}
\setcounter{table}{4}
\renewcommand{\thesection}{\Alph{section}}  

\section{Overall of Appendix}
\begin{itemize}
    \item \cref{sec: More Details about LEMON}: \textbf{More Details of Benchmark Construction}
    \item \cref{sec: Additional Dataset Statistics}: \textbf{Additional Dataset Statistics}
    \item \cref{sec: Detailed Experimental Settings}: \textbf{Detailed Experimental Settings}
    \item \cref{sec: Further Experiments}: \textbf{Further Experiments}
    \item \cref{sec: Failure Case}: \textbf{Failure Case}
\end{itemize}

\section{More Details of Benchmark Construction} \label{sec: More Details about LEMON}
\subsection{Video Selection}
To construct a diverse and pedagogically meaningful benchmark for temporal multimodal understanding on instructional videos, we carefully selected five STEM domains, \textbf{Mathematics}, \textbf{Artificial Intelligence}, \textbf{Computer Science}, \textbf{Electronic Engineering}, and \textbf{Robotics}, as the core content areas. They are chosen based on three considerations.

First, they collectively cover a broad spectrum of cognitive demand, ranging from symbolic reasoning (Mathematics), conceptual abstraction (AI), procedural logic (Computer Science), signal–hardware interpretation (Electronic Engineering), to dynamic physical interaction (Robotics).

Second, these domains are prevalent in real-world instructional content and encompass core academic and practical skills, providing a natural setting to evaluate whether MLLMs can understand step-by-step demonstrations, tool usage, and domain-specific terminology.

Third, STEM videos inherently involve multimodal complexity, formulas written on screen, code demonstrations, diagrams, circuit boards, robot motion, spoken explanation, making them ideal for probing temporal grounding, cross-modal alignment, and causal reasoning.

\subsubsection{Selection Criteria}
We established a set of requirements to ensure high-quality, instruction-centric videos:
\begin{itemize}
    \item \textbf{Instructional nature}: Videos must contain explicit teaching, explanation, step-by-step demonstration, or popular science of a certain knowledge rather than entertainment-style content.
    \item \textbf{Clear multimodal signals}: The visual modality should include interpretable diagrams, slides, code, equations, or physical apparatus. And audio must include coherent narration or explanation.
    \item \textbf{Sufficient temporal structure}: Videos should exhibit clear progression of steps or stages, enabling the evaluation of temporal reasoning.
    \item \textbf{Language constraints}: Videos must include English narration to maintain consistent transcript and subtitle.
    \item \textbf{Noise control}: Videos with excessive background noise, music overlays, or rapid scene switching are excluded.
    \item \textbf{Duration bounds}: We selected videos less than one hour, long enough to capture temporal dependencies but short enough to maintain consistent semantic density.
\end{itemize}

\subsubsection{YouTube Filtering Pipeline}
We implemented a multi-stage pipeline to retrieve and curate videos from YouTube:

\textbf{Keyword-based Retrieval}. For each domain, we construct keyword lists such as ``linear algebra'', ``neural networks'', ``data structures'', ``analog electronics'', and ``robotics''. And using YouTube's search API \footnote{\url{https://www.youtube.com}}, we collect several candidates per domain.

\textbf{Metadata Filtering}. We remove videos with insufficient technical quality or poor instructional relevance. Videos with low resolution ($<$ 720p), misleading tags or thumbnails, or extremely low engagement metrics were excluded at this stage, as these signals often correlate with unclear or low-quality instructional content.

\textbf{Automatic Content Inspection}. A lightweight scan using Whisper \footnote{\url{https://huggingface.co/openai/whisper-large-v3}} + CLIP \footnote{\url{https://huggingface.co/openai/clip-vit-large-patch14}} features is then applied to filter out silent videos, static slide decks, and monologue-style recordings without meaningful instructional visuals, ensuring that selected videos contain multimodal cues rather than purely verbal explanations.

\textbf{Manual Verification}. Finally, we manually check the remaining videos to confirm correct subject categorization and the presence of demonstrative actions, such as solving equations, writing code, assembling circuits, or operating robots, together with clear narration, visual alignment. Only videos meeting all criteria are retained.

\begin{figure*}
    \centering
    \includegraphics[width=\linewidth]{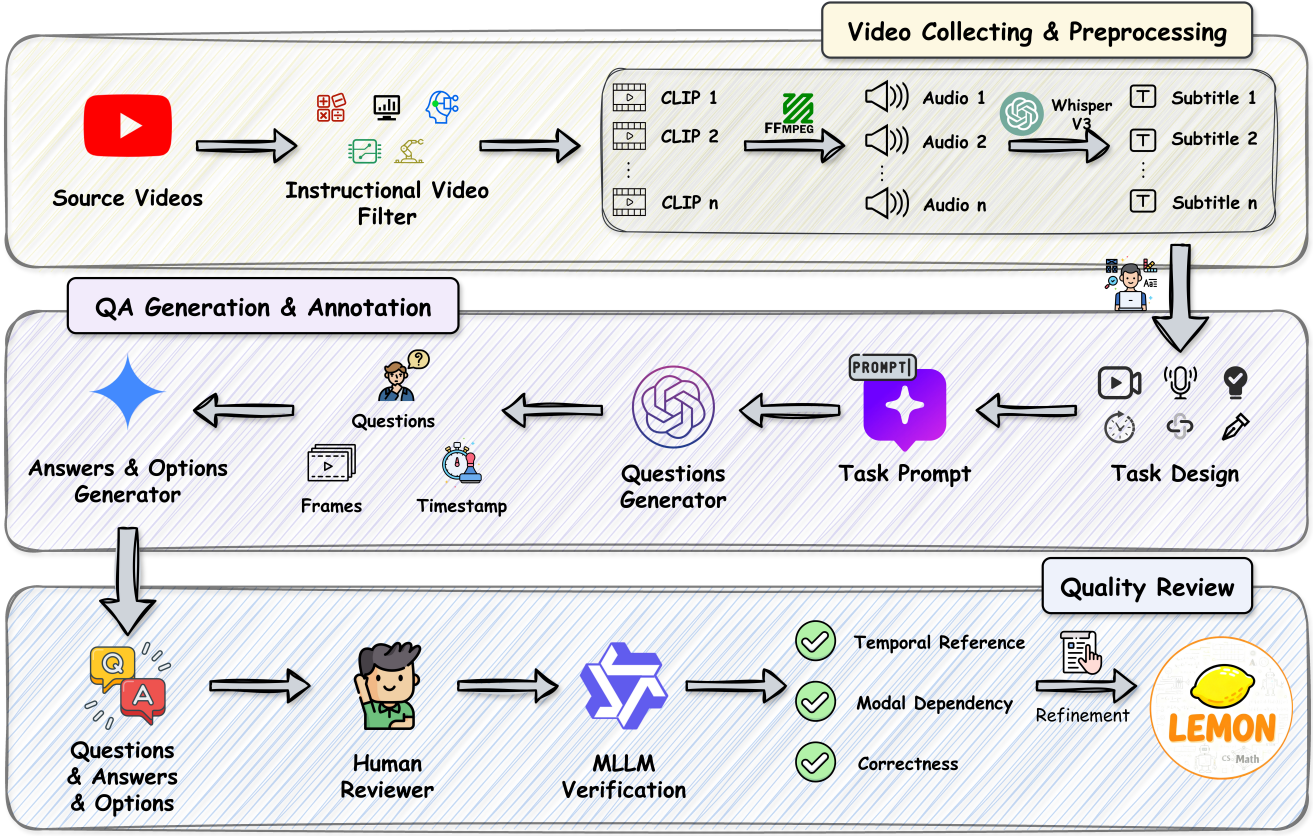}
    \caption{\textbf{The three-stage QA annotation process of LEMON.}}
    \label{fig: production process}
\end{figure*}

\begin{figure*}[t]
    \centering
    \includegraphics[width=\linewidth]{figures/guidelines.pdf}
    \caption{\textbf{Comprehensive guidelines for Generators and Annotators covering all six tasks in LEMON.}}
    \label{fig: guidelines}
\end{figure*}

\subsection{QA Generation and Annotation}
Figure \ref{fig: production process} illustrates the three-stage data construction workflow of LEMON, while Figure \ref{fig: guidelines} further shows the specific guidelines provided to generators and annotators for all six major tasks in LEMON.

The following sections detail the task-specific pipelines, outlining how multimodal signals are extracted, processed, and annotated to ensure consistent and high-quality supervision across the benchmark.

\subsubsection{Streaming Perception (SP)}
For this task, we segment each instructional video into clips of up to ten minutes and pair them with time-aligned subtitles. These multimodal segments are then provided to GPT-4o to generate a sequence of perception-oriented questions. To reflect realistic streaming comprehension, GPT-4o is explicitly instructed to ensure that consecutive questions exhibit semantic continuity and causal dependence, and each new question must build directly on the answer to the previous one rather than starting from an unrelated visual event.

Subsequently, human annotators review generated question sequences to verify logical coherence, the validity of causal links, and proper temporal grounding within the video segment. Annotators revise or discard questions that break the dependency chain, ensuring that Streaming Perception ultimately evaluates models on incremental understanding under temporally unfolding instructional content. Finally, hand over the handled issues to Gemini for generating answers and options.

\subsubsection{OCR-Based Reasoning (OR)}
For this task, annotators first watch all instructional videos and identify segments that contain example questions or homework problems. They then record the precise time spans in which the instructor presents the problem statement, but has not yet begun the solution. Based on these timestamps, we extract the corresponding video clips and select a key frame that fully contains the problem text. More importantly, annotators are required to draw bounding boxes around all relevant textual regions to ensure complete and accurate OCR grounding.

Using the extracted problem statement and its correct solution, we construct a multiple-choice question by adding carefully designed distractor options. These distractors are generated to be plausible, typically reflecting common mistakes, misapplied formulas, or visually similar answer patterns, so that the task evaluates a model’s ability to both read the textual content and reason about it, rather than relying on superficial cues.

\subsubsection{Audio Comprehension (AC)}
For this task, annotators first scan each video’s audio and subtitle track to locate short segments with clear speech and minimal background noise. Centered on each selected audio snippet, they extract a five-minute surrounding video window and feed both the video and aligned subtitles into GPT-4o to generate \textit{Contextual Listening} questions together with the corresponding timestamps. Based on these timestamps, annotators then return to the video and select four frames that exhibit distinct, visually discriminative cues, ensuring that the final multimodal item couples audio understanding with grounded visual evidence.

\subsubsection{Temporal Awareness (TA)}
For this task, annotators watch each full video and select three to four segments, each lasting 20 to 40 seconds, that exhibit clear temporal structure. These chosen segments must display coherent visual and textual continuity, such as enumerated summaries, step-wise derivations, or the instructional progression from concept explanation to worked examples, so that the resulting clips contain explicit cues for temporal reasoning. Through this selection strategy, the task evaluates a model’s ability to trace sequential logic rather than interpret isolated moments.

\subsubsection{Instructional Prediction (IP)}
For this task, annotators focus on segments that contain substantive instructional content, such as concept explanations, formula definitions, or worked-through derivations, ensuring that each selected clip provides enough information for answering the primary question. Crucially, the clip must also support a second, forward-looking question that tests whether a model can predict the next instructional step. To enable this, annotators prioritize segments with strong pedagogical logic: transitions where an instructor naturally progresses from definition to application, from theorem to proof sketch, or from problem setup to solution strategy. Even without relying on visuals, the combination of subtitles and audio should provide sufficient cues for a model to anticipate what the instructor will introduce next, making the task a direct evaluation of multimodal and discourse-level instructional forecasting.

\subsubsection{Advanced Expression (AE)}
For this task, annotators deliberately select segments with high conceptual density, such as lectures rich in technical terminology, theoretical explanations, or domain-specific constructs, while filtering out low-information content, such as exercise walkthroughs or coding demos. The selected video clips and their aligned transcripts are then provided to both GPT-4o and Gemini 2.0 Flash, each of which is prompted to extract a set of key domain terms along with their correct multilingual expressions. Reviewers subsequently consolidate, refine, and validate the outputs from both models, ensuring that only accurate, unambiguous, and domain-faithful terminology is retained as the final evaluation labels. This process yields a rigorous benchmark for assessing the model’s ability to produce fine-grained content, cross-lingual translation, and interpret advanced technical expressions.

\section{Additional Dataset Statistics} \label{sec: Additional Dataset Statistics}
\begin{table}[t]
    \centering
    \resizebox{0.9\linewidth}{!}{
    \begin{tabular}{lcc}
        \toprule
        \midrule
        \textbf{Subject} & \textbf{Number} & \textbf{QA Tokens} \\
        \midrule
        \rowcolor{orange!10}
        \multicolumn{3}{c}{\textbf{\textit{Streaming Perception (SP)}}} \\
        \midrule
        \textbf{Mathematics} & 322 & 40.82\\
        \textbf{Artificial Intelligence} & 226 & 38.61 \\
        \textbf{Computer Science} & 295 & 38.29\\
        \textbf{Electronic Engineering} & 267 & 43.89\\
        \textbf{Robotics} & 175 & 42.88\\
        \midrule
        \rowcolor{purple!10}
        \multicolumn{3}{c}{\textbf{\textit{OCR-Based Reasoning (OR)}}} \\
        \midrule
        \textbf{Mathematics} & 294 & 114.86 \\
        \textbf{Computer Science} & 310 & 101.99 \\
        \textbf{Electronic Engineering} & 202 & 103.77 \\
        \midrule
        \rowcolor{green!10}
        \multicolumn{3}{c}{\textbf{\textit{Audio Comprehension (AC)}}} \\
        \midrule
        \textbf{Mathematics} & 176 & 191.78 \\
        \textbf{Artificial Intelligence} & 276 & 170.93 \\
        \textbf{Robotics} & 126 & 195.67 \\
        \midrule
        \rowcolor{red!10}
        \multicolumn{3}{c}{\textbf{\textit{Temporal Awareness (TA)}}} \\
        \midrule
        \textbf{Mathematics} & 45 & 312.04 \\
        \textbf{Artificial Intelligence} & 141 & 309.56 \\
        \textbf{Electronic Engineering} & 42 & 313.92\\
        \textbf{Robotics} & 82 & 313.93\\
        \midrule
        \rowcolor{blue!10}
        \multicolumn{3}{c}{\textbf{\textit{Instructional Prediction (IP)}}} \\
        \midrule
        \textbf{Artificial Intelligence} & 434 & 42.88 \\
        \midrule
        \rowcolor{pink!10}
        \multicolumn{3}{c}{\textbf{\textit{Advanced Expression (AE)}}} \\
        \midrule
        \textbf{Mathematics} & 160 & 12.50 \\
        \textbf{Artificial Intelligence} & 144 & 12.50 \\
        \textbf{Computer Science} & 176 & 12.50 \\
        \textbf{Electronic Engineering} & 112 & 12.50 \\
        \textbf{Robotics} & 176 & 12.50 \\
        \midrule
        \bottomrule
    \end{tabular}
    }
    \caption{\textbf{Distribution across five STEM subjects for all six tasks in LEMON.} \textbf{QA Tokens} represents the average token count in QA pairs. The variation in counts reflects task-specific modality requirements and the differing availability of suitable instructional content across subjects.}
    \label{tab: distribution across five STEM}
\end{table}

\begin{figure}
    \centering
    \includegraphics[width=\linewidth]{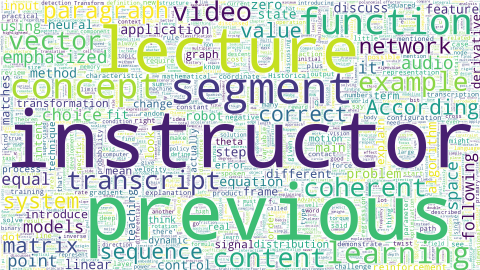}
    \caption{\textbf{Word cloud of QA content across LEMON.} The visualization highlights the most frequent terms appearing in QA pairs across six tasks.}
    \label{fig: wordcloud}
\end{figure}

\begin{table*}[t]
    \centering 
    \resizebox{0.85\linewidth}{!}{
    \begin{tabular}{l cccccccc}
        \toprule
        \midrule
        \multirow{2}{*}{\textbf{Models}} & 
        \multirow{2}{*}{\textbf{Hardware}} & 
        \multirow{2}{*}{\textbf{Precision}} &
        \multicolumn{6}{c}{\textbf{Runtime (s)}} \\
        \cmidrule(lr){4-9}
        & & & SP & OR & AC & TA & IP & AE \\
        \midrule
        IXC2.5-OL & A100-SXM4-80GB & fp16 & 32.39 & 15.74 & \textbf{5.32} & 12.52 & 23.34 & 9.60 \\
        Baichuan-Omni 1.5 & A100-SXM4-80GB & fp16 & \textbf{296.03} & \textbf{21.78} & 1.03 & 5.36 & \textbf{66.07} & \textbf{56.00} \\
        MiniCPM-o 2.6 & A100-SXM4-80GB & fp16 & 24.13 & 10.13 & 1.44 & 9.62 & 13.56 & 10.97 \\
        M4 & A100-PCIE-40GB & fp16 & 4.66 & 3.05 & 0.75 & 23.65 & 3.83 & 10.77 \\
        Ola & A100-PCIE-40GB & fp16 & 5.73 & 3.39 & 0.59 & 5.15 & 4.87 & 7.87 \\
        Video-SALMONN 2+ & A800-80GB & fp16 & 11.97 & 2.19 & 3.40 & \textbf{29.40} & 5.66 & 8.44 \\
        LLaVA-OneVision & A100-PCIE-40GB & fp16 & 17.93 & 6.82 & 0.61 & 12.75 & 10.57 & 19.46 \\
        Dispider & A100-PCIE-40GB & fp16 & 45.90 & 4.69 & 0.84 & 15.89 & 13.71 & 1.38 \\
        InternVL3 & A100-PCIE-40GB & fp16 & 18.69 & 5.38 & 0.57 & 9.26 & 11.23 &  16.46 \\
        LongVA & A100-PCIE-40GB & fp16 & 7.53 & 5.61 & 0.48 & 5.49 & 5.43 & 23.34 \\
        LongVU & A100-PCIE-40GB & fp16 & 64.74 & 6.67 & 0.95 & 16.77 & 20.67 & 8.59 \\
        ShareGPT4Video & A100-PCIE-40GB & fp16 & 17.18 & 6.30 & 0.84 & 10.54 & 9.33 & 8.38 \\
        VideoLLaMA3 & A100-PCIE-40GB & fp16 & 17.64 & 4.46 & 2.84 & 3.84 & 9.71 & 13.37 \\
        Video-XL-2 & A100-PCIE-40GB & fp16 & 90.43 & 15.45 & 0.71 & 6.14 & 16.53 & 11.69 \\
        \midrule
        \bottomrule
    \end{tabular}
    }
    \caption{\textbf{Average inference time per QA pair for each model across tasks.} 
    Open-source models are deployed locally and run with fp16 precision. The table reports the mean processing time (in seconds) for different models across six task categories.}
    \label{tab: inference time on each task}
\end{table*}

Table \ref{tab: distribution across five STEM} summarizes the distribution of samples across the five STEM subjects and their average QA token lengths within LEMON. It is worth noting that the average QA length in \textbf{OCR-Based Reasoning} is driven primarily by the inherent length of in-class problem statements. For \textbf{Audio Comprehension}, token counts reflect the four transcribed candidate options. The higher token length in \textbf{Temporal Awareness} arises from UUID-based clip identifiers used in the options. In contrast, \textbf{Advanced Expression} yields very short QA pairs because the model is only given a query, no options or answers are required.

Figure \ref{fig: wordcloud} presents a word cloud constructed from all QA pairs in LEMON. The dominant terms correspond to high-frequency instructional concepts such as definitions, examples, and problem-solving steps. Larger tokens reflect commonly referenced entities (e.g., variables, formulas, diagram elements), while long-tail distribution demonstrates the diversity of domain-specific terminology across the five STEM subjects. This linguistic profile aligns with LEMON’s goal of capturing authentic multimodal classroom discourse and supporting evaluations that require both semantic understanding and pedagogical reasoning.

\section{Detailed Experimental Settings} \label{sec: Detailed Experimental Settings}
\subsection{Implement Details}

This section provides a detailed breakdown of computational resources, runtime characteristics, and modality-handling strategies across MLLMs. All evaluations are conducted under a consistent hardware environment unless otherwise specified.

\subsubsection{Input Preprocessing}
We adopt consistent preprocessing strategies across all models to ensure comparability, while distinguishing between streaming and non-streaming video input modes.

\begin{itemize}
    \item \textbf{Video Frames}: Streaming-capable models process video input sequentially at 1 FPS, whereas non-streaming models operate on 64 sampled frames or follow their native frame-selection strategy when specified. For Open-Source MLLMs, frames are resized to 224$\times$224 and then sent into the image encoder; For Proprietary MLLMs, frames are adjusted to 512$\times$512 and then converted to base64 format.

    \item \textbf{Audio}: For Proprietary MLLMs and Open-Source Omni MLLMs with audio capability, we provide resampled 16 kHz audio with loudness and channel normalization. 

    \item \textbf{Subtitle}: Subtitle inputs are tokenized using the model-specific tokenizer and punctuation is normalized.
\end{itemize}

\subsubsection{Model Deployment and Evaluation}

As shown in Table \ref{tab: inference time on each task}, we summarize the deployment and evaluation strategies for open-source models and report the average runtime for each task type.

\begin{itemize}
    \item \textbf{Proprietary MLLMs}: We evaluate these models via API calls, thus internal GPU/CPU usage not directly measurable.

    \item \textbf{Open-Source MLLMs}: All open-source models are executed locally in fp16 precision. Depending on the model’s computational footprint and recommended deployment settings, inference is conducted on either an A100 40GB/80GB GPU or an A800 80GB GPU. GPU memory usage and runtime behavior are monitored consistently across all evaluations. 
    
\end{itemize}

\subsection{Evaluation Metrics}


Our evaluation is tailored to the diverse task formats in LEMON, ensuring consistent and reliable assessment across all settings. For multiple-choice tasks, model performance is assessed directly using accuracy. For open-ended tasks, where responses may vary in phrasing or granularity, we leverage the curated reference labels and employ GPT-based scoring to assess semantic alignment and correctness.

\subsubsection{Multiple-Choice Tasks}
Each QA pair contains a model-generated response and a ground-truth answer label. Evaluation is conducted using a two-stage matching pipeline.

\textbf{Direct Answer Matching}. To standardize option predictions across heterogeneous model outputs, we apply a unified rule-based extractor that normalizes responses into one of the four discrete choices (A, B, C or D). 
As summarized in Table \ref{tab: output format}, the extractor covers the most common surface forms produced by MLLMs, ranging from simple letter outputs to descriptive patterns and stylistic variations.
When none of these textual signatures is detected, the raw response is preserved and passed to the LLM-based extraction stage. 

\begin{table}[h]
    \centering
    \resizebox{\linewidth}{!}{
    \begin{tabular}{llc}
        \toprule
        \midrule
        \textbf{Output Format} & \textbf{Example} & \textbf{Matchable} \\
        \midrule
        
        Single Letter Output & A & \textcolor{teal}{\usym{2713}} \\
        Parenthesized Letter & (A) & \textcolor{teal}{\usym{2713}} \\
        Prefixed Letter Format & A:, A), A1 & \textcolor{teal}{\usym{2713}} \\
        Descriptive Pattern & Answer: ..., Answer is... & \textcolor{teal}{\usym{2713}} \\
        Sentence-Initial Marker & A. \{OPTION CONTENT\} & \textcolor{teal}{\usym{2713}} \\
        Sentence-Final Marker & After ..., the correct answer is D. & \textcolor{teal}{\usym{2713}} \\
        Natural Language Response & I think ... & \textcolor{Red}{\usym{2717}} \\
        \midrule
        \bottomrule
    \end{tabular}
    }
    \caption{\textbf{Common output patterns produced by evaluated MLLMs.}}
    \label{tab: output format}
\end{table}

\textbf{LLM-Assisted Extraction}. When a model's output is written in free-form natural language and does not explicitly state a discrete option, we employ GPT-4o to infer the most probable choice from the predefined label set. If the extractor fails to map the response to a valid option, the prediction is marked as incorrect. The prompt for the LLM to extract the answer is as follows:

\begin{tcolorbox}
    [colback=Emerald!10,
    colframe=cyan!40!black,
    title=\textbf{Prompt for Answer Extractor}]
    
    You are an answer extraction assistant. Please extract the answer choice (A, B, C, or D) from the following response. 
    Return \textbf{ONLY} a single letter or \textbf{INVALID} if no clear answer can be determined.

    \textbf{Response}: {response}

    \textbf{The extracted answer is}:
\end{tcolorbox}

\textbf{Accuracy Calculation.} All successfully identified labels are normalized to uppercase before evaluation. The final accuracy for each task is computed as

\begin{equation}
    Accuracy = \frac{1}{N}\sum^N_{i=1}1\{{\hat{y}_i=y_i}\},
\end{equation}

where $\hat{y}_i$ denotes the extracted prediction and $y_i$ is the corresponding ground-truth answer.

\subsubsection{Open-Ended Tasks: VS Evaluation}
For \textit{Video Summarization}, we employ a hybrid evaluation protocol that integrates GPT-based holistic scoring with keyword-level semantic matching. 

\textbf{GPT-Based Holistic Scoring.} GPT-4o is used as an expert evaluator to score each generated summary on a 0–10 scale according to four predefined dimensions. And the GPT score for sample $i$ is denoted as:
\begin{equation}
    S^{\text{GPT}}_i \in [0, 10].
\end{equation}
A detailed scoring prompt used for this evaluation is shown below.

\begin{tcolorbox}
[breakable, colback=SeaGreen!10!CornflowerBlue!10,
colframe=RoyalPurple!55!Aquamarine!100!,
title=\textbf{Prompt for GPT Scoring on VS}]

You are an evaluation assistant. Please score the following summary on a scale of 0-10.

\# \textbf{\textit{Scoring Criteria (total 10 points)}}:

1. \textbf{Keyword Coverage} (4 points): How many key terms from the reference list are mentioned or covered in the summary. More keywords = higher score.

2. \textbf{Content Completeness} (3 points): Does the summary comprehensively cover the main content and concepts?

3. \textbf{Language Fluency} (2 points): Is the summary well-written, coherent, and fluent? Are sentences grammatically correct?

4. \textbf{Structure and Organization} (1 point): Is the summary well-structured and logically organized?

\# \textbf{\textit{Detailed Scoring Standards}}:

- 0 points: Completely irrelevant, no keywords covered, or empty response

- 1–3 points: Very poor quality – $<$30\% keywords covered, major content missing, poor language, unclear structure

- 4–6 points: Average quality – 30-60\% keywords covered, partial content, acceptable language, basic structure

- 7–9 points: Good quality – 60–90\% keywords covered, most content present, fluent language, clear structure

- 10 points: Excellent quality – $>$90\% keywords covered, fully comprehensive, excellent fluency, strong organization

\textbf{Reference Keywords}: \{keywords\}

\textbf{Summary to Evaluate}: \{response\}

\textbf{Please provide: Score}: \{number\}

\end{tcolorbox}

\textbf{Keyword Matching.} To measure factual grounding, we compute the number of reference key terms explicitly mentioned in the generated summary. Let $\mathcal{K}$ be the set of reference keywords, and $\text{resp}_i$ be the model-generated summary. Then the keyword match score is:
\begin{equation}
    S^{\text{KW}}_i = \bigl|\{ k \in \mathcal{K} \mid k \text{ appears in } \text{resp}_i \}\bigr|.
\end{equation}

\textbf{Final Video Summarization Score.} We combine the two components through a weighted sum that prioritizes holistic assessment while retaining factual rigor:
\begin{equation}
    \text{VS-Score}_i
        = W_{\text{GPT}} \cdot 10 \cdot S^{\text{GPT}}_i
        + W_{\text{KW}} \cdot 100 \cdot S^{\text{KW}}_i,
\end{equation}
where we set $W_{\text{GPT}}$ to 0.7 and $W_{\text{KW}}$ to 0.3.

\subsubsection{Open-Ended Tasks: KT Evaluation}
Knowledge Translation is evaluated using two complementary methods: GPT-based scoring and automatic translation metrics.

\textbf{GPT-Based Holistic Scoring.}GPT-4o serves as an expert evaluator, assigning a score from 0 to 10 based on three principles:
\begin{itemize}
    \item \textbf{Accuracy:} Correctness of key term translations.
    \item \textbf{Fluency:} Naturalness and readability of the translation.
    \item \textbf{Elegance:} Idiomatic and elegant expression.
\end{itemize}
And the score from GPT is denoted as:
\begin{equation}
    S^{\text{GPT}}_i \in [0, 10].
\end{equation}
The detailed evaluation prompt is as follows:

\begin{tcolorbox}
[breakable, colback=Salmon!20, 
colframe=Salmon!90!Black,
title = \textbf{Prompt for GPT Scoring on KT}, 
]

You are an evaluation assistant. Please score the following Chinese translation on a scale of 0-10.

\# \textbf{\textit{Scoring Criteria}}:

1. \textbf{Accuracy} (4 points): Are the key terms translated correctly? Check if the translation matches the reference key terms.

2. \textbf{Fluency} (3 points): Is the translation fluent and natural in Chinese? Does it read smoothly?

3. \textbf{Elegance} (3 points): Is the translation elegant and idiomatic? Does it follow Chinese expression conventions?

\# \textbf{\textit{Detailed Scoring Standards}}:

- 0 points: Completely incorrect translation, no key terms correct, or empty response

- 1-3 points: Very poor quality - $<$ 30\% key terms correct, many translation errors, unnatural Chinese, poor readability

- 4-6 points: Average quality - 30-60\% key terms correct, some translation errors but generally understandable, acceptable Chinese fluency, basic readability

- 7-9 points: Good quality - 60-90\% key terms correct, minor translation errors, fluent and natural Chinese, good readability and style

- 10 points: Excellent quality - $>$ 90\% key terms correct, accurate translation, perfect Chinese fluency, elegant and idiomatic expression

\textbf{Reference Keywords}: \{keywords\}

\textbf{Reference Full Translation}: \{reference full translation\}

\textbf{Translation to Evaluate}: \{response\}

\end{tcolorbox}

\textbf{Automatic Translation Metrics.} Meanwhile, we also evaluated four translation metrics: BLEU \footnote{\url{https://github.com/mjpost/sacrebleu}}, ROUGE \footnote{\url{https://pypi.org/project/rouge-score}}, BERTScore \footnote{\url{https://github.com/Tiiiger/bert_score}}, and COMET22 \footnote{\url{https://huggingface.co/Unbabel/wmt22-comet-da}}.

\textbf{BLEU} evaluates n-gram precision between the candidate translation and reference:
\begin{equation}
\text{BLEU} = \text{BP} \cdot \exp\Big( \sum_{n=1}^N w_n \log p_n \Big),
\end{equation}
where $p_n$ is the precision of n-grams, $w_n$ are the weights (usually uniform), and BP is the brevity penalty:
\begin{equation}
\text{BP} = 
\begin{cases}
1, & \text{if } c > r,\\
e^{1 - r/c}, & \text{if } c \le r,
\end{cases}
\end{equation}
with $c$ the length of candidate translation and $r$ the reference length.

\textbf{ROUGE-L} measures the longest common subsequence (LCS) overlap between candidate $C$ and reference $R$:
\begin{equation}
\text{ROUGE-L} = \frac{(1+\beta^2) \cdot \text{LCS}(C,R)}{|C| + \beta^2 |R|},
\end{equation}
where $|C|$ and $|R|$ are lengths of candidate and reference, and $\beta$ balances precision and recall.

\textbf{BERTScore} computes semantic similarity using contextual embeddings:
\begin{equation}
\text{BERTScore} = \frac{1}{|C|} \sum_{c_i \in C} \max_{r_j \in R} \cos(\mathbf{e}_{c_i}, \mathbf{e}_{r_j}),
\end{equation}
where $\mathbf{e}_{c_i}$ and $\mathbf{e}_{r_j}$ are contextual embeddings for candidate token $c_i$ and reference token $r_j$.

\textbf{COMET22} is a reference-based learned metric that predicts translation quality using pretrained cross-lingual models. Given source $S$, candidate $C$, and reference $R$, the COMET22 score is:
\begin{equation}
\text{COMET22}(S, C, R) = f_{\text{COMET}}(S, C, R),
\end{equation}
where $f_{\text{COMET}}$ is the pretrained neural regressor that outputs a quality score reflecting adequacy, fluency, and semantic similarity.

\begin{figure}[t]
    \centering
    \includegraphics[width=\linewidth]{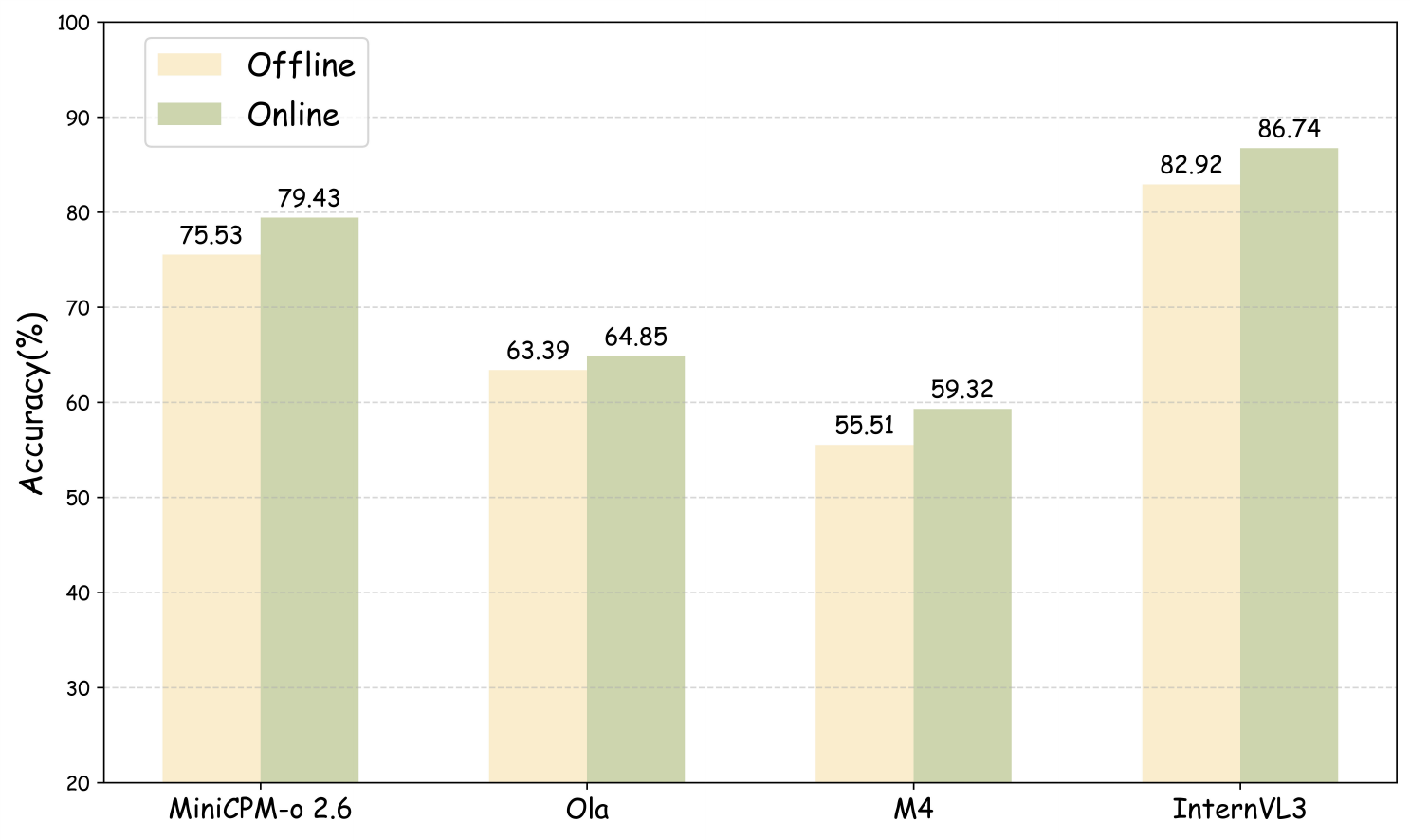}
    \caption{\textbf{Performance Comparison between Online and Offline Mode.}}
    \label{fig: Online vs Offline}
\end{figure}

\begin{figure*}[t]
    \centering
    \includegraphics[width=0.8\linewidth]{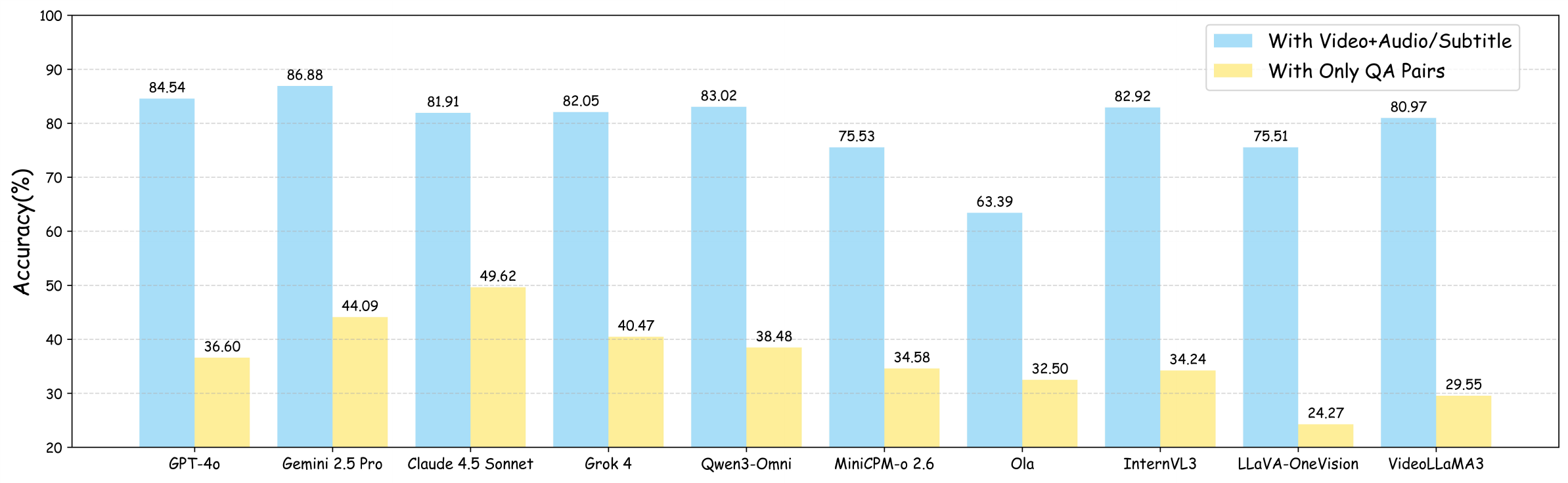}
    \caption{\textbf{Streaming Perception with and without Multimodal Inputs.} Performance comparison between full multimodal inputs (video + audio/subtitles) and only the QA pair, showing how much the models rely on their inherent prior knowledge without external signals.}
    \label{fig: SP Prior Knowledge Probing}
\end{figure*}

\begin{figure*}[t]
    \centering
    \begin{subfigure}[t]{0.33\linewidth}
        \centering
        \includegraphics[width=\linewidth]{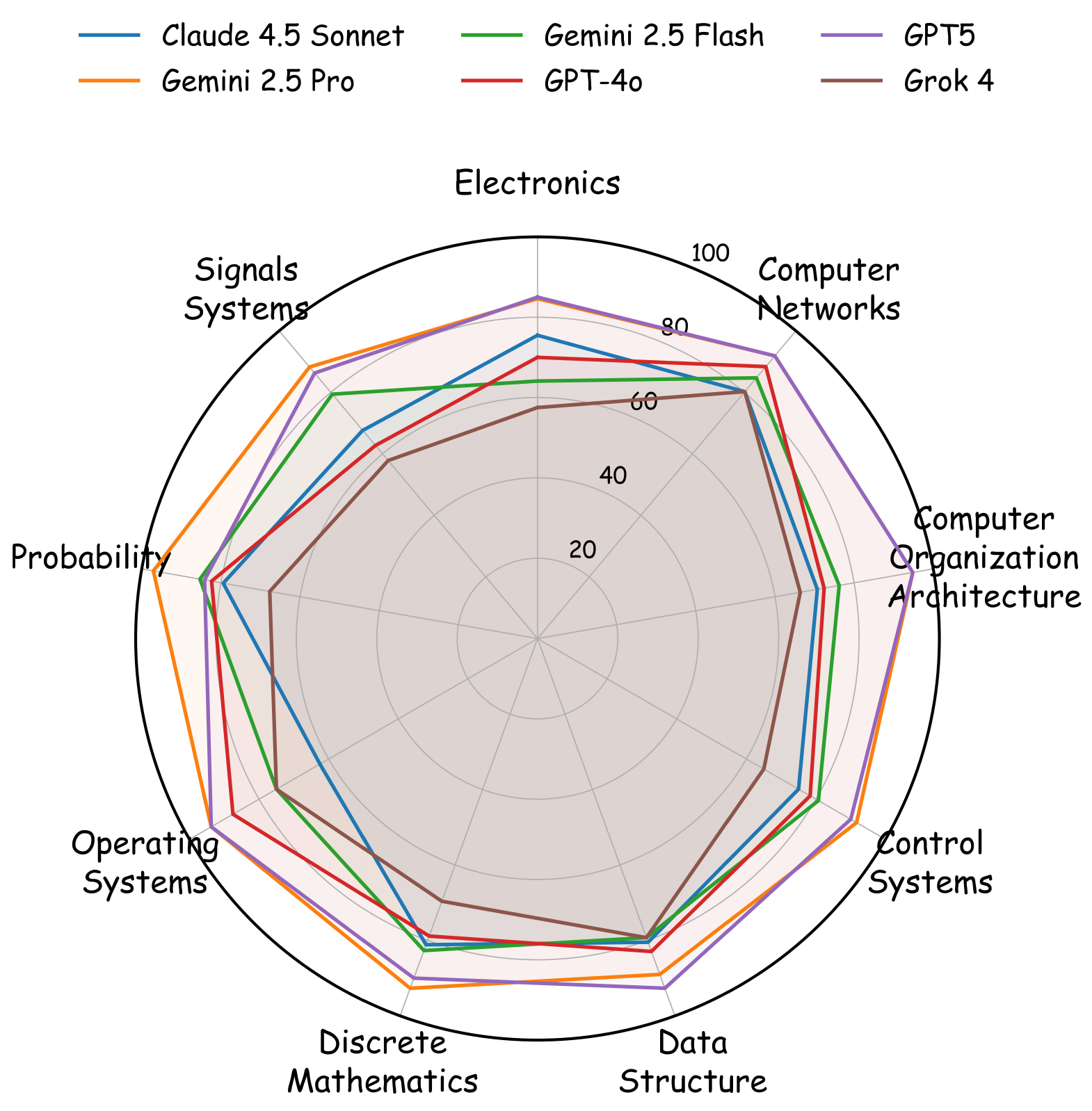}
        \caption{Proprietary MLLMs.}
        \label{subfig: close mllms}
    \end{subfigure}
    \hfill
    \begin{subfigure}[t]{0.33\linewidth}
        \centering
        \includegraphics[width=\linewidth]{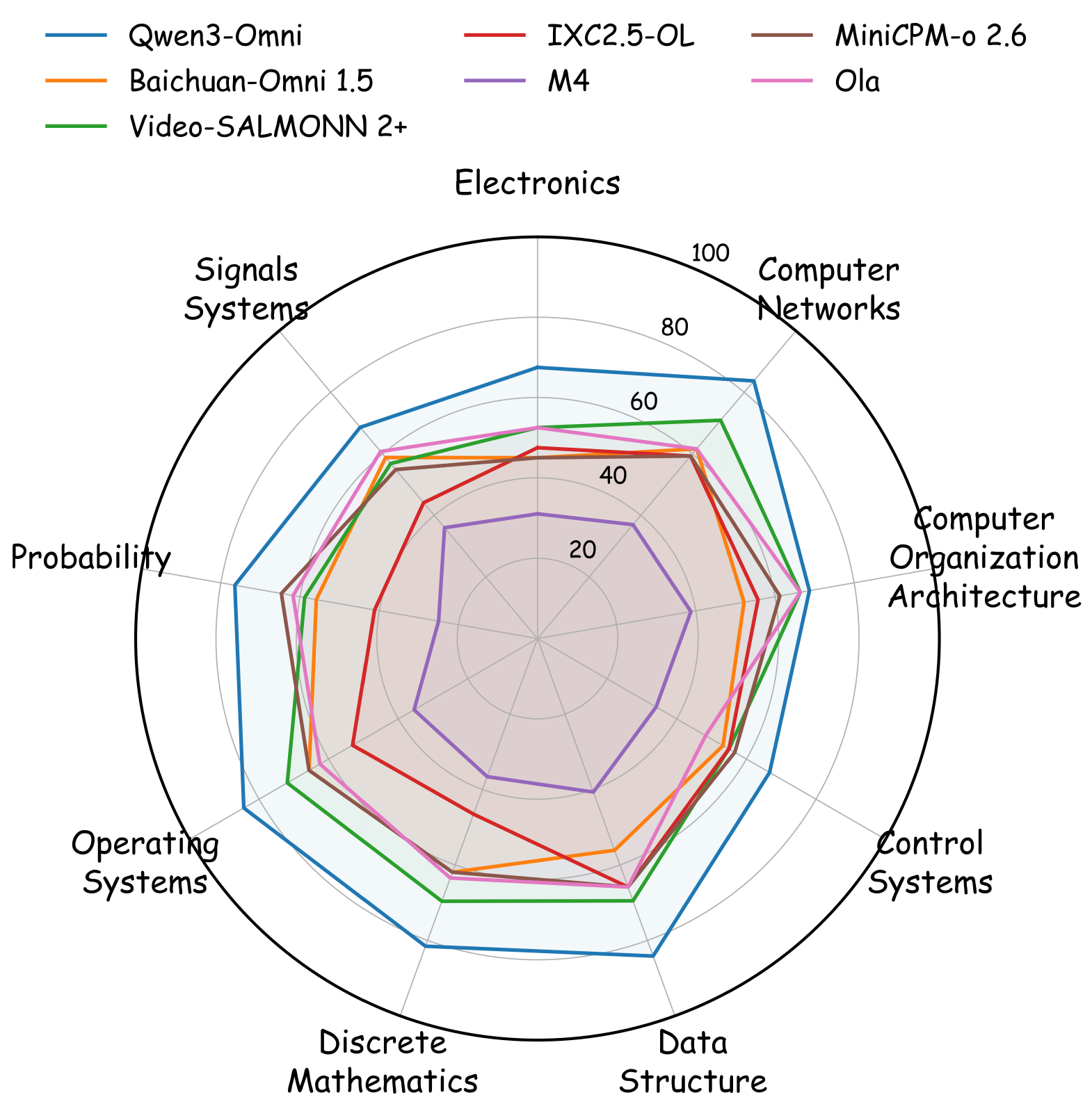}
        \caption{Open-Source Omni MLLMs.}
        \label{subfig: omni mllms}
    \end{subfigure}
    \hfill
    \begin{subfigure}[t]{0.33\linewidth}
        \centering
        \includegraphics[width=\linewidth]{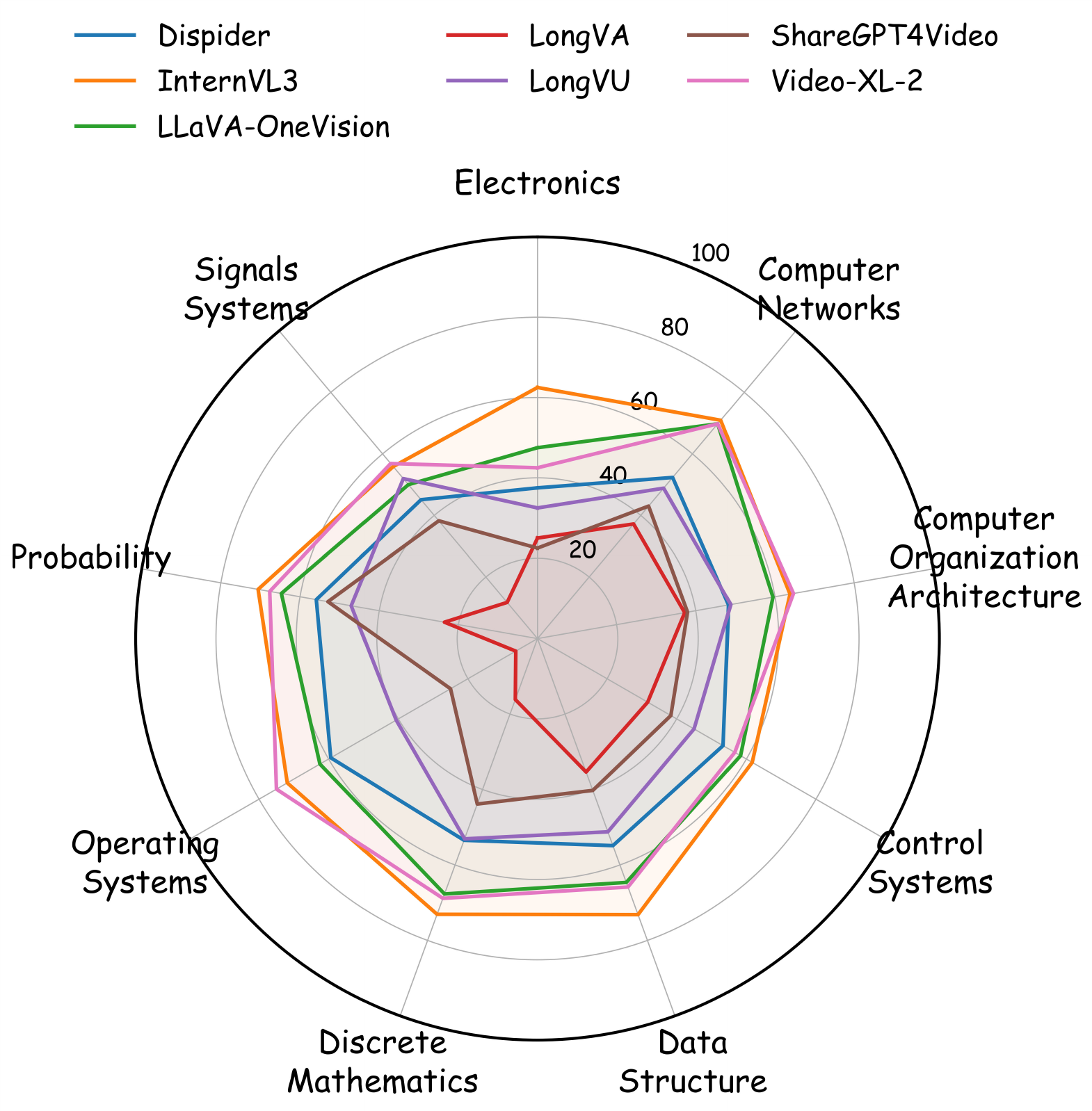}
        \caption{Open-Source Video MLLMs.}
        \label{subfig: video mllms}
    \end{subfigure}
    \caption{\textbf{Performance comparison on OCR-Based Reasoning across nine university courses.}}
    \label{fig: different class}
\end{figure*}

\section{Further Experiments} \label{sec: Further Experiments}

To obtain a more comprehensive understanding of model behavior beyond the main results, we conduct a series of additional experiments. All experiments follow the same computational setup and evaluation protocol as described in the main sections.

\subsection{Online vs. Offline Inference}

Current MLLMs provide limited support for true online (streaming) inference, while most models still rely on offline processing pipelines that have access to the full video segment before generating predictions.
In fact, Online inference is implemented by providing the model only with video frames observed up to the current timestamp for each QA pair, ensuring that no future information is accessible when generating the answer. In Offline mode, the model has full access to the entire video, allowing answers to leverage the complete temporal context.

To examine whether existing ``online-capable'' models exhibit meaningful advantages in Streaming Perception, we compare several representative MLLMs under both online and offline settings, as shown in Figure \ref{fig: Online vs Offline}. Despite their nominal streaming functionality, our evaluation shows that the performance gap between online and offline modes remains minimal across tested models. This observation suggests that current MLLMs have yet to achieve genuine real-time understanding, and significant progress is still required before online multimodal reasoning can become reliably effective.

\begin{table*}[t]
    \centering
    \begin{tabular}{l ccc cc}
        \toprule
        \midrule
        \multirow{2}{*}{\textbf{Models}} & 
        \multicolumn{3}{c}{\textbf{Temporal Awareness}} & 
        \multicolumn{2}{c}{\textbf{Video Summarization}} \\

        \cmidrule(lr){2-4} \cmidrule(lr){5-6}
        & Video & +Subtitle & +Audio & Video & +Subtitle \\ \midrule

        GPT-4o & 29.03 & 29.68$_{\textcolor{teal}{\uparrow 0.65}}$ & - & 52.39 & 50.19$_{\textcolor{Red}{\downarrow 2.20}}$ \\
        GPT-5 & 23.71 & 31.29$_{\textcolor{teal}{\uparrow 7.58}}$ & - & 55.90 & 58.76$_{\textcolor{teal}{\uparrow 2.86}}$ \\
        Gemini 2.5 Pro & 22.58 & 35.48$_{\textcolor{teal}{\uparrow 12.90}}$ & 26.45$_{\textcolor{teal}{\uparrow 3.87}}$ & 55.74 & 57.76$_{\textcolor{teal}{\uparrow 2.02}}$ \\
        Gemini 2.5 Flash & 23.87 & 31.61$_{\textcolor{teal}{\uparrow 7.74}}$ & 23.30$_{\textcolor{Red}{\downarrow 0.57}}$ & 55.91 & 23.90$_{\textcolor{Red}{\downarrow 32.01}}$ \\
        Claude 4.5 Sonnet & 25.48 & 29.68$_{\textcolor{teal}{\uparrow 4.20}}$ & - & 53.14 & 60.49$_{\textcolor{teal}{\uparrow 7.35}}$ \\
        Grok 4 & 21.94 & 27.42$_{\textcolor{teal}{\uparrow 5.48}}$ & 20.32$_{\textcolor{Red}{\downarrow 1.62}}$ & \textbf{56.75} & \textbf{60.57}$_{\textcolor{teal}{\uparrow 3.82}}$ \\
        Qwen3-Omni & \textbf{65.81} & 61.29$_{\textcolor{Red}{\downarrow 4.52}}$ & - & 50.54 & 57.46$_{\textcolor{teal}{\uparrow 6.92}}$ \\
        MiniCPM-o 2.6 & 55.81 & \textbf{53.23}$_{\textcolor{Red}{\downarrow 2.58}}$ & \textbf{65.48}$_{\textcolor{teal}{\uparrow 9.67}}$ & 15.21 & 55.95$_{\textcolor{teal}{\uparrow 40.74}}$ \\
        Ola & 38.06 & 36.45$_{\textcolor{Red}{\downarrow 1.61}}$ & 37.79$_{\textcolor{Red}{\downarrow 0.27}}$ & 11.56 & 54.31$_{\textcolor{teal}{\uparrow 42.75}}$ \\
        M4 & 27.42 & 38.67$_{\textcolor{teal}{\uparrow 11.25}}$ & 38.74$_{\textcolor{teal}{\uparrow 11.32}}$ & 11.44 & 44.85$_{\textcolor{teal}{\uparrow 33.41}}$ \\
        InternVL3 & 23.23 & 25.58$_{\textcolor{teal}{\uparrow 2.35}}$ & - & 49.66 & 54.69$_{\textcolor{teal}{\uparrow 5.03}}$ \\
        LLaVA-OneVision & 24.19 & 28.39$_{\textcolor{teal}{\uparrow 4.20}}$ & - & 35.60 & 47.28$_{\textcolor{teal}{\uparrow 11.68}}$ \\
        VideoLLaMA3 & 24.52 & 28.39$_{\textcolor{teal}{\uparrow 3.87}}$ & - & 41.13 & 55.85$_{\textcolor{teal}{\uparrow 14.72}}$ \\
        \midrule
        \bottomrule
    \end{tabular}
    \caption{\textbf{Ablation study on whether subtitles or audio improve performance in Temporal Awareness and the \textit{Video Summarization} subtask of Advanced Expression.}}
    \label{fig: TA and AE}
\end{table*}

\begin{table}[t]
    \centering
    \resizebox{1.0\linewidth}{!}{
    \begin{tabular}{l cc}
    \toprule
    \midrule
    \textbf{Model} & \textbf{Self-Transcribed} & \textbf{Ground-Truth} \\
    \midrule
    Gemini 2.5 Pro & \textbf{79.92} & \textbf{86.85}$_{\textcolor{teal}{\uparrow 6.93}}$ \\
    Gemini 2.5 Flash & 79.22 & 85.12$_{\textcolor{teal}{\uparrow 5.9}}$ \\
    Grok 4 & 74.62 & 78.89$_{\textcolor{teal}{\uparrow 4.27}}$ \\
    Baichuan-Omni 1.5 & 27.34 & 36.23$_{\textcolor{teal}{\uparrow 8.89}}$ \\
    MiniCPM-o 2.6 & 42.56 & 54.67$_{\textcolor{teal}{\uparrow 12.11}}$ \\
    Ola & 31.83 & 29.41$_{\textcolor{Red}{\downarrow 2.42}}$ \\
    M4 & 48.35 & 54.60$_{\textcolor{teal}{\uparrow 6.25}}$ \\
    Video-SALMONN 2+ & 77.34 & 77.86$_{\textcolor{teal}{\uparrow 0.52}}$ \\
    \midrule
    \bottomrule
    \end{tabular}
    }
    \caption{\textbf{Comparison of audio comprehension accuracy using self-transcribed versus ground-truth transcripts.}}
    \label{tab: ac for audio mllms}
\end{table}

\begin{table*}
    \centering
    \resizebox{1.0\linewidth}{!}{
    \begin{tabular}{l cccc}
        \toprule
        \midrule
        
        \textbf{Models} & 
        \textbf{Arabic} &
        \textbf{Chinese} &
        \textbf{French} &
        \textbf{German} \\
        \midrule

        GPT-4o & 29.77 / 0.5460 / 0.7962 / 0.7436 & 7.04 / 0.6815 / 0.5943 / 0.8371 & \textcolor{red!60!black}{\textbf{56.86}} / 0.7163 / 0.7069 / 0.8153 & 43.31 / 0.5646 / 0.5527 / 0.7430 \\
        GPT-5 & 29.74 / 0.6885 / \textcolor{red!60!black}{\textbf{0.8850}} / \textcolor{red!60!black}{\textbf{0.8929}} & 5.67 / 0.7405 / 0.7270 / 0.8961 & 46.19 / 0.7417 / 0.7781 / 0.8769 & 42.06 / 0.6722 / 0.7496 / 0.8756 \\
        Gemini 2.5 Pro & 32.33 / 0.7175 / 0.8591 / 0.8866 & 4.84 / 0.8090 / 0.6743 / 0.8955 & 48.55 / 0.7840 / 0.7827 / \textcolor{red!60!black}{\textbf{0.8782}} & 44.29 / 0.7233 / 0.7362 / 0.8731 \\
        Gemini 2.5 Flash & \textcolor{red!60!black}{\textbf{34.29}} / 0.7093 / 0.8743 / 0.8905 & 2.78 / 0.7811 / 0.6871 / 0.8868 & 49.79 / \textcolor{red!60!black}{\textbf{0.7922}} / 0.8025 / 0.8754 & \textcolor{red!60!black}{\textbf{45.07}} / \textcolor{red!60!black}{\textbf{0.7333}} / 0.7530 / 0.8732 \\
        Claude 4.5 Sonnet & 31.28 / 0.7046 / 0.8668 / 0.8856 & 5.25 / 0.7728 / 0.6847 / 0.8959 & 49.21 / 0.7636 / 0.7657 / 0.8742 & 40.05 / 0.6720 / 0.7217 / 0.8680 \\
        Grok 4 & 32.66 / \textcolor{red!60!black}{\textbf{0.7233}} / 0.8746 / 0.8722 & 5.88 / 0.8204 / 0.7009 / 0.8904 & 47.38 / 0.7817 / 0.7753 / 0.8664 & 40.80 / 0.7040 / 0.7103 / 0.8648 \\
        Qwen3-Omni & 30.61 / 0.7116 / 0.8697 / 0.8785 & \textcolor{red!60!black}{\textbf{12.48}} / \textcolor{red!60!black}{\textbf{0.8418}} / \textcolor{red!60!black}{\textbf{0.7288}} / \textcolor{red!60!black}{\textbf{0.9038}} & 48.89 / 0.7798 / \textcolor{red!60!black}{\textbf{0.8131}} / 0.8754 & 42.89 / 0.7084 / \textcolor{red!60!black}{\textbf{0.7706}} / \textcolor{red!60!black}{\textbf{0.8768}} \\
        IXC2.5-OL & 0.00 / 0.0099 / 0.5594 / 0.2862 & 0.03 / 0.0101 / -0.1422 / 0.2992 & 0.00 / 0.0340 / -0.0907 / 0.3863 & 0.00 / 0.0151 / -0.1148 / 0.3256 \\
        Baichuan-Omni-1.5 & 1.54 / 0.0963 / 0.6639 / 0.5046 & 1.44 / 0.3632 / 0.3989 / 0.7292 & 10.73 / 0.2394 / 0.2234 / 0.6302 & 5.47 / 0.1517 / 0.1670 / 0.5807 \\
        MiniCPM-o 2.6 & 0.57 / 0.0807 / 0.6090 / 0.4139 & 1.15 / 0.2298 / 0.2163 / 0.6246 & 10.62 / 0.2714 / 0.2884 / 0.6429 & 4.27 / 0.1747 / 0.2015 / 0.6004 \\
        Ola & 0.08 / 0.0490 / 0.6043 / 0.4042 & 0.25 / 0.0747 / 0.1962 / 0.5584 & 0.18 / 0.1011 / 0.0964 / 0.5487 & 0.12 / 0.0782 / 0.0879 / 0.5236 \\
        M4 & 0.02 / 0.0434 / 0.4461 / 0.3611 & 0.45 / 0.0995 / 0.1667 / 0.5631 & 0.53 / 0.1161 / 0.0574 / 0.5346 & 0.56 / 0.0897 / 0.0821 / 0.4892 \\
        video SALMONN 2+ & 2.40 / 0.2582 / 0.7054 / 0.6363 & 4.55 / 0.4272 / 0.4957 / 0.7670 & 4.27 / 0.2499 / 0.1590 / 0.6656 & 6.07 / 0.2491 / 0.2588 / 0.6643 \\
        ShareGPT4Video & 0.00 / 0.0383 / 0.5942 / 0.3750 & 0.01 / 0.1278 / 0.1959 / 0.6235 & 0.05 / 0.1187 / 0.1140 / 0.4976 & 0.06 / 0.0401 / 0.0523 / 0.4565 \\
        LongVA & 1.17 / 0.1215 / 0.6787 / 0.5341 & 3.78 / 0.3700 / 0.4557 / 0.7590 & 4.39 / 0.2887 / 0.2642 / 0.6330 & 5.51 / 0.2392 / 0.2927 / 0.6314 \\
        LLaVA-OneVision & 0.75 / 0.0610 / 0.6430 / 0.4550 & 0.49 / 0.0971 / 0.1048 / 0.5878 & 3.62 / 0.1546 / 0.1712 / 0.5761 & 2.01 / 0.1152 / 0.1437 / 0.5440 \\
        LongVU & 0.00 / 0.0212 / 0.5921 / 0.3616 & 0.11 / 0.0382 / 0.1198 / 0.5797 & 0.58 / 0.0307 / -0.0072 / 0.4748 & 0.13 / 0.0224 / -0.0070 / 0.4271 \\
        VideoLLaMA3 & 0.18 / 0.1365 / 0.5444 / 0.5002 & 1.75 / 0.1757 / 0.1867 / 0.6372 & 3.63 / 0.2066 / 0.0277 / 0.6026 & 0.35 / 0.0885 / -0.1857 / 0.4999 \\
        InternVL3 & 0.24 / 0.1486 / 0.6588 / 0.6306 & 1.84 / 0.3469 / 0.4180 / 0.7777 & 6.43 / 0.2808 / 0.2225 / 0.7289 & 7.99 / 0.2628 / 0.2653 / 0.7210 \\
        Video-XL-2 & 0.58 / 0.1103 / 0.6388 / 0.4856 & 1.14 / 0.3309 / 0.3301 / 0.6234 & 1.61 / 0.1359 / 0.1356 / 0.5631 & 0.28 / 0.0739 / 0.0881 / 0.5050 \\

        \midrule
        & \textbf{Japanese} & \textbf{Korean} & \textbf{Russian} & \textbf{Spanish} \\
        \midrule

        GPT-4o & 7.94 / 0.5857 / 0.8033 / 0.8007 & \textcolor{red!60!black}{\textbf{23.09}} / 0.5677 / 0.7962 / 0.7770 & 32.68 / 0.5645 / 0.7920 / 0.7257 & 62.35 / 0.7378 / 0.7368 / 0.8328 \\
        GPT-5 & 4.56 / 0.6632 / 0.8763 / 0.9199 & 21.57 / 0.6422 / \textcolor{red!60!black}{\textbf{0.8657}} / 0.9136 & \textcolor{red!60!black}{\textbf{38.32}} / 0.7602 / \textcolor{red!60!black}{\textbf{0.8979}} / \textcolor{red!60!black}{\textbf{0.9063}} & 57.18 / 0.7966 / \textcolor{red!60!black}{\textbf{0.8337}} / \textcolor{red!60!black}{\textbf{0.8851}} \\
        Gemini 2.5 Pro & 7.05 / 0.7628 / 0.8654 / 0.9220 & 20.22 / 0.6967 / 0.8600 / 0.9134 & 36.59 / 0.8144 / 0.8864 / 0.8989 & 59.31 / 0.8332 / 0.7897 / 0.8812 \\
        Gemini 2.5 Flash & \textcolor{red!60!black}{\textbf{11.25}} / \textcolor{red!60!black}{\textbf{0.7857}} / 0.8804 / 0.9210 & 19.97 / \textcolor{red!60!black}{\textbf{0.7056}} / 0.8591 / 0.9068 & 35.36 / 0.7974 / 0.8850 / 0.8943 & \textcolor{red!60!black}{\textbf{62.50}} / \textcolor{red!60!black}{\textbf{0.8420}} / 0.8141 / 0.8822 \\
        Claude 4.5 Sonnet & 7.19 / 0.7181 / 0.8690 / 0.9185 & 19.65 / 0.6232 / 0.8511 / 0.9082 & 33.39 / 0.7591 / 0.8797 / 0.9001 & 59.01 / 0.8052 / 0.7864 / 0.8719 \\
        Grok 4 & 5.58 / 0.7242 / 0.8613 / 0.9137 & 21.01 / 0.6729 / 0.8484 / 0.9028 & 36.10 / 0.7832 / 0.8815 / 0.8949 & 60.50 / 0.8371 / 0.8218 / 0.8735 \\
        Qwen3-Omni & 11.18 / 0.7615 / \textcolor{red!60!black}{\textbf{0.8824}} / \textcolor{red!60!black}{\textbf{0.9249}} & 22.91 / 0.6920 / 0.8639 / \textcolor{red!60!black}{\textbf{0.9144}} & 36.70 / \textcolor{red!60!black}{\textbf{0.8161}} / 0.8950 / 0.8997 & 57.08 / 0.8156 / 0.7895 / 0.8736 \\
        IXC2.5-OL & 0.04 / 0.0251 / 0.5418 / 0.2983 & 0.00 / 0.0103 / 0.5536 / 0.3288 & 0.00 / 0.0126 / 0.5419 / 0.2560 & 0.00 / 0.0310 / -0.1168 / 0.3827 \\
        Baichuan-Omni-1.5 & 0.98 / 0.1607 / 0.7055 / 0.6885 & 2.38 / 0.1285 / 0.6797 / 0.6401 & 4.96 / 0.1655 / 0.6920 / 0.5951 & 21.02 / 0.3186 / 0.3057 / 0.6765 \\
        MiniCPM-o 2.6 & 0.34 / 0.1250 / 0.6641 / 0.5785 & 0.53 / 0.0973 / 0.6446 / 0.5473 & 2.55 / 0.1591 / 0.6572 / 0.5010 & 13.11 / 0.2867 / 0.2871 / 0.6490 \\
        Ola & 0.37 / 0.0647 / 0.6597 / 0.5512 & 0.29 / 0.0709 / 0.6528 / 0.5354 & 0.13 / 0.0629 / 0.6293 / 0.4790 & 0.08 / 0.0888 / 0.0590 / 0.5483 \\
        M4 & 0.62 / 0.0743 / 0.6004 / 0.5338 & 0.65 / 0.0747 / 0.6436 / 0.5331 & 0.41 / 0.0674 / 0.6348 / 0.4898 & 0.50 / 0.1165 / 0.0255 / 0.5414 \\
        video SALMONN 2+ & 2.56 / 0.2560 / 0.7147 / 0.7691 & 0.64 / 0.0788 / 0.6434 / 0.5342 & 1.53 / 0.1450 / 0.6591 / 0.7080 & 4.67 / 0.2705 / 0.2395 / 0.6965 \\
        ShareGPT4Video & 0.38 / 0.2026 / 0.7099 / 0.6501 & 1.90 / 0.2120 / 0.7228 / 0.6636 & 1.14 / 0.1194 / 0.6548 / 0.4470 & 3.65 / 0.2966 / 0.4418 / 0.7173 \\
        LongVA & 1.97 / 0.2431 / 0.7471 / 0.7347 & 3.28 / 0.1996 / 0.7162 / 0.6786 & 9.55 / 0.3523 / 0.7655 / 0.7638 & 17.46 / 0.4295 / 0.4566 / 0.7598 \\
        LLaVA-OneVision & 0.40 / 0.0785 / 0.6781 / 0.5876 & 1.84 / 0.0862 / 0.6705 / 0.5557 & 2.86 / 0.0921 / 0.6610 / 0.5122 & 0.08 / 0.0738 / -0.0625 / 0.4386 \\
        LongVU & 0.12 / 0.0086 / 0.5643 / 0.3905 & 0.05 / 0.0165 / 0.5707 / 0.4081 & 0.09 / 0.0589 / 0.6052 / 0.3782 & 0.00 / 0.0115 / -0.0576 / 0.4780 \\
        VideoLLaMA3 & 0.72 / 0.1180 / 0.6225 / 0.6252 & 3.10 / 0.2435 / 0.6636 / 0.7038 & 1.60 / 0.1206 / 0.6347 / 0.6021 & 3.69 / 0.2278 / 0.0892 / 0.6524 \\
        InternVL3 & 2.81 / 0.2465 / 0.7351 / 0.7752 & 2.71 / 0.2361 / 0.7090 / 0.7599 & 1.00 / 0.1283 / 0.6582 / 0.7496 & 5.02 / 0.2581 / 0.2024 / 0.7276 \\
        Video-XL-2 & 0.10 / 0.1193 / 0.6745 / 0.5771 & 0.80 / 0.0695 / 0.6617 / 0.5362 & 0.73 / 0.0901 / 0.6475 / 0.5228 & 1.62 / 0.1641 / 0.1475 / 0.5780 \\

        \midrule
        \bottomrule
    \end{tabular}
    }
    \caption{\textbf{Evaluation across eight target languages, with results formatted as ``BLEU / ROUGE-L / BERTScore / COMET22''.}}
    \label{fig: translation}
\end{table*}

\subsection{How Much Do MLLMs Rely on Prior Knowledge?}
Instructional videos typically contain a high density of domain-specific knowledge, much of which overlaps with factual, conceptual, or procedural information already encoded in large pretrained MLLMs.
To assess whether SP questions can be answered without any perceptual evidence, we evaluate a prompt-only setting where no video, audio, or subtitle input is provided and models receive only the textual query. This experiment probes the extent to which responses rely purely on prior knowledge or dataset biases, thereby revealing whether individual QA pairs genuinely require visual context and whether the task formulation appropriately enforces grounding in the video content. 

As shown in Figure \ref{fig: SP Prior Knowledge Probing}, accuracy drop under this QA pairs-only setting indicates that models cannot reliably answer instructional questions using prior knowledge alone; instead, effective reasoning in SP tasks requires integrating information from video and audio/subtitle cues.

\subsection{Can MLLMs Reliably Reason Over Classroom Problems via OCR?}
To assess whether modern MLLMs can handle OCR-based academic reasoning, we evaluate proprietary MLLMs, open-source omni MLLMs, and open-source video MLLMs across nine representative university-level courses. Figure \ref{fig: different class} illustrates that proprietary MLLMs consistently outperform the others, with Gemini 2.5 Pro and GPT-5 achieving the highest accuracy, often exceeding 90\% in structured CS courses such as Computer Networks and Computer Organization \& Architecture.

In contrast, open-source omni MLLMs show moderate performance but exhibit noticeable instability across subjects, suggesting weaker domain reasoning. Video MLLMs perform the worst overall, reflecting their limited ability to provide correct solutions for assignment-style questions. Their accuracy drops severely in symbol-heavy courses such as Signals Systems or Discrete Mathematics, confirming that current video-oriented architectures remain insufficient to handle structured academic problem solving.

\subsection{Do MLLMs Truly Understand Audio Content?} 
To assess whether audio-capable MLLMs genuinely comprehend auditory content—rather than relying on language priors—we re-evaluate them under a controlled setting where the ground-truth transcript of the first subtask \textit{Speech Transcription} is directly provided to the second subtask \textit{Contextual Listening}. This removes the need for real audio parsing and isolates the effect of perfect acoustic cues.

As shown in Table \ref{tab: ac for audio mllms}, most models exhibit clear accuracy gains once the true transcript is supplied, revealing that their original performance was constrained by limited audio understanding rather than downstream reasoning.


\subsection{Supplementary Ablation Experiment}
In the main paper, we only performed subtitle/audio ablation for Streaming Perception, OCR-Reasoning, and Instructional Prediction. Here, we extend the analysis to Temporal Awareness and the \textit{Video Summarization} subtask of Advanced Expression to evaluate whether additional modalities improve performance.

Figure \ref{fig: TA and AE} shows that for Temporal Awareness, subtitles and audio provide mixed results. Strong models such as GPT-5 and Gemini 2.5 Pro gain substantially from subtitles (+7.58 and +12.90, respectively), while some models exhibit slight degradation, indicating inconsistent multimodal fusion and sensitivity to noisy cues.

For Video Summarization, almost all models benefit significantly from subtitles (e.g., MiniCPM-o 2.6: +40.74, Ola: +42.75), highlighting that models rely heavily on textual cues rather than directly extracting fine-grained visual details. These results suggest that current MLLMs still struggle with detailed video perception, and subtitles largely compensate for this limitation.

\subsection{How Well Do Models Translate Across Multiple Languages?}
We evaluate all models on the \textit{Knowledge Translation} subtask of Advanced Expression across eight target languages using BLEU, ROUGE, BERTScore, and COMET22. As shown in Table \ref{fig: translation}, the results show that proprietary MLLMs, such as Gemini 2.5 Pro/Flash and GPT-5, generally achieve strong performance in European languages. 
Notably, Qwen3-Omni performs competitively across multiple languages, including Chinese, Japanese, and Korean, likely due to its extensive multilingual pretraining and robust cross-lingual representation capabilities.
However, open-source omni and video MLLMs exhibit limited performance, with scores remaining low in most languages, indicating challenges in consistent high-quality multilingual translation.

\section{Failure Case} \label{sec: Failure Case}
\begin{figure*}[t]
    \centering
    \includegraphics[width=\linewidth]{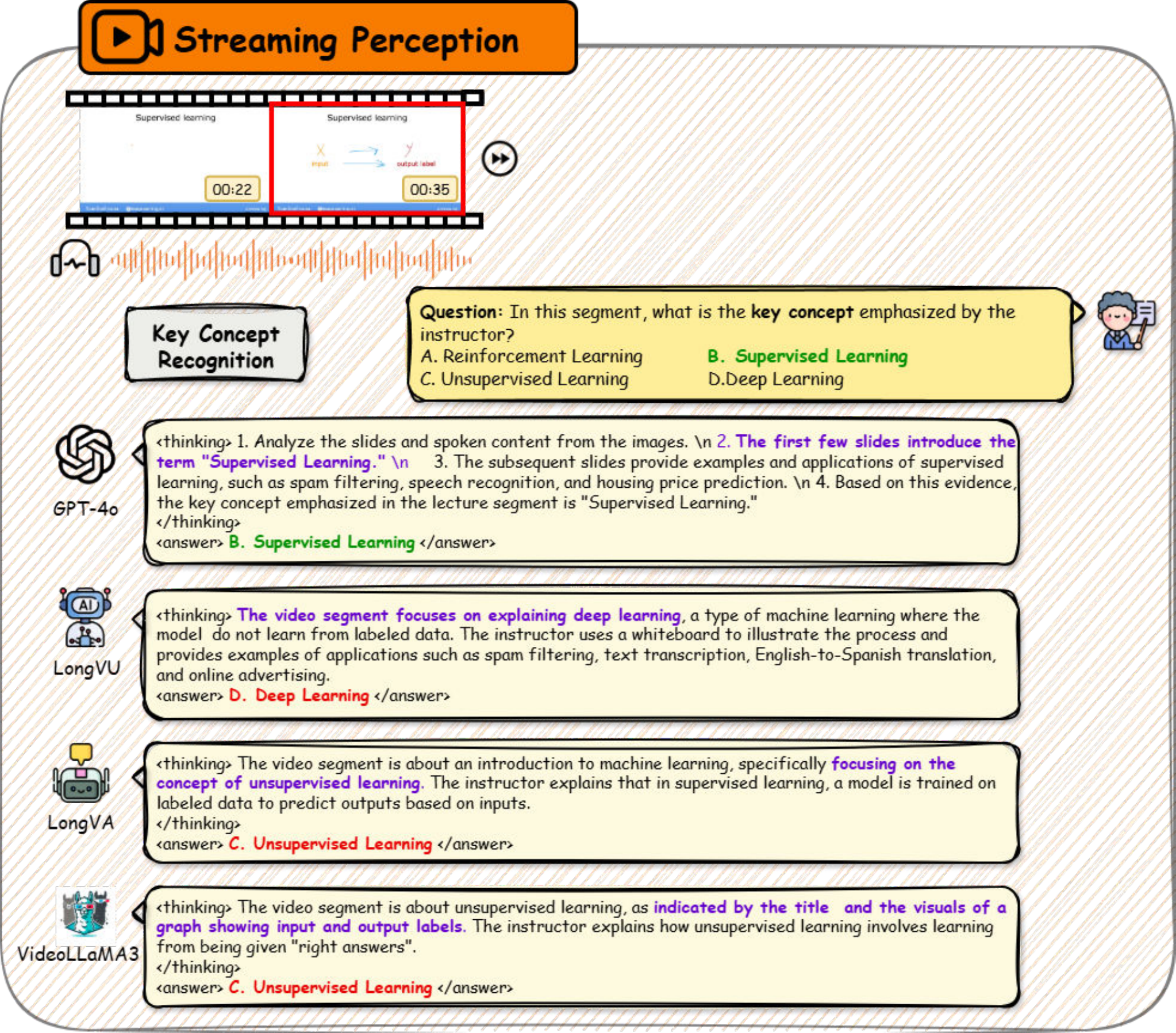}
    \caption{\textbf{Failure Case of key concept perception in Streaming Perception.}}
    \label{fig: key concept perception}
\end{figure*}

\begin{figure*}[t]
    \centering
    \includegraphics[width=\linewidth]{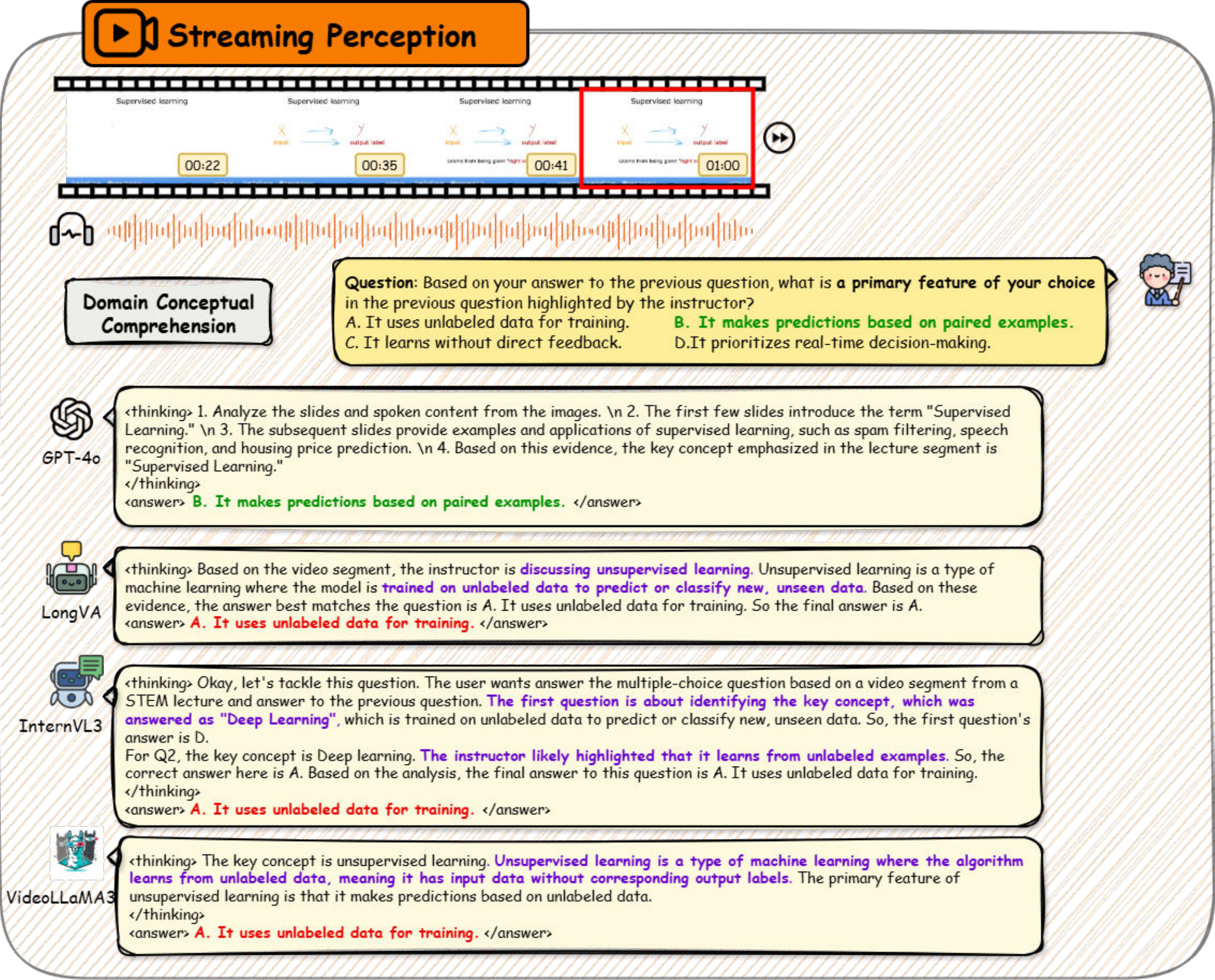}
    \caption{\textbf{Failure Case of domain conceptual comprehension in Streaming Perception.}}
    \label{fig: domain conceptual comprehension}
\end{figure*}

\begin{figure*}[t]
    \centering
    \includegraphics[width=\linewidth]{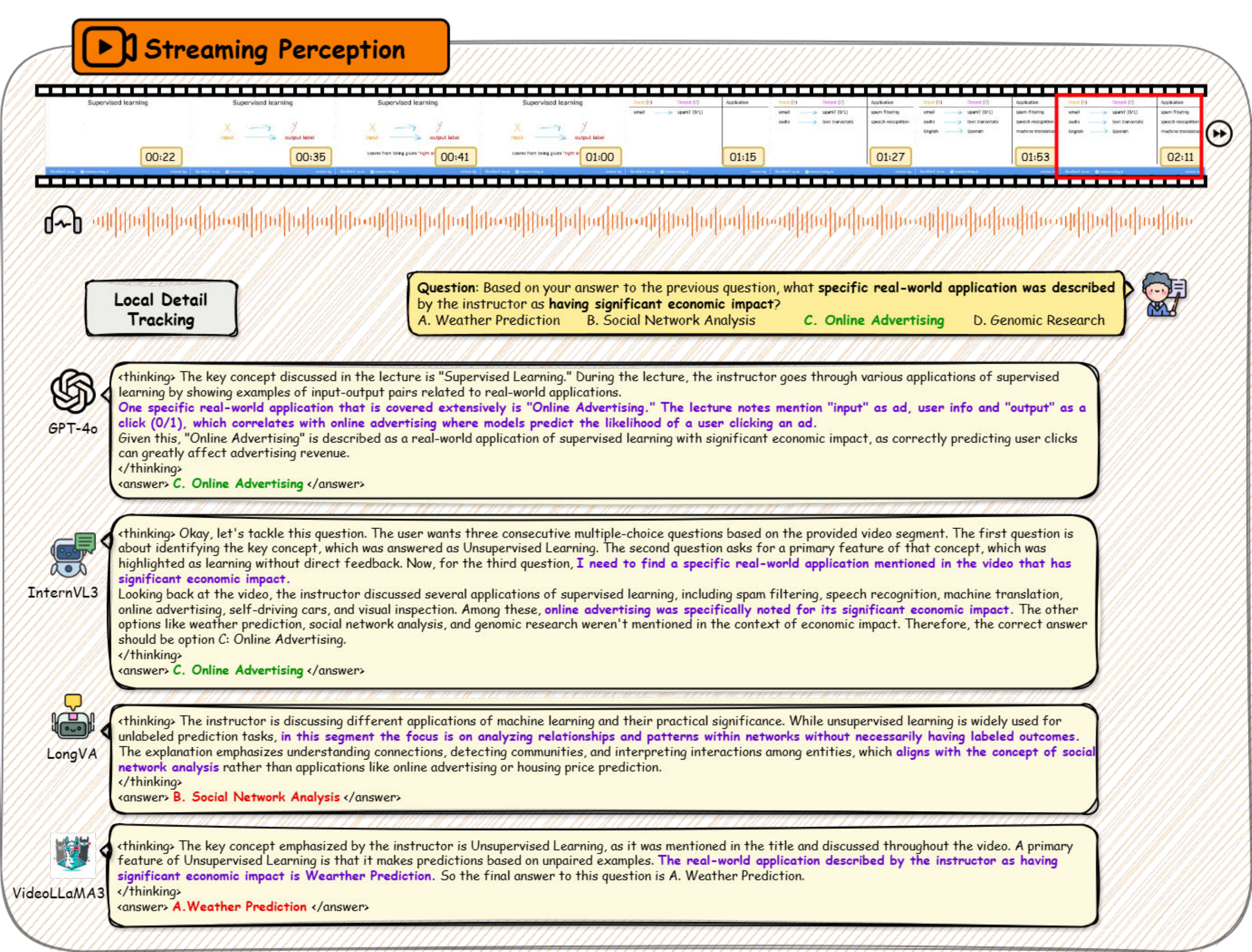}
    \caption{\textbf{Failure Case of local detail tracking in Streaming Perception.}}
    \label{fig: local detail tracking}
\end{figure*}

\begin{figure*}[t]
    \centering
    \includegraphics[width=\linewidth]{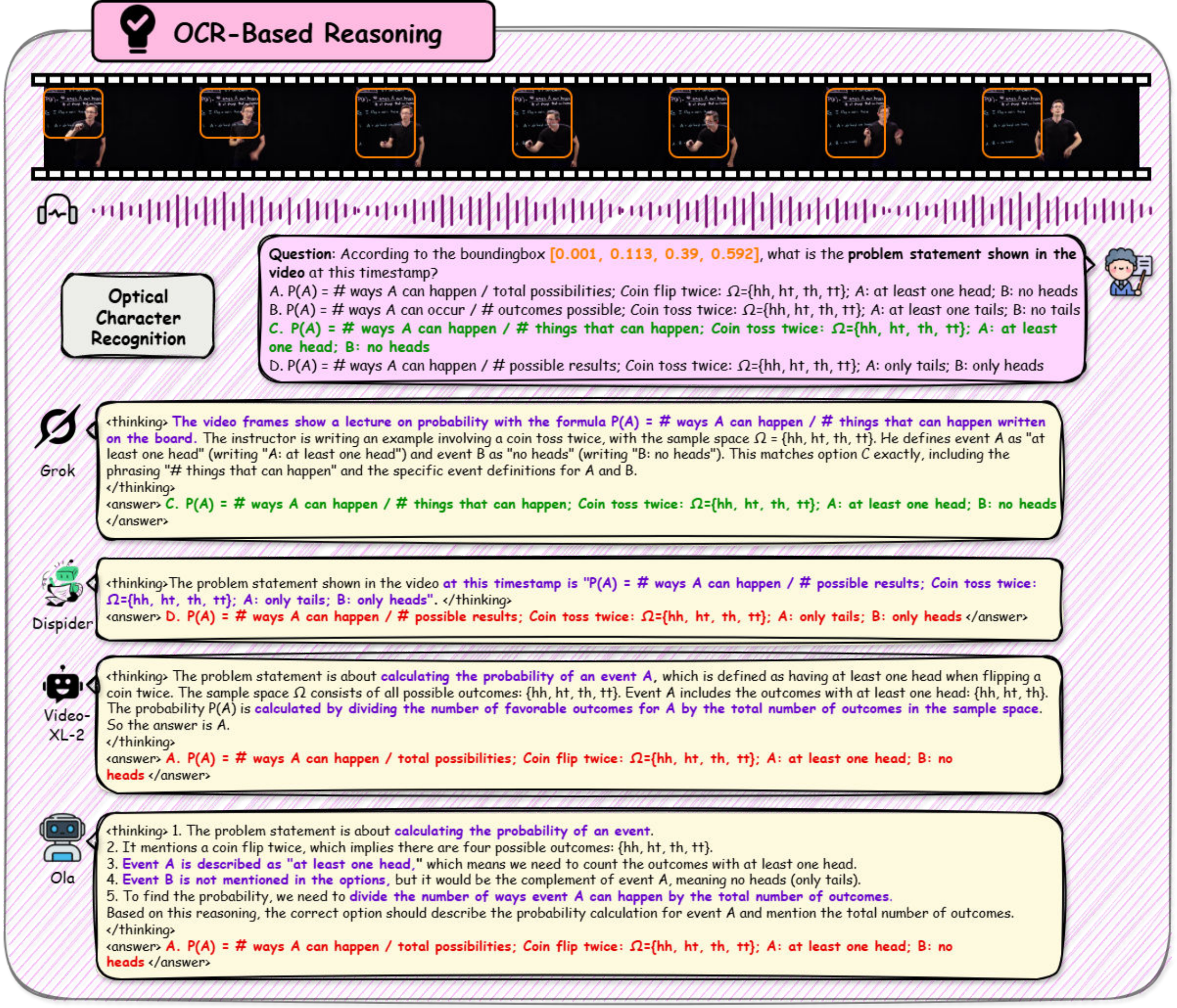}
    \caption{\textbf{Failure Case of optical character recognition in OCR-Based Reasoning.}}
    \label{fig: optical character recognition}
\end{figure*}

\begin{figure*}[t]
    \centering
    \includegraphics[width=\linewidth]{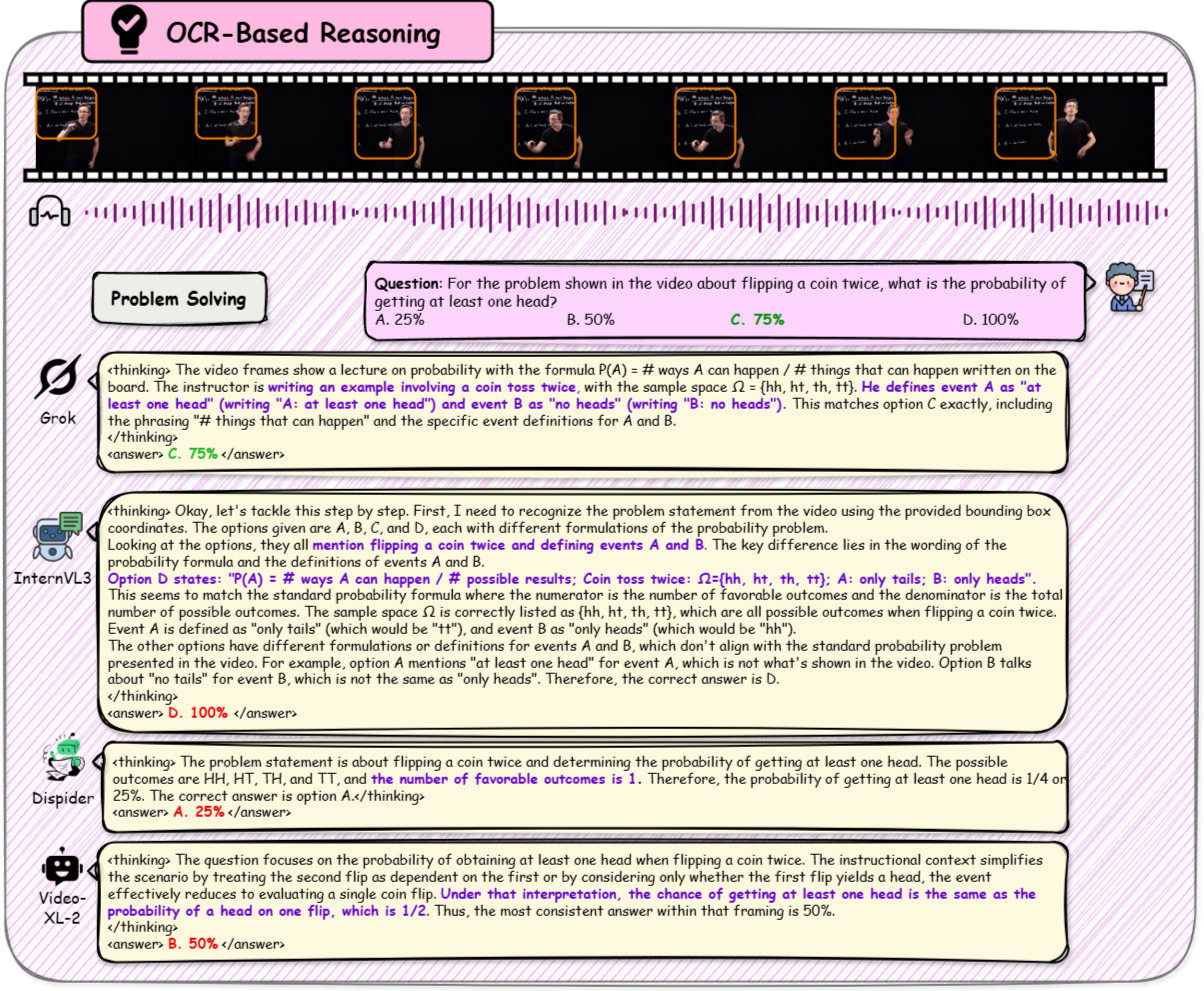}
    \caption{\textbf{Failure Case of problem solving in OCR-Based Reasoning.}}
    \label{fig: problem solving}
\end{figure*}

\begin{figure*}[t]
    \centering
    \includegraphics[width=\linewidth]{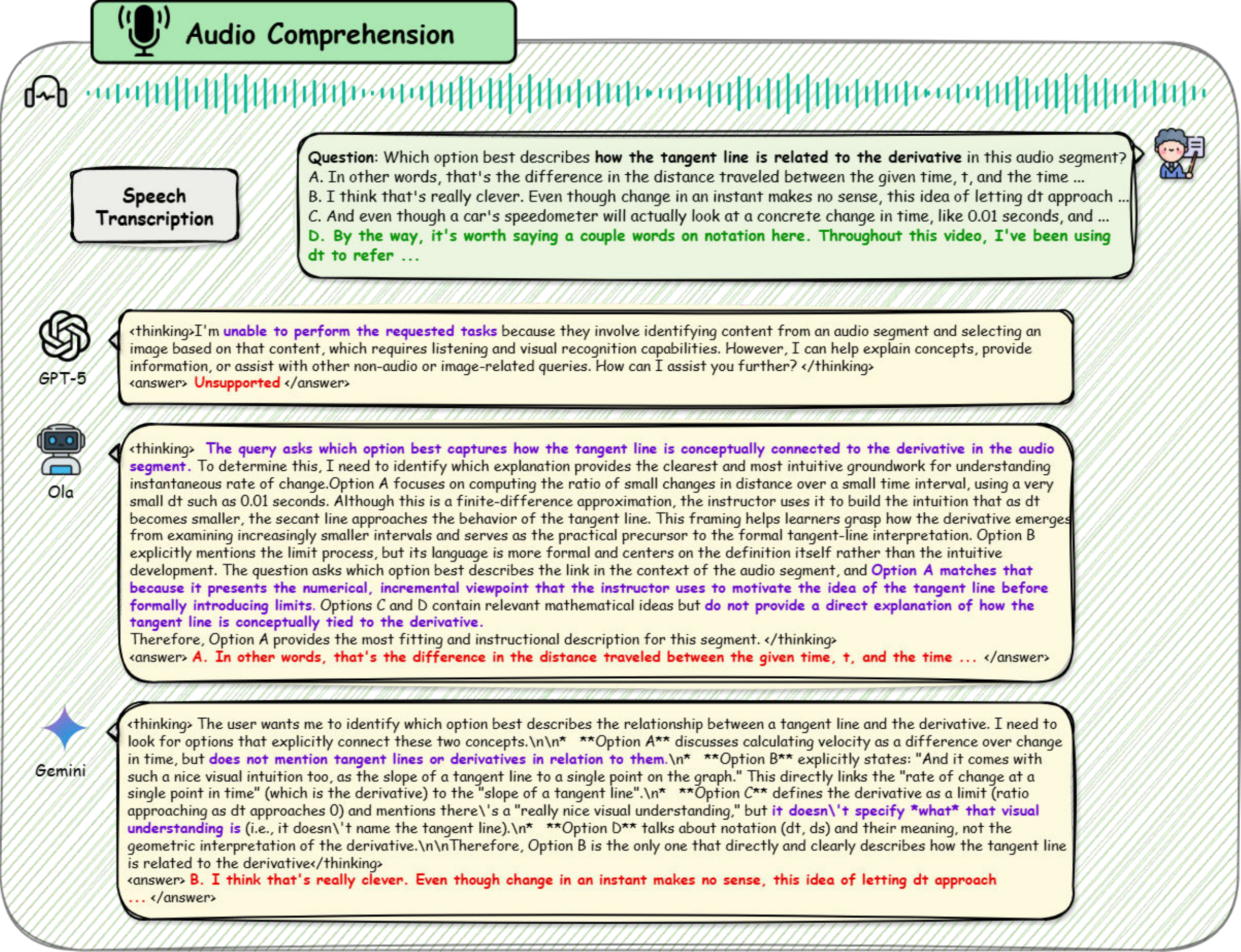}
    \caption{\textbf{Failure Case of speech transcription in Audio Comprehension.}}
    \label{fig: speech transcription}
\end{figure*}

\begin{figure*}[t]
    \centering
    \includegraphics[width=\linewidth]{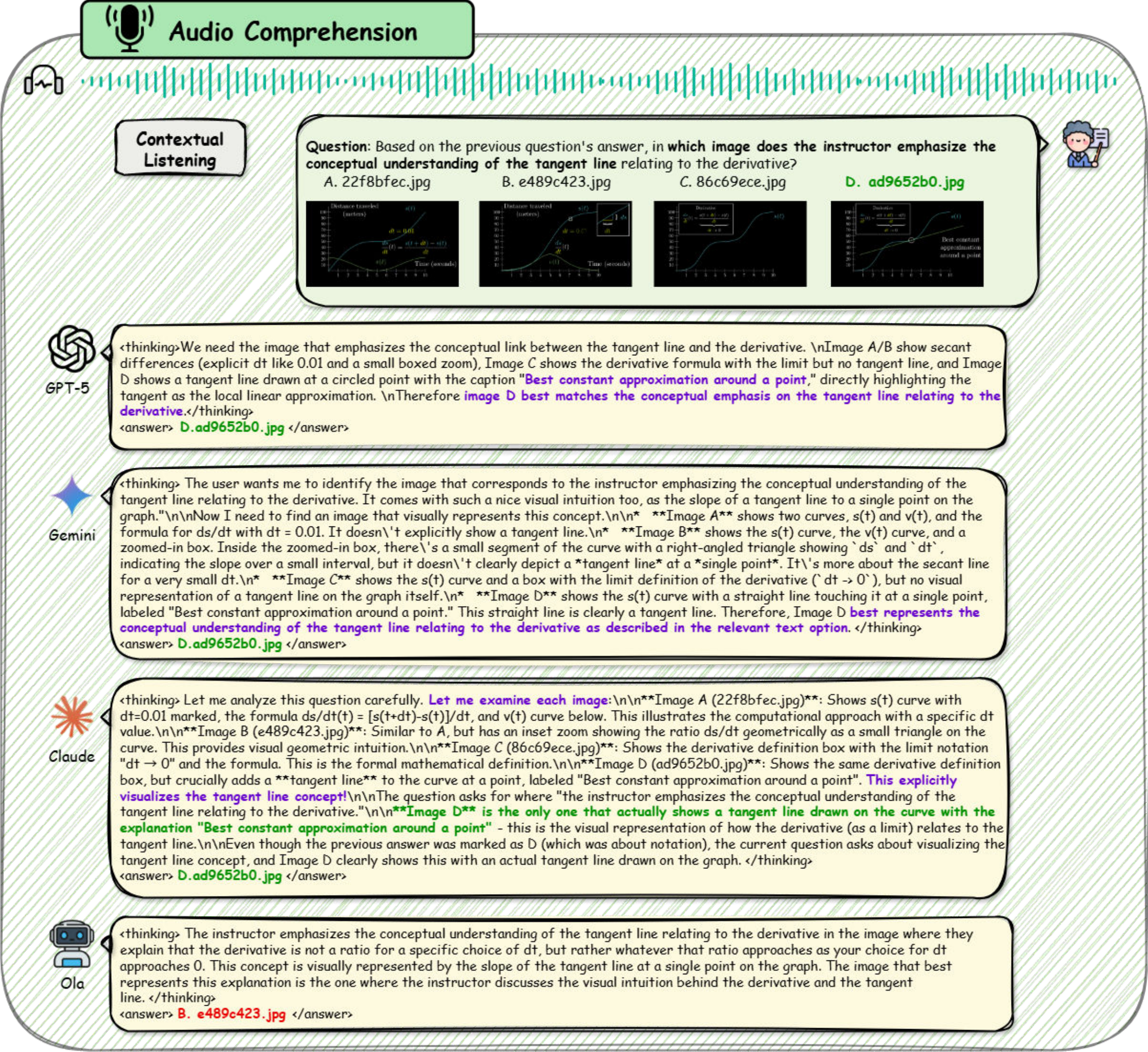}
    \caption{\textbf{Failure Case of contextual listening in Audio Comprehension.}}
    \label{fig: contextual listening}
\end{figure*}

\begin{figure*}[t]
    \centering
    \includegraphics[width=\linewidth]{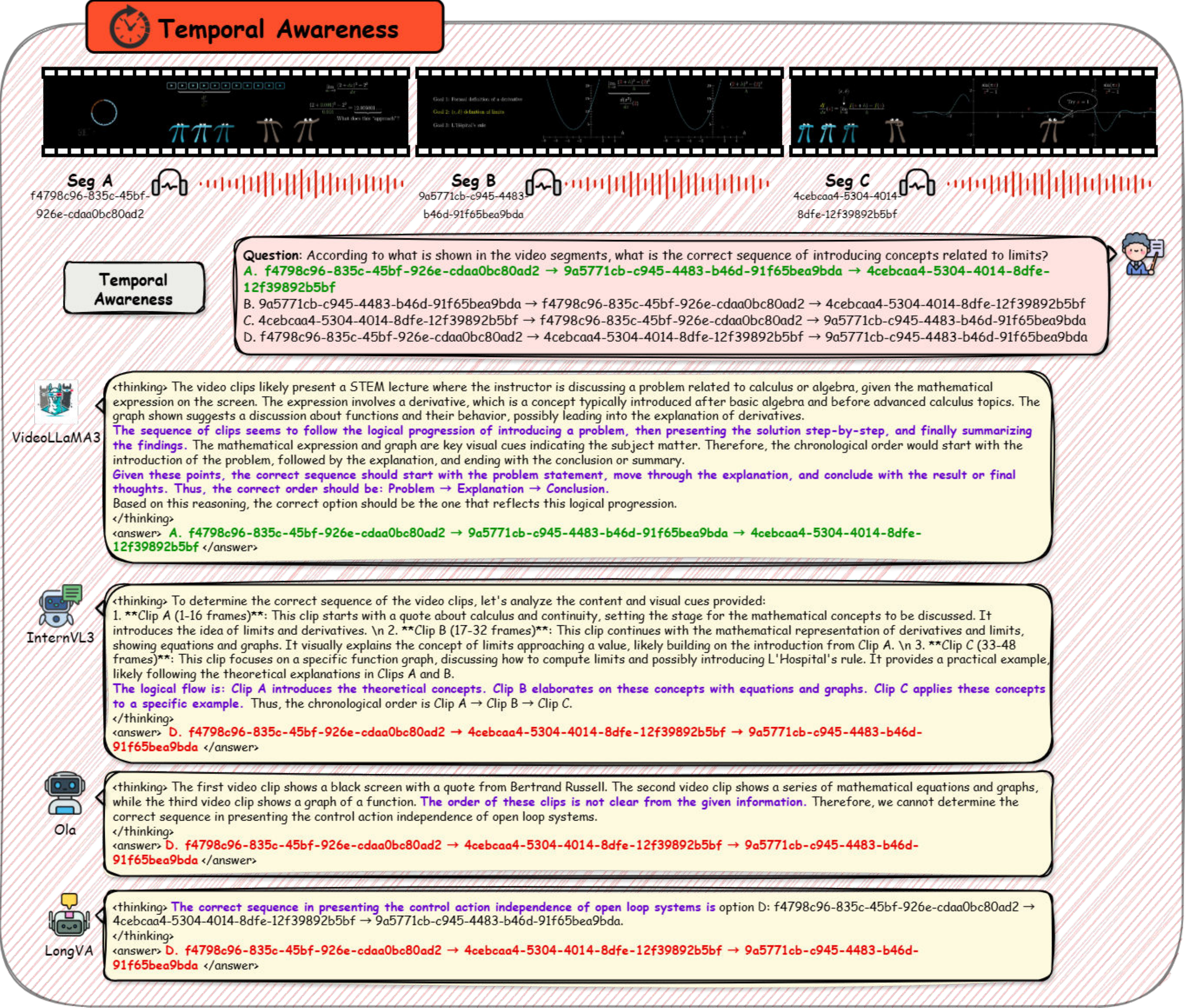}
    \caption{\textbf{Failure Case of Temporal Awareness.}}
    \label{fig: temporal awareness}
\end{figure*}

\begin{figure*}[t]
    \centering
    \includegraphics[width=\linewidth]{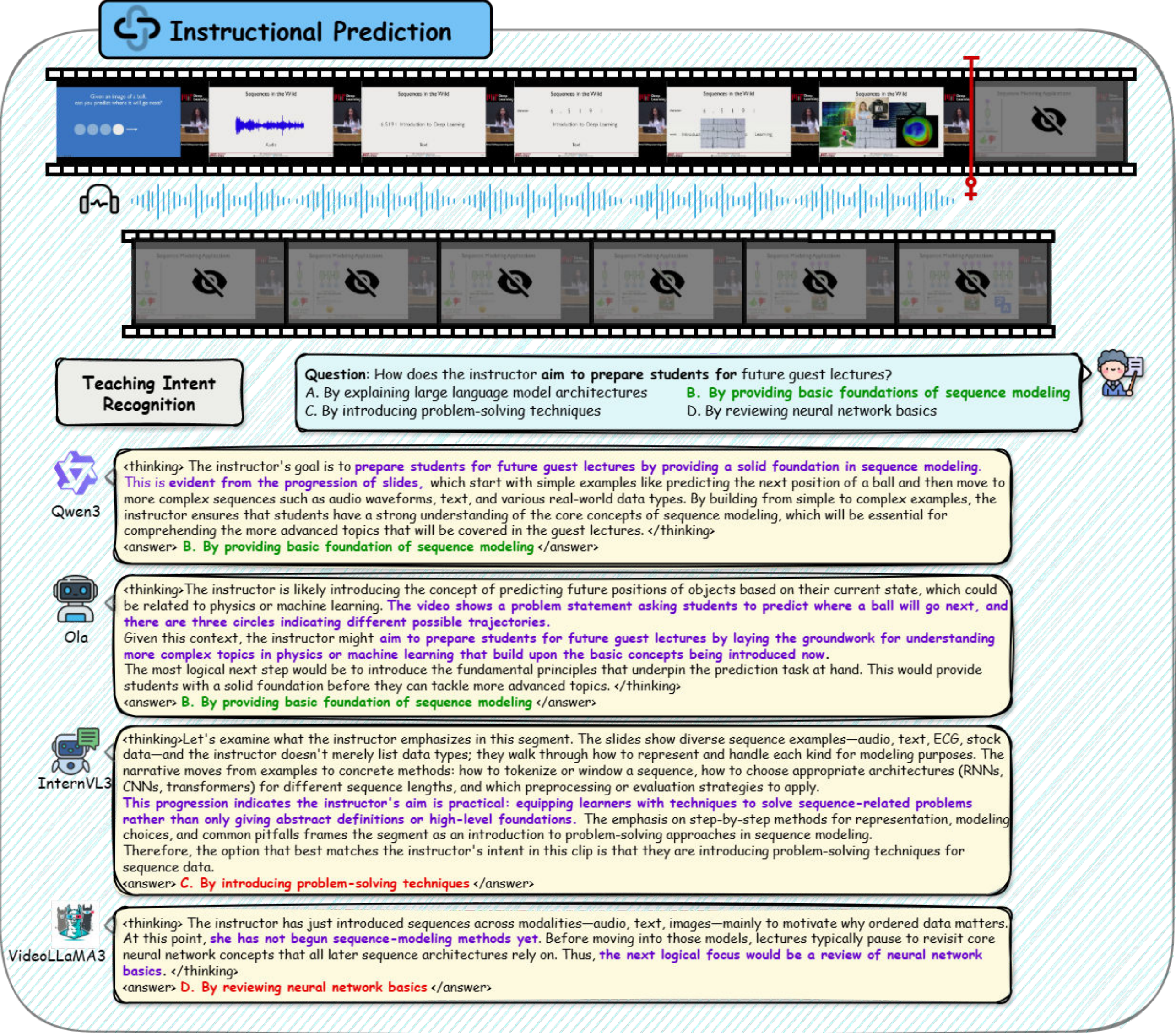}
    \caption{\textbf{Failure Case of teaching intent recognition in Instructional Prediction.}}
    \label{fig: teaching intent recognition}
\end{figure*}

\begin{figure*}[t]
    \centering
    \includegraphics[width=\linewidth]{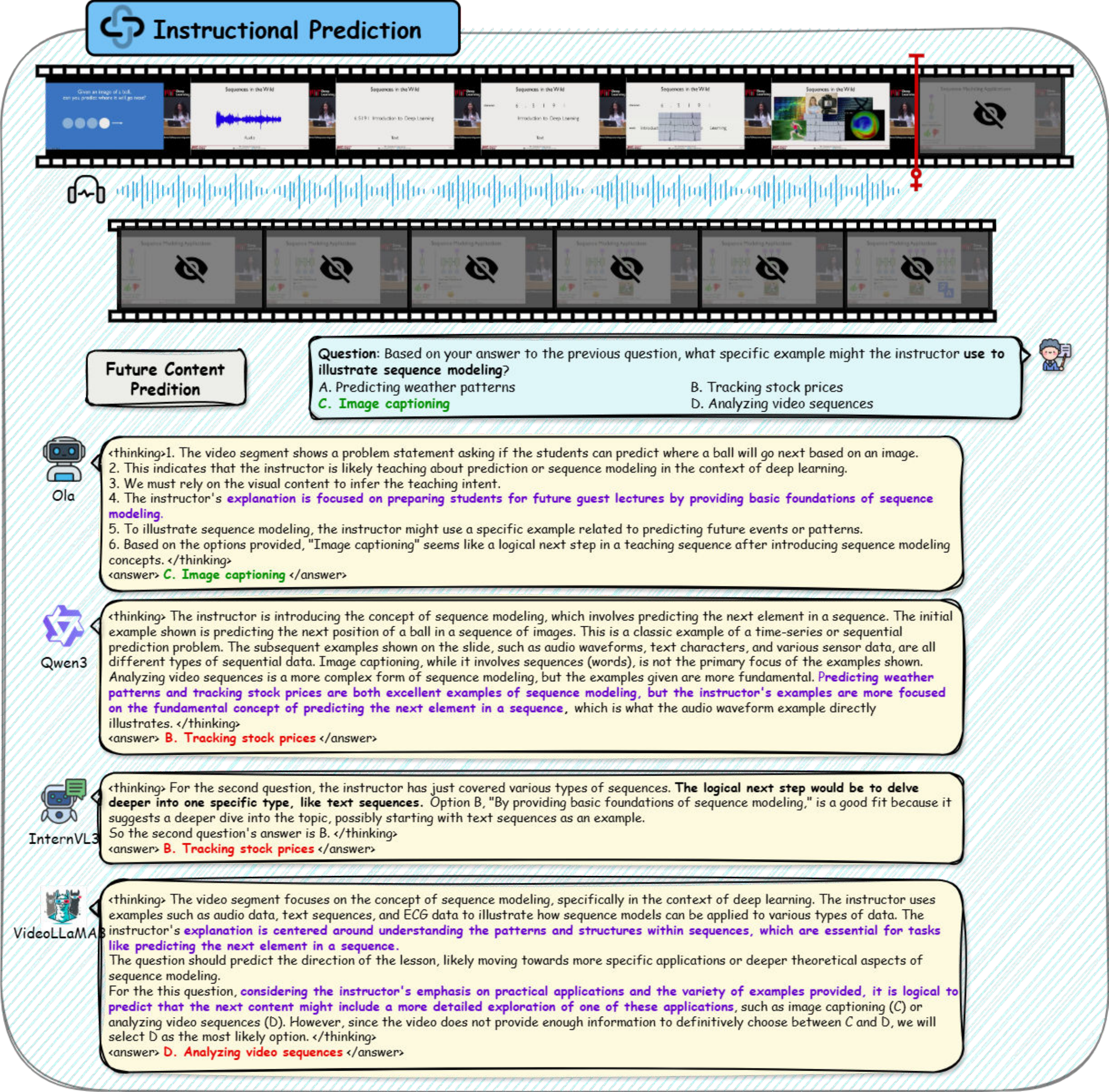}
    \caption{\textbf{Failure Case of future content prediction in Instructional Prediction.}}
    \label{fig: future content prediction}
\end{figure*}

\begin{figure*}[t]
    \centering
    \includegraphics[width=\linewidth]{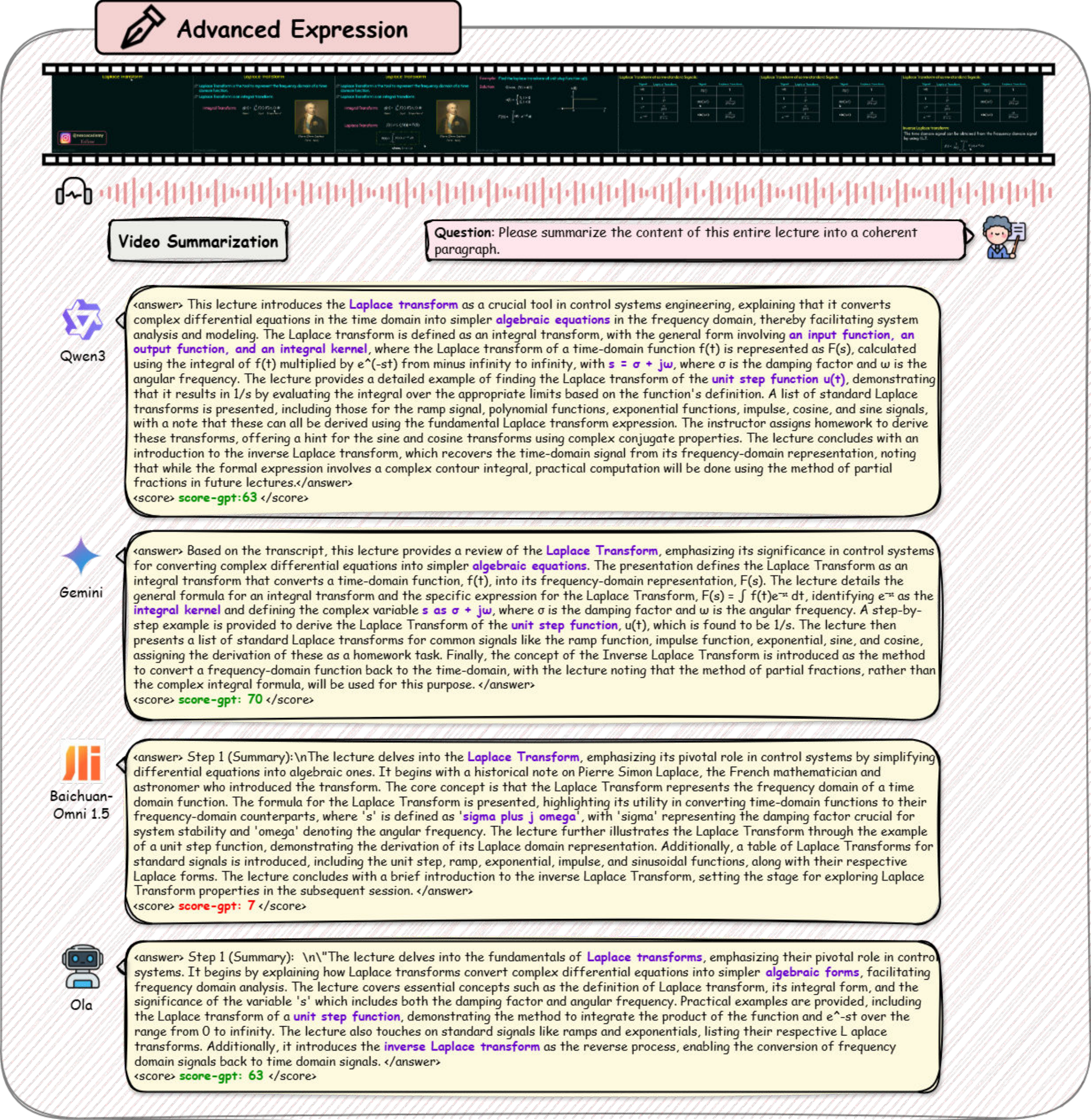}
    \caption{\textbf{Failure Case of video summarization in Advanced Expression.}}
    \label{fig: video summarization}
\end{figure*}

\begin{figure*}[t]
    \centering
    \includegraphics[width=\linewidth]{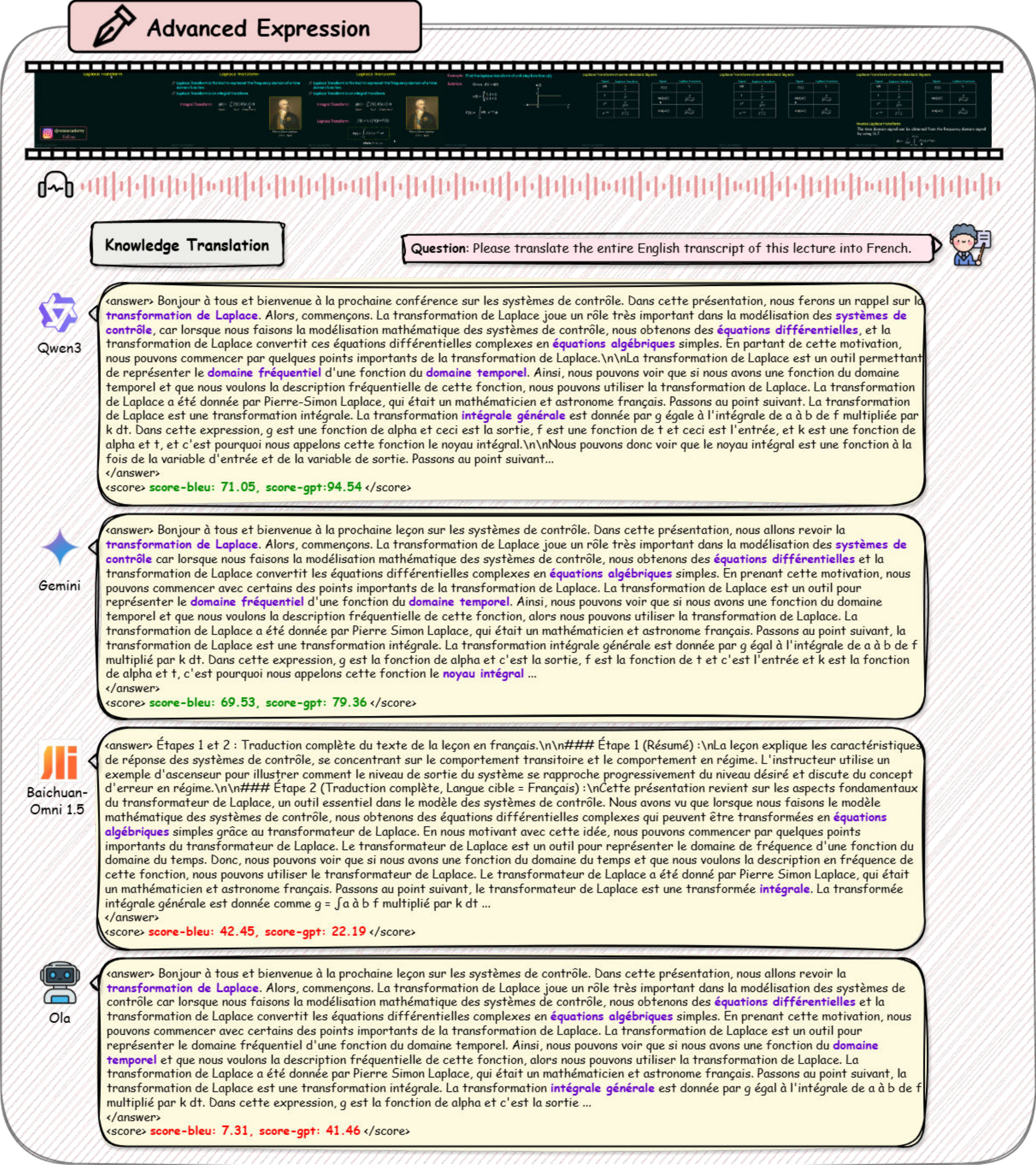}
    \caption{\textbf{Failure Case of knowledge translation in Advanced Expression.}}
    \label{fig: knowledge translation}
\end{figure*}

We present additional visualizations of our LEMON failure cases in Figure \ref{fig: key concept perception}, \ref{fig: domain conceptual comprehension}, \ref{fig: local detail tracking}, \ref{fig: optical character recognition}, \ref{fig: problem solving}, \ref{fig: speech transcription}, \ref{fig: contextual listening}, \ref{fig: temporal awareness}, \ref{fig: teaching intent recognition}, \ref{fig: future content prediction}, \ref{fig: video summarization}, and \ref{fig: knowledge translation}.

Across the first three subtasks of \textbf{Streaming Perception (SP)}, evaluated models demonstrate consistent failure patterns. Models frequently fail to accurately extract relevant content, producing answers that contradict the observed event. Distinguishing semantically similar concepts is challenging, particularly when multiple concepts appear simultaneously, and errors in early steps tend to propagate, compounding subsequent mistakes. In addition, errors in earlier steps can propagate, further compounding mistakes in subsequent decisions. These issues underscore the difficulty of fine-grained visual understanding and precise concept selection in instructional scenarios.

In \textbf{OCR-Based Reasoning (OR)}, models frequently fail to recognize text in designated video regions, resulting in incorrect answers. Errors in the initial questions also propagate, further compounding subsequent mistakes. Models occasionally perform incorrect calculations and misinterpret concepts during reasoning, resulting in answers that are inconsistent with the video content. These failure patterns highlight challenges in integrating multimodal perception, numerical reasoning, and concept-level inference in instructional videos.

For the two subtasks of \textbf{Audio Comprehension (AC)}, these models exhibit several failure modes. In the first subtask, models struggle with audio understanding, misinterpret audio content, or fail to align audio information with options. In the second subtask, errors arise from cross-modal association failures, where audio cues cannot be correctly matched to corresponding visual content or options. These failures highlight challenges in audio understanding and multi-modal correspondence in instructional video scenarios.

For \textbf{Temporal Awareness (TA)}, models show three main failure modes. First, they frequently fail at cross-modal association, unable to correctly link video segments with the corresponding options. Second, the model struggles to perform reasoning based on the information already observed, leading to errors in temporal inference. Third, the model sometimes generates answers without properly integrating the relevant video content, effectively hallucinating information that is not grounded in the observed input. These failure patterns highlight challenges in multi-modal temporal reasoning and the risk of ungrounded predictions.

Across two subtasks of \textbf{Instructional Predicting (IP)}, models consistently struggle with content comprehension and reasoning. In Subtask 1, misinterpretation of video content leads to incorrect answers. In Subtask 2, models have difficulty selecting the most appropriate concept when multiple candidates appear and often fail to incorporate video content into their reasoning. These failures reflect challenges in fine-grained content understanding and context-aware inference.

In \textbf{Advanced Expression (AE)}, models fail to perceive fine-grained content, leading to inaccurate or incomplete answers. Second, some of the models does not have multi-language support, resulting in errors when processing non-English content. These limitations highlight challenges in fine-grained visual understanding and multi-lingual translation.

\end{document}